\documentclass{article} 
\usepackage{iclr2027_conference,times}

\usepackage{amsmath,amsfonts,bm}

\def\eqref#1{equation~\ref{#1}}

\def\1{\bm{1}}

\DeclareMathAlphabet{\mathsfit}{\encodingdefault}{\sfdefault}{m}{sl}
\SetMathAlphabet{\mathsfit}{bold}{\encodingdefault}{\sfdefault}{bx}{n}

\usepackage{hyperref}
\usepackage{url}

\usepackage{amsmath}
\usepackage{adjustbox}
\usepackage{booktabs} 
\usepackage{amsfonts}
\usepackage{array}
\usepackage{xcolor}
\usepackage{multirow}
\usepackage[table]{xcolor}
\usepackage{pifont}
\definecolor{groupgray}{HTML}{F5F6F8}
\definecolor{proposedbg}{HTML}{EAF4EC}
\definecolor{bestgreen}{HTML}{0B5D2A}
\definecolor{mutedgray}{HTML}{6B7280}
\newcommand{\cmark}{\textcolor{bestgreen}{\ding{51}}}
\newcommand{\best}[1]{\textbf{\color{bestgreen}#1}}

\usepackage{wrapfig}
\definecolor{groupgray}{RGB}{235,235,235}
\definecolor{proposedbg}{RGB}{230,242,255}
\usepackage{pifont} 
\usepackage[table]{xcolor}
\usepackage{natbib}  
\usepackage{caption} 
\usepackage{algorithm}
\usepackage{algorithmic}

\usepackage{newfloat}
\usepackage{listings}
\usepackage[most]{tcolorbox}
\newtcolorbox{promptbox}[1][]{
  enhanced, colback=gray!4, colframe=gray!55, colbacktitle=gray!18, coltitle=black,
  boxrule=0.4pt, arc=1.5pt, left=4pt, right=4pt, top=3pt, bottom=3pt,
  toptitle=1pt, bottomtitle=1pt, fonttitle=\sffamily\scriptsize\bfseries,
  fontupper=\ttfamily\footnotesize,
  before upper={\frenchspacing\raggedright\setlength{\parindent}{0pt}\setlength{\parskip}{0.45\baselineskip}},
  #1}
\newcommand{\pcode}[2]{\hspace*{1em}\makebox[3em][l]{#1}#2}
\newcommand{\pline}[1]{\par\hangindent=2.6em\hangafter=1\hspace*{1em}#1}
\DeclareCaptionStyle{ruled}{labelfont=normalfont,labelsep=colon,strut=off}
\floatstyle{ruled}
\newfloat{listing}{tb}{lst}{}
\floatname{listing}{Listing}
\title{Autoregressive Drillhole Modelling under Distribution Shift}
\iclrfinalcopy

\author{Yihao Ding, Daniel Yitian Su, Yiran Zhang, Christopher M. Gonzalez \& Wei Liu\thanks{Corresponding author: \texttt{wei.liu@uwa.edu.au}} \\
The University of Western Australia \\
Crawley, WA 6009, Australia \\
\texttt{\{yihao.ding, wei.liu, christopher.gonzalez\}@uwa.edu.au} \\
\texttt{\{daniel.su, yiran.zhang\}@research.uwa.edu.au} \\
}

\begin{document}

\maketitle \lhead{Preprint}

\begin{abstract}
Autoregressive modelling has achieved remarkable success in language and sequence tasks by learning to predict future states from previous observation. Mineral-exploration drillholes provide a natural but largely unexplored setting for this paradigm: as drilling proceeds, lithology is revealed sequentially from shallow to deep, making prediction of deeper strata inherently autoregressive. Existing drillhole modelling, however, is dominated by spatial interpolation and reconstruction, or largely rely on masked modelling, leaving strictly autoregressive prediction largely underexplored. We introduce DrillBench, a benchmark of 49,671 Western Australian drillholes for next-layer prediction and autoregressive stratigraphic generation across a graded transfer spectrum, from local prediction through spatial shift to cross geological province transfer. Benchmarking classical, geostatistical, and neural models reveals a clear \emph{transfer boundary}: spatial and geochemical conditioning provides large local gains but deteriorates sharply under stronger shift, whereas lithology-sequence autoregressive models transfer more robustly. Guided by this finding, we develop a backbone-agnostic recipe combining large-scale pretraining on historical drillholes with spatial retrieval of neighbouring lithology. Retrieval is most effective in weathered cover, when local spatial continuity remains informative, whereas pretraining contributes more strongly in bedrock and under broader geological shift. Together, they retain strong local performance while improving generalisation under spatial and cross-province shift, most markedly on the most distant splits. The benchmark and code are available at \url{https://github.com/yihaoding/drillbench}.

\end{abstract}

\section{Introduction}
\label{sec:intro}
Autoregressive modelling has expanded beyond language~\citep{achiam2023gpt,brown2020language} to scientific sequences such as proteins~\citep{madani2023large}, genomes~\citep{nguyen2024sequence}, and general time series~\citep{garza2023timegpt}. By repeatedly predicting the next state from the observed sequence, autoregressive models can learn recurring structural dependencies that support both next-step prediction and free-running generation. Such dependencies may also transfer across settings when similar underlying structure persists. Meanwhile, the machine-learning community has shown growing interest in geoscience, particularly through remote sensing \citep{bai2026geolink,ji2026cf} and Earth observation \citep{danish2026terrafm}. However, most of this progress focuses on surface and near-surface observations, while subsurface geological sequences, including large drillhole archives, remain comparatively underexplored by modern sequence modelling.

\begin{wrapfigure}{r}{0.55\textwidth}
\vspace{-1.3em}
\centering
\includegraphics[width=\linewidth]{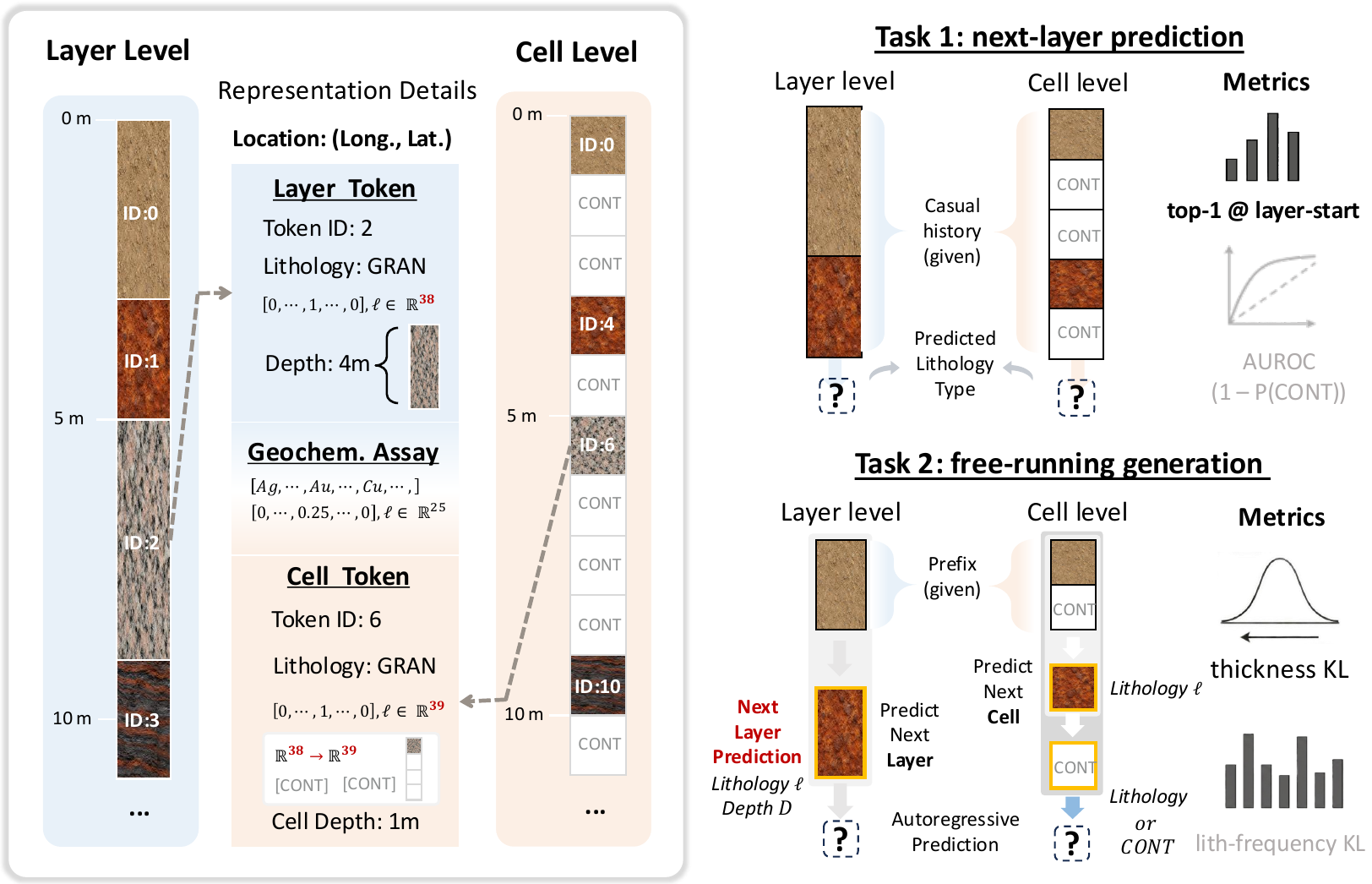}
\caption{Two granularities of drillhole representation and DrillBench task definitions.}
\label{fig:task_definition}
\vspace{-1.5em}
\label{fig:data}
\end{wrapfigure}

The drilling process naturally induces an autoregressive prediction setting (Figure~\ref{fig:data}). As drilling progresses downwards, lithology is revealed layer by layer, so at any stage only the strata already encountered above the drill bit are available, while deeper formations remain unknown. This makes it natural to predict the next stratum from the sequence observed so far and recursively extend the hole to deeper. More importantly, autoregressive modelling can learn recurring stratigraphic dependencies within a hole, for example, a weathering profile progressing from laterite to saprolite and then to fresher bedrock, rather than relying entirely on nearby observations. 
Existing mineral-drillhole modelling, however, is dominated by interpolation, simulation, and 3D reconstruction~\citep{strebelle2002conditional,ren2025deep,li2026generalized}, which primarily exploit spatial relationships between observations. Petroleum well-log studies retain the depth sequence but largely focus on masked representation learning or point-wise classification~\citep{qi2025wlfm,sun2023cross,carvalho2026lithology}. Neither setting directly evaluates whether a model can use the observed stratigraphic sequence to roll a partially drilled hole forward from its own predictions, leaving autoregressive subsurface generation without a public benchmark or shared evaluation protocol.

Thus, we introduce \textbf{DrillBench}, a public benchmark of 49,671 mineral-exploration drillholes, together with about 112k-hole pretraining archive (WA-DASC-full) spanning several tectonic provinces in Western Australia. DrillBench evaluates two tasks: \textit{next-layer prediction}, which predicts the next lithology from the ground-truth history above it, and \textit{free-running generation}, which recursively generates the remaining lithology from an observed prefix. Both tasks are evaluated across progressively larger spatial and geological shifts, from in-distribution prediction through increasing spatial shift to cross-province transfer. Benchmarking a broad range of classical, geostatistical, and neural methods reveals a clear advantage of sequence-only autoregressive modelling: it provides a better balance between in-distribution accuracy and transfer. By relying on the observed sequence rather than location-specific signals, these models degrade more gradually as spatial and geological shift increases. In contrast, direct spatial and geochemical conditioning improves local accuracy but can become harmful under stronger shift, exposing a clear \emph{\textbf{transfer boundary}}.

Building on this transfer-safe autoregressive foundation, we further develop a backbone-agnostic training-and-inference recipe that introduces additional geological information. \textbf{(i) Large-scale pretraining} strengthens transferable stratigraphic structure using historical drillholes, with sparse geochemistry used only as auxiliary supervision. \textbf{(ii) Spatial retrieval} then introduces neighbouring lithology at inference without directly exposing absolute coordinates at test time. The two mechanisms complement the autoregressive model in different geological regimes. Retrieval is most effective in weathered cover within the source province but adds little across provinces. Pretraining shows the opposite trend: it contributes little in-distribution but becomes more valuable in bedrock and under broader geological transfer. Together, these results show how the three components complement each other: autoregressive sequence modelling provides the transferable foundation, retrieval strengthens prediction where local continuity is reliable, particularly in weathered cover, and pretraining becomes more important in bedrock and under broader geological transfer.

Our contributions are:
\textbf{i) Formulation.} We formulate drillhole modelling as generative next-stratum prediction: the model uses only the strata already observed to predict the unseen geology below. We evaluate this through next-stratum prediction and free-running generation with a unified protocol.
\textbf{ii) The transfer boundary.} Coordinates and assays improve local accuracy but degrade sharply under spatial and geological shift, whereas sequence-based models transfer more robustly.
\textbf{iii) Pretraining and retrieval.} We extend autoregressive models with large-scale pretraining on large-scale drillholes and retrieval of neighbouring lithology, without directly conditioning on absolute coordinates or assays. Retrieval is most useful for local continuity in weathered cover, while pretraining contributes more in bedrock and cross-province transfer.
\textbf{iv) Resources.} We release DrillBench with 49,671 drillholes, the WA-DASC-full pretraining corpus, pretrained checkpoints, and evaluation code.


\section{Related Work}
\label{sec:related}
Recent machine-learning research has increasingly explored geoscience through remote sensing \citep{xiao2026geommbench} and Earth observation \citep{li2026reobench}, including large-scale and multimodal foundation models trained on satellite imagery \citep{bai2026geolink,danish2026terrafm,perron2026universat}. These advances primarily model the Earth's surface and near-surface environment, whereas large-scale learning of subsurface geological structure has received substantially less attention.
Mineral-drillhole modelling is dominated by \textit{spatial interpolation and geostatistical simulation}, ~\citep{carle1996transition,strebelle2002conditional,caers2011modeling,ren2025deep,li2026generalized}. These approaches rely heavily on coordinates and neighbouring observations, making them effective for local reconstruction but less suited to learning stratigraphic relationships that may recur across locations. Petroleum well-log research instead models observations along depth, but most methods use masked modelling and bidirectional encoders~\citep{qi2025wlfm,liu2020well,sun2023cross,carvalho2026lithology}. Because they can access information from deeper positions or predict individual depths independently, these formulations do not match the one-directional drilling process in which only the already drilled sequence is available. In addition, existing public benchmarks are generally small and geographically restricted~\citep{hall2016facies,bormann2020force,carvalho2026lithology}, limiting the study of graded transfer, large-scale pretraining, and retrieval-based use of neighbouring drillholes. This paper addresses these gaps by formulating drillhole modelling autoregressively, evaluating prediction and generation across progressively stronger spatial and geological shifts, and using a historical archive to study how pretraining and retrieval contribute to transfer.

\section{Benchmark Protocol}
\label{sec:benchmark}

\subsection{Drillhole Representation}
\label{subsec:tokenization}

A single drillhole representation format does not suit all modelling approaches. Autoregressive models naturally operate on geological layers, while spatial and point-wise methods make predictions at specific depths. We therefore provide two representations of each drillhole: a compact \emph{layer-level} sequence and a \emph{cell-level} sequence defined over depth. 

\textbf{a) Layer-level representation.}
A drillhole is represented as
$\mathcal H^{\mathrm{layer}} = \big((\ell_1,D_1),\ldots,(\ell_L,D_L)\big), $
where $\ell_i\in\mathcal L$ is the lithology of stratum $i$, $D_i>0$ is its thickness, and $\mathcal L$ contains the 37 standardised lithology codes. This representation directly models \emph{which lithology comes next} and \emph{how thick it is}, without repeating the same lithology. It is used by layer-transition baselines such as Markov models and by layer-level neural models such as LSTM and Transformer.
\textbf{b) Cell-level representation.}
The drillhole is represented as a depth-indexed sequence
$ \mathcal H^{\mathrm{cell}} = (x_1,\ldots,x_T), \qquad x_t\in\mathcal L', $
where $\mathcal L'=\mathcal L\cup\{\texttt{CONT},\texttt{UNK}\}$ and \texttt{UNK} denotes missing or unmappable lithology. Given a depth spacing $\Delta d$, the first cell of each stratum carries its lithology code and the remaining cells are encoded as \texttt{CONT}. The number of \texttt{CONT} cells therefore preserves stratum thickness without revealing where the stratum ends in advance. Cell-level sequence models such as LSTM, and Mamba use this sequence directly. For spatial methods such as $k$-NN, and feature-based models such as XGBoost, \texttt{CONT} cells are replaced by the corresponding stratum lithology so that each depth cell has an explicit label.


\subsection{Task Definition}
\label{sec:tasks-metrics}

\subsubsection{Task 1: Next-Layer Prediction}
\label{sec:task1}

\textbf{Definition.}
At each true stratum boundary, the model predicts the lithology of the next stratum using only information available up to that boundary. Ground-truth boundaries are provided so that the task isolates \emph{which lithology comes next}, rather than jointly evaluating boundary or thickness prediction. Layer-level models use the preceding strata, while cell-level models use the corresponding cell sequence.
\textbf{Evaluation.}
All methods are evaluated over the same 37 lithology classes at the ground-truth stratum boundaries. Layer-level models directly predict the 37 lithology classes. Cell-level models also predict \texttt{CONT}; at the true stratum boundaries, we remove \texttt{CONT} and renormalise the lithology probabilities. We report macro-F1 over the lithology classes with at least 20 events in each split as the main metric (Appendix~\ref{app:evaluation}). 
\textbf{Baselines.}
We include methods operating at both levels. Cell-level baselines include frequency and Markov models, geostatistical methods such as spatial $k$-NN and TPROGS~\citep{caers2011modeling,strebelle2002conditional,carle1996transition}, XGBoost~\citep{chen2016xgboost}, and neural sequence models including GRU, LSTM, TCN, Transformer, and Mamba. Layer-level baselines include corresponding transition models and neural sequence architectures applied directly to the stratum sequence.

\subsubsection{Task 2: Free-Running Generation}
\label{sec:task2}
\textbf{Definition.} Given the true first $p\%$ of a drillhole, the model generates the remainder autoregressively. At each step, the model decides whether a new layer begins and, if so, predicts its lithology. Each prediction becomes the next input until the known hole depth is reached. No ground-truth
boundaries are provided, so errors can accumulate.
\textbf{Evaluation.} Our primary metric is the \emph{joint KL},
$\mathrm{KL}(\text{true}|\text{generated})$, between the empirical joint distributions of lithology and layer thickness over each split. Lithology uses the 37 DrillBench classes and thickness seven log-spaced bins, so the metric penalises mismatches in either rock composition or thickness. We average over $M{=}5$ rollouts per hole; lower is better. (Details and additional metrics are reported in Appendix~\ref{app:results_task2}).
\textbf{Baselines.} Markov, frequency, and neural sequence models from Task~1 can use their own predictions as input and therefore support
free-running generation. Spatial methods such as kriging, spatial $k$-NN, and simulation cannot recursively extend a hole and are reported for reference only.




\section{DrillBench Dataset}
\label{sec:data}

\begin{wrapfigure}{r}{0.5\textwidth}
\vspace{-1.5em}
\centering
\includegraphics[width=\linewidth]{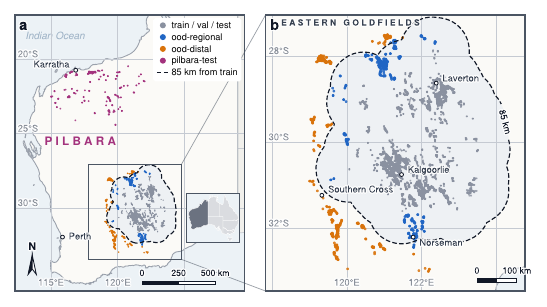}
\vspace{-2em}
\caption{DrillBench six splits spatial layout.}
\vspace{-1.5em}

\label{fig:dataset_analysis}
\end{wrapfigure}

\subsection{Source Corpus}
DrillBench is derived from the Western Australia\footnote{\url{https://dasc.dmirs.wa.gov.au/}}, a public archive of mineral exploration submissions released under CC-BY-4.0. We focus on two geologically distinct areas: the \textit{Eastern Goldfields} (EG; Archean greenstone--granite, gold-dominated), with 46{,}393 drillholes, and the \textit{Pilbara} (Archean BIF iron-formation, iron-dominated), with 3{,}278 drillholes. After curation (\S\ref{subsec:curation}), DrillBench contains \textbf{49{,}671 drillholes and 1{,}499{,}181 depth intervals} collected over 30 years data, stored as nested Apache Parquet files. To our knowledge, this is the first public drillhole corpus formatted explicitly for 1D autoregressive sequence modelling.

\subsection{Dataset Curation}
\label{subsec:curation}

To construct a clean, standardised, and leakage-resistant benchmark, we apply six preprocessing steps (details and per-step counts in Appendix~\ref{app:dataset_curation}).
\textbf{(1) Invalid-record removal.} We remove 26 corrupted records with implausible drill depths ($>10\,\mathrm{km}$) or malformed \texttt{hole\_id} fields. \textbf{(2) Hole-level quality filtering.} We retain drillholes that are at least $30\,\mathrm{m}$ deep, containing at least five logged intervals, and have substantive lithology text for at least half of them, ensuring sufficient context for autoregressive evaluation. \textbf{(3) Lithology vocabulary normalisation.} Free-text descriptions are mapped to a canonical 37-code vocabulary shared across data providers. About 68\% of intervals are mapped; unmapped intervals retain their raw text and are treated as unobserved. \textbf{(4) Physical-bounds masking.} For each of the 25 elements, values above a stoichiometric upper bound (e.g., $\mathrm{Fe}\geq72\%$) are set to \texttt{null}. Negative values within the valid range are retained because they encode below-detection-limit measurements. \textbf{(5) Coordinate harmonisation.} Collar coordinates reported in different zones and datums, including GDA94 and AGD84, are transformed into a common regional reference frame so that splitting, de-duplication, and retrieval use consistent ground distances. We exclude 27 drillholes without valid coordinates. \textbf{(6) Twin-hole de-duplication.} The same physical drillhole may appear in multiple statutory reports under different identifiers, creating evaluation leakage. We remove evaluation holes with a training-set twin within $25\,\mathrm{m}$ whose lithology sequence is identical or whose interval grid matches within $0.11\,\mathrm{m}$. This removes 1{,}051 duplicate holes from \textsc{val} and \textsc{test}, leaving 49{,}671 drillholes and 1{,}499{,}181 intervals.

\subsection{Drillhole Data Modalities}
Each drillhole comprises hole-level metadata and a sequence of depth intervals. (1)~\textit{Collar metadata} provides drillhole-level context, including location, elevation, region and planned maximum depth. (2)~\textit{Lithology} records the rock type at each depth interval and serves as the main prediction target. Each mapped interval is assigned one of 37 predefined lithology labels (s ee Appendix~\ref{app:litho_detail}), such as banded iron formation (BIF). (3)~\textit{Geochemistry} records the concentrations of 25 elements at each interval. Because measurements are incomplete across depths, each geochemical vector is accompanied by a binary mask ($m_t \in \{0,1\}^{25}$) indicating the available elements.

\subsection{DrillBench Splits}
\label{subsec:splits}

\begin{wraptable}{r}{0.5\textwidth}
\centering
\vspace{-1em}
\caption{DrillBench split summary across splits.}

\label{tab:drillbench_splits}
\scriptsize
\setlength{\tabcolsep}{3.5pt}
\renewcommand{\arraystretch}{1.15}
\begin{adjustbox}{width=\linewidth}
\begin{tabular}{llrrrrr}
\toprule
\textbf{Split} &
\textbf{Region} &
\multicolumn{1}{c}{\textbf{Holes}} &
\multicolumn{1}{c}{\textbf{Intervals}} &
\multicolumn{1}{c}{\textbf{1m tokens}} &
\multicolumn{1}{c}{\textbf{Depth}} &
\multicolumn{1}{c}{\textbf{Lith.}} \\
&
&
\multicolumn{1}{c}{$n$} &
\multicolumn{1}{c}{$n$} &
\multicolumn{1}{c}{$n$} &
\multicolumn{1}{c}{$\bar{d}$ (m)} &
\multicolumn{1}{c}{(\%)} \\
\midrule
\rowcolor{black!4}
\multicolumn{7}{l}{\textit{In-distribution: Eastern Goldfields (EG) core}} \\
\textsc{train} & EG
& 27{,}435 & 818{,}350 & 2{,}076{,}335 & 75.9 & 70.1 \\
\textsc{val} & EG
& 1{,}951 & 60{,}998 & 152{,}817 & 78.5 & 69.4 \\
\textsc{test} & EG
& 3{,}856 & 120{,}196 & 301{,}531 & 78.3 & 70.0 \\
\addlinespace[2pt]
\rowcolor{black!4}
\multicolumn{7}{l}{\textit{Graded spatial shift: EG outer rings}} \\
\textsc{ood-regional} & EG
& 8{,}874 & 201{,}110 & 743{,}314 & 83.7 & 62.7 \\
\textsc{ood-distal} & EG
& 4{,}277 & 118{,}908 & 251{,}170 & 58.8 & 59.2 \\
\addlinespace[2pt]
\rowcolor{black!4}
\multicolumn{7}{l}{\textit{Cross-province transfer: Pilbara}} \\
\textsc{pilbara-test} & Pilbara
& 3{,}278 & 179{,}619 & 239{,}708 & 73.6 & 70.5 \\
\midrule
\rowcolor{black!2}
\textbf{Total} &
&
\textbf{49{,}671} &
\textbf{1{,}499{,}181} &
\textbf{3{,}764{,}875} &
\textbf{76.0} &
\textbf{68.3} \\
\bottomrule
\end{tabular}
\end{adjustbox}
\end{wraptable}

Each drillhole in DrillBench is assigned to one of six subsets (Table~\ref{tab:drillbench_splits}), defining three evaluation settings with progressively stronger distribution shifts: \textit{i) \textbf{In-distribution}}, using spatially clustered Eastern Goldfields (EG) drillholes for training, validation, and testing; \textit{ii) \textbf{Graded spatial shift}}, using two groups of EG drillholes increasingly separated from the training data (with same geological province but different spatial distance), with \textsc{ood-regional} located 15--98km from the nearest \textsc{train} collar (median 58km) and \textsc{ood-distal} 86--205km away (median 129km); and \textit{iii) \textbf{Cross-province transfer}}, using \textsc{pilbara-test} to evaluate generalisation to a geologically distinct province. Together, these settings form a controlled transfer spectrum from local prediction, through within-province spatial separation, to cross-province geological generalisation (Figure~\ref{fig:dataset_analysis}).

\textbf{WA-DASC-full.} Alongside DrillBench, we release WA-DASC-full, containing 498{,}713 drillholes across 30 DASC tectonic-province labels, 13 of which have more than 5{,}000 holes. About 112k holes with at least two logged lithology layers are used for pretraining. Unlike DrillBench, WA-DASC-full is only lightly standardised and is used for pretraining and retrieval rather than benchmark evaluation. To prevent leakage, all DrillBench evaluation holes, and their archive twins within 500\,m, are excluded from pretraining, fine-tuning, and retrieval. 

\section{A Transfer-Safe Training-and-Inference Recipe}
\label{sec:drillmar}

We introduce a backbone-agnostic training-and-inference recipe, which aims to bring in broader geological knowledge and local evidence without providing absolute coordinates and assay values out of the prediction input. The recipe has three components: a \textit{layer-level autoregressive} backbone over strata and thicknesses (\S\ref{subsec:drillmar-ar}), large-scale \textit{pretraining on the uncurated} WA-DASC-full archive with geochemistry as auxiliary supervision (\S\ref{subsec:drillmar-pretrain}), and \textit{spatial retrieval} that uses coordinates only to look up neighbouring lithology (\S\ref{subsec:retrieval}). \textit{Fine-tuning, decoding, and inference} follow in \S\ref{subsec:drillmar-finetune}.

\begin{figure*}[t]
\centering
\includegraphics[width=\linewidth]{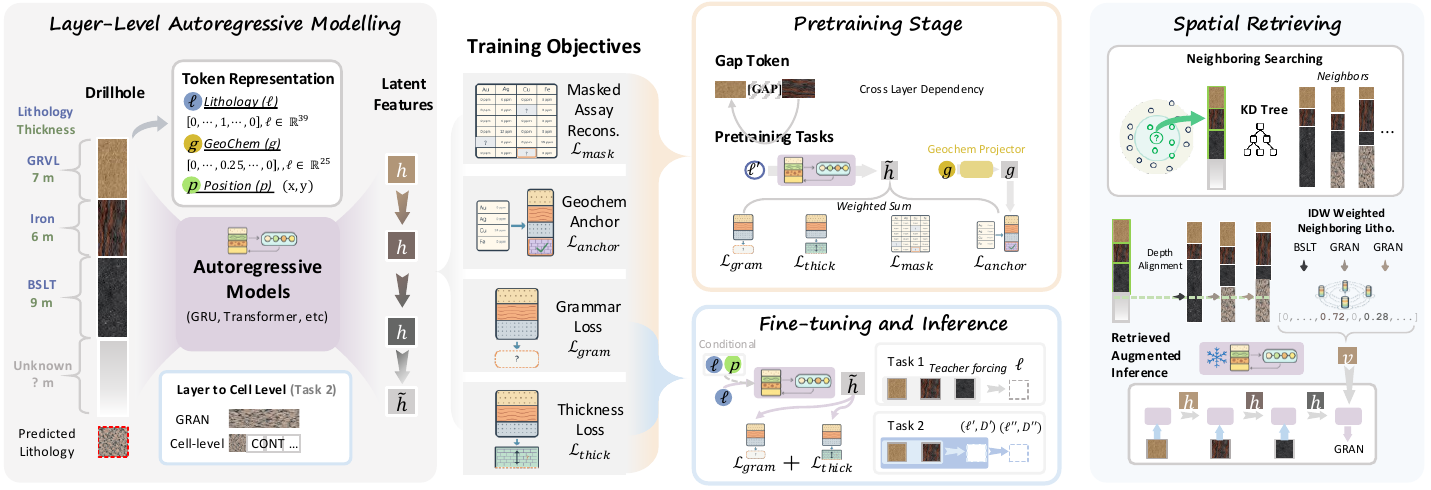}
\caption{A Transfer-Safe Recipe. Pretraining, fine-tuning, and retrieving augmented pipeline.}
\label{fig:drillmar}
\vspace{-2em}
\end{figure*}
\subsection{Layer-Level Autoregressive Model}
\label{subsec:drillmar-ar}
To directly model the dependencies between successive lithological strata, we represent each drillhole as a sequence of layers, $(\ell_1,D_1),\dots,(\ell_L,D_L)$, where $\ell_i$ denotes the lithology of layer $i$ and $D_i>0$ is its thickness. It follows a layer-level autoregressive architecture: given the first $i$ layers, it predicts the lithology and thickness of layer $i{+}1$. A sequence backbone $f_\theta$ maps the embedded lithologies to $h_i=f_\theta(\mathbf{e}(\ell_1),\dots,\mathbf{e}(\ell_i))$, where $\mathbf{e}$ is a learned lithology embedding.
The hidden state is first passed to a \textbf{lithology head}, which predicts a categorical distribution $q_i(k)$ over the lithology classes of the next layer. In addition, a separate \textbf{thickness head} predicts the corresponding layer thickness using a log-normal distribution. The predicted lithology and thickness together form the next layer $(\ell_{i+1},D_{i+1})$, which is appended to the generated sequence and the process repeats autoregressively until the drillhole is completed.

\subsection{Pretraining on Uncurated Archives}
\label{subsec:drillmar-pretrain}
 Large-scale pretraining allows model to learn lithological dependencies that generalise beyond individual regions. Geological sequences contain recurring patterns across regions, such as weathering profiles from laterite through saprolite to fresher bedrock. 
 Pretraining on a much larger archive can therefore provide more robust priors for cross-region generalisation. However, WA-DASC-full is highly heterogeneous, with only ${\sim}15\%$ of intervals mapping to the target lithology vocabulary. We address this heterogeneity through gap tokenisation and multi-task supervision.

\textbf{Gap tokenisation.}
To preserve stratigraphic continuity when lithology records are incomplete, we introduce an explicit \textsc{gap} token for unlogged intervals rather than breaking a drillhole into disconnected fragments. For example, $A \rightarrow [\textsc{gap}] \rightarrow B$ preserves the relative ordering of the observed strata while marking the missing section. This allows models to learn from longer layer sequences without treating missing records as observed lithologies.


\textbf{Multi-task pretraining.}
To learn transferable geological representations from the large-scale archive, we introduce four training objectives. The two primary tasks predict the next lithology and its thickness through $\mathcal{L}_{\mathrm{gram}}$ and $\mathcal{L}_{\mathrm{thick}}$. We further use geochemical assays as auxiliary supervision where available: $\mathcal{L}_{\mathrm{mask}}$ reconstructs randomly masked element values to capture compositional relationships, while $\mathcal{L}_{\mathrm{anchor}}$ predicts the corresponding lithology from chemistry when a mapped label is available. These auxiliary tasks allow assayed intervals, including those with unreliable or unmappable lithology descriptions, to provide additional training signals. Geochemistry \textit{is used only during pretraining} and removed during fine-tuning and inference, so the downstream model does not depend on assay availability in unseen regions. The overall objective is
\begin{equation}
\mathcal{L}_{\mathrm{pre}}
= \mathcal{L}_{\mathrm{gram}}
+ \lambda_{\mathrm{thick}}\mathcal{L}_{\mathrm{thick}}
+ \lambda_{\mathrm{mask}}\mathcal{L}_{\mathrm{mask}}
+ \lambda_{\mathrm{anchor}}\mathcal{L}_{\mathrm{anchor}},
\end{equation}
where $\lambda$ balances the pretraining objectives.

\subsection{Spatial retrieval without explicit coordinate inputs}
\label{subsec:retrieval}
We use neighbouring drillholes as local geological evidence without exposing absolute coordinates to the model. Coordinates are used only to retrieve the $K{=}10$ nearest logged holes in the same region from WA-DASC-full and DrillBench \textsc{train}. Evaluation holes are excluded from the retrieval pool, and neighbours within $25\,\mathrm{m}$ of the query collar are skipped to reduce the risk of retrieving duplicate holes.
To predict layer $i{+}1$, each retrieved neighbour is queried at the corresponding elevation when valid elevation information is available; otherwise, the same down-hole depth is used. A neighbour contributes a vote only if one of its beds starts within the corresponding $1\,\mathrm{m}$ depth cell. Its lithology is mapped to the 37-class vocabulary, and votes are combined using inverse-distance weighting:
\begin{equation}
r_i=\sum_{j\in \mathcal{N}_i}\alpha_{ij}\,\mathbf{1}(c_{ij}),
\qquad
\alpha_{ij}=\frac{1/d_j}{\sum_{m\in \mathcal{N}_i}1/d_m},
\end{equation}
where $\mathcal{N}_i$ is the set of voting neighbours, $\mathbf{1}(c_{ij})$ is the one-hot lithology vector, and $d_j$ is the collar distance. Thus, $r_i$ captures local lithological context without directly encoding absolute position.
We fuse the retrieval signal with the current layer representation as
$
x_i=\mathbf{e}(\ell_i)+\phi(r_i),
$
where $\phi$ is a small MLP. If no neighbour provides a valid vote, we set $r_i=\mathbf{0}$. Model parameters are fixed across all test sets, so at test time target-region information enters only through retrieved neighbours.

\subsection{Fine-Tuning, Decoding, and Evaluation}
\label{subsec:drillmar-finetune}

\textbf{Fine-tuning.}
The model $f_\theta$ is fine-tuned on \textsc{train} for the 37 lithology classes and thickness (chemistry pathway removed). The retrieval encoder $\phi$, absent from pretraining, is learned jointly under label smoothing ($\varepsilon=0.3$).
The layer input is
$
x_i=\mathbf e(\ell_i)+\phi(r_i),
$
and $h_i=f_\theta(x_1,\ldots,x_i)$. Retrieval is zeroed for 30\% of training holes, so the model can fall back to the sequence alone. The heads give
\begin{equation}
q_i(k)=P(\ell_{i+1}=k\mid h_i),\qquad
p(D_{i+1}\mid h_i,\ell_{i+1}),
\end{equation}
where $p(D)$ is a discretised log-normal over thickness in cells.


\textbf{Decoding and evaluation.}
Every baseline answers the same question on the 1\,m grid: whether the next cell is \texttt{CONT}, meaning that the current stratum continues, or a lithology $k$, meaning that a new stratum begins. Our model instead predicts the next stratum's lithology distribution $q$ and thickness distribution. We convert these predictions to the same cell-level form. If $\lambda$ is the probability that the current stratum ends at the present cell, then
\begin{equation}
P(\texttt{CONT})=1-s\lambda,\qquad P(k)=s\lambda\,q(k),
\end{equation}
where $s$ is a single calibration factor for the predicted boundary probability. A larger $s$ produces more stratum boundaries, while a smaller $s$ produces fewer. From this point, all methods have the same output space and are evaluated by the same code.
In \emph{Task~1}, each method is given the true history above a stratum boundary and is scored by whether its most probable lithology matches the stratum that actually begins there. In \emph{Task~2}, each method is given the true upper half of a hole and generates the remainder autoregressively until the logged depth. The generated cells are then grouped into strata and compared with the true lower half using their joint lithology--thickness distribution.
For each generator, $s$ is fitted only on \textsc{train} holes to match the observed boundary rate, and is then kept fixed for all test sets (Appendix~\ref{app:evaluation}).




\section{Results and Discussion}
\label{sec:results}

\providecolor{unitgray}{HTML}{E5E7EB}

\begin{wraptable}{r}{0.45\textwidth}
\vspace{-1.3em}
\centering
\footnotesize
\setlength{\tabcolsep}{3.2pt}
\renewcommand{\arraystretch}{1.05}

\caption{\textbf{Tasks~1 and 2 performance.}
Task~1 reports macro-F1 (\%, $\uparrow$);
Task~2 reports joint KL ($\downarrow$).
L: lithology; S: location; R: retrieval.}
\vspace{-0.5em}
\label{tab:main-tasks}

\begin{adjustbox}{width=\linewidth}
\begin{tabular}{@{}ll ccc rrrr@{}}
\toprule
& & \multicolumn{3}{c}{\textbf{Inputs}}
& \multicolumn{4}{c}{\textbf{Increasing spatial shift $\rightarrow$}} \\
\cmidrule(lr){3-5}
\cmidrule(lr){6-9}

\textbf{Family} & \textbf{Method}
& L & S & R
& Test & OOD-R & OOD-D & Pil-T \\

\midrule
\rowcolor{unitgray}
\multicolumn{9}{@{}l}{\textbf{Task~1: next-layer macro-F1 (\%, $\uparrow$)}} \\

\rowcolor{unitgray}
\multicolumn{9}{@{}l}{\textbf{Cell level} (1 meter token)} \\

\rowcolor{black!2}
& Unigram
& & & &
2.0 & 0.8 & 1.1 & 1.7 \\

\rowcolor{black!2}
\multirow{-2}{*}{\textbf{Freq.}}
& $3$-gram
& \cmark & & &
9.4 & 3.2 & 4.0 & 3.7 \\

\addlinespace[1.5pt]

\rowcolor{black!2}
& TPROGS
& \cmark & \cmark & &
4.5 & 1.1 & 1.2 & 1.7 \\

\rowcolor{black!2}
\multirow{-2}{*}{\textbf{Geostat.}}
& Spatial $k$-NN
& & & \cmark &
68.9 & 11.5 & 9.6 & 8.0 \\

\addlinespace[1.5pt]

\rowcolor{black!2}
\textbf{ML}
& XGBoost
& \cmark & \cmark & &
64.8 & 2.4 & 2.5 & 2.2 \\

\addlinespace[1.5pt]

\rowcolor{blue!3}
& GRU
& \cmark & & &
58.8 & 21.4 & 23.0 & 23.2 \\

\rowcolor{blue!3}
\multirow{-2}{*}{\textbf{DL}}
& Transformer
& \cmark & & &
60.3 & 26.5 & 23.6 & 22.5 \\

\midrule
\rowcolor{unitgray}
\multicolumn{9}{@{}l}{\textbf{Layer level} (stratum token)} \\

\rowcolor{black!2}
\textbf{Markov}
& Layer Markov
& \cmark & & &
56.7 & 36.3 & 26.1 & 22.0 \\

\addlinespace[1.5pt]

\rowcolor{blue!3}
& GRU
& \cmark & & &
59.4 & 37.4 & 28.3 & 28.2 \\

\rowcolor{blue!3}
\multirow{-2}{*}{\textbf{DL}}
& Transformer
& \cmark & & &
58.9 & 36.6 & 29.1 & 26.0 \\

\addlinespace[1.5pt]

\rowcolor{proposedbg}
& GRU + R
& \cmark & & \cmark &
\best{70.0} & 37.4 & 33.9 & 26.9 \\

\rowcolor{proposedbg}
& GRU + PT
& \cmark & & &
58.5 & \best{39.2} & 33.0 & \best{31.7} \\

\rowcolor{proposedbg}
\multirow{-3}{*}{\textbf{Ours}}
& GRU + PT + R
& \cmark & & \cmark &
69.2 & 38.8 & \best{37.2} & 30.6 \\

\midrule
\rowcolor{unitgray}
\multicolumn{9}{@{}l}{\textbf{Task~2: joint KL ($\downarrow$)}} \\

\rowcolor{unitgray}
\multicolumn{9}{@{}l}{\textbf{Cell level} (1 meter token)} \\

\rowcolor{black!2}
& Unigram
& & & &
0.226 & 0.599 & 0.872 & 1.111 \\

\rowcolor{black!2}
\multirow{-2}{*}{\textbf{Freq.}}
& $3$-gram
& \cmark & & &
0.186 & 0.533 & 0.781 & 0.960 \\

\addlinespace[1.5pt]

\rowcolor{black!2}
& TPROGS
& \cmark & \cmark & &
0.183 & 0.492 & 0.716 & 0.885 \\

\rowcolor{black!2}
\multirow{-2}{*}{\textbf{Geostat.}}
& Spatial $k$-NN$^{*}$
& & & \cmark &
0.553 & 0.950 & 0.852 & 0.729 \\

\addlinespace[1.5pt]

\rowcolor{black!2}
\textbf{ML}
& XGBoost
& \cmark & \cmark & &
0.206 & 0.518 & 0.950 & 1.776 \\

\addlinespace[1.5pt]

\rowcolor{blue!3}
& GRU
& \cmark & & &
0.340 & 0.612 & 0.559 & 1.003 \\

\rowcolor{blue!3}
\multirow{-2}{*}{\textbf{DL}}
& Transformer
& \cmark & & &
0.356 & 0.585 & 0.533 & 0.995 \\

\midrule
\rowcolor{unitgray}
\multicolumn{9}{@{}l}{\textbf{Layer level} (stratum token)} \\

\rowcolor{black!2}
\textbf{Markov}
& Layer Markov
& \cmark & & &
0.195 & 0.385 & 0.509 & 0.849 \\

\addlinespace[1.5pt]

\rowcolor{blue!3}
& GRU
& \cmark & & &
0.183 & 0.356 & 0.464 & 0.550 \\

\rowcolor{blue!3}
\multirow{-2}{*}{\textbf{DL}}
& Transformer
& \cmark & & &
\best{0.172} & 0.320 & 0.495 & 0.544 \\

\addlinespace[1.5pt]

\rowcolor{proposedbg}
& GRU + R
& \cmark & & \cmark &
0.192 & 0.341 & 0.479 & 0.561 \\

\rowcolor{proposedbg}
& GRU + PT
& \cmark & & &
0.185 & \best{0.303} & 0.488 & 0.551 \\

\rowcolor{proposedbg}
\multirow{-3}{*}{\textbf{Ours}}
& GRU + PT + R
& \cmark & & \cmark &
0.186 & 0.322 & \best{0.408} & \best{0.458} \\

\bottomrule
\end{tabular}
\end{adjustbox}

\vspace{-3.5em}
\end{wraptable}

We present the results in three stages. We first benchmark autoregressive models against classical, geostatistical, and learning-based alternatives on next-layer prediction and free-running generation. We then study how performance changes with increasing spatial and geological shift, using this analysis to expose the \emph{transfer boundary} and assess how pretraining and retrieval mitigate it. Finally, case studies and error analysis highlight the remaining limitations and point to future directions. Additional results and analyses are reported in the Appendix~\ref{app:results}.

\subsection{Benchmark Performance}

\textbf{Task~1: Next-layer prediction.}
Frequency-based methods remain weak across all settings. Spatial $k$-NN and XGBoost perform well on the in-distribution test set, but their performance drops sharply on more distant regions. In contrast, autoregressive neural models degrade more gradually and maintain much stronger performance under spatial shift.
Cell- and layer-level neural baselines perform similarly in-distribution, but layer-level models are generally stronger under spatial shift. The benefit becomes clearer when pretraining and retrieval are added. Retrieval gives the largest improvement in-distribution, with GRU+R reaching 70.0\% macro-F1. Pretraining is more helpful for transfer, giving the best OOD-R and Pilbara-test results (39.2\% and 31.7\%). Combining pretraining and retrieval achieves the best OOD-D result at 37.2\% and remains competitive in the other settings. Overall, these results show that retrieval mainly strengthens local prediction, while pretraining improves transfer to more distant regions. We therefore focus the following analysis on layer-level models.

\textbf{Task~2: Free-running generation.}
This task is more challenging because errors accumulate over the rollout. Frequency, geostatistical, and ML methods are competitive in some settings but degrade under larger spatial shifts. Neural autoregressive models are more stable, with layer-level models generally outperforming cell-level models on the harder transfer splits.
At the layer level, retrieval and pretraining provide complementary gains. GRU+PT+R achieves the best OOD-D and Pilbara-test results, with joint KL of 0.408 and 0.458, while GRU+PT is strongest among our models on OOD-R. Together with Task~1, these results support modelling lithology directly at the stratum level for more transferable autoregressive generation.

\subsection{transfer boundary and Mitigation Strategies - Our Transfer Safe Recipe}
\begin{figure*}[h]
\centering
\vspace{-1em}

\includegraphics[width=\textwidth]{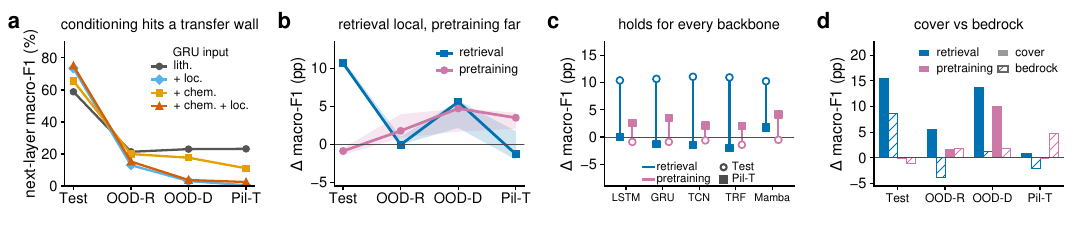}
\vspace{-2.5em}

\caption{
\textbf{What transfers.} Next-layer prediction performance under increasing shift.
\textbf{(a)} Direct conditioning and the transfer boundary (lith only).
\textbf{(b)} Local retrieval vs.\ broad pretraining.
\textbf{(c)} Consistency across five backbones.
\textbf{(d)} Regolith cover vs.\ bedrock.
}
\vspace{-1em}
\label{fig:transfer}
\end{figure*}

\textbf{i) Direct conditioning causes the transfer boundary.}
Figure~\ref{fig:transfer}(a) shows a clear trade-off between local accuracy and transfer. Adding location and chemistry improves \textsc{test} performance, but these gains collapse under distribution shift: the fully conditioned model drops from 75\% on \textsc{test} to below 3\% on \textsc{pilbara-test}, while the lithology-only model remains above 20\% on all shifted splits. We call this reversal the \emph{transfer boundary}: signals that are highly predictive locally can become harmful when their regional associations no longer hold. This is consistent with known limitations of coordinates as spatial predictors~\citep{meyer2019importance,ploton2020spatial}; our receip therefore avoids using raw coordinates and assays as model inputs.
\textbf{ii) Pretraining supports broad geological transfer.}
Figure~\ref{fig:transfer}(b) shows that pretraining provides little gain in-distribution, but becomes positive once the data shift and stays positive on every shifted split, indicating that it mainly contributes transferable geological knowledge rather than local fit. This trend is consistent across multiple deep-learning backbones (Figure~\ref{fig:transfer}(c)): pretraining has almost no effect on \textsc{test}, but consistently improves \textsc{pilbara-test} performance.
\textbf{iii) Retrieval supports local spatial adaptation.}
Retrieval shows the opposite pattern. It provides the largest gains in distribution, where nearby drillholes supply directly relevant geological evidence, and its benefit is smallest across provinces (Figure~\ref{fig:transfer}(b)). The same behaviour appears across multiple backbones (Figure~\ref{fig:transfer}(c)). This suggests that retrieval is most useful for local adaptation rather than broad geological transfer.
\textbf{iv) Pretraining and retrieval are complementary across geological regimes.} Figure~\ref{fig:transfer}(d) separates their effects between regolith cover and bedrock. Cover comprises weathered or transported material above the bedrock and is strongly shaped by local regolith--landform processes, whereas bedrock more directly records the geological units relevant to primary mineralisation~\citep{anand2010guide}. Consistent with this distinction, retrieval is most effective in Eastern Goldfields cover, where nearby holes provide useful local context, while pretraining is more valuable in bedrock on OOD-R and across provinces. OOD-D is the exception because most scored layers are regolith (Table~\ref{tab:app-task1-zones}). Overall, retrieval captures local near-surface continuity, whereas pretraining better supports transfer of deeper geological structure.

\subsection{Failure Modes and Case Studies.}

\begin{figure}[t]
\centering
\includegraphics[width=\textwidth]{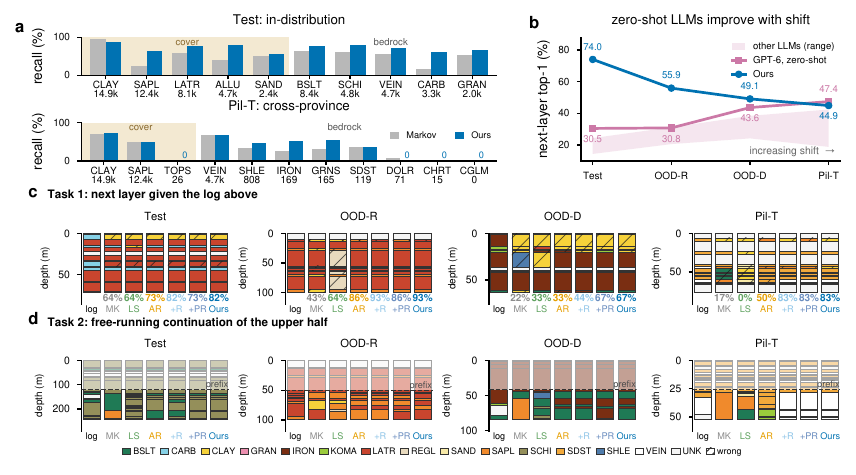}
\vspace{-2em}
\caption{\textbf{Failure analysis under geological shift.}
(a) Cross-province errors concentrate on lithologies with limited training support. (b) Without retrieval, zero-shot LLMs degrade less under shift (top-1 accuracy on 150 randomly selected holes per split; our model is scored on the same holes). (c--d) Case studies show the complementary effects of retrieval and pretraining, together with remaining lithology and boundary errors.}

\label{fig:failure-cases}
\vspace{-2em}
\end{figure}

\textbf{Failures Concentrate on Poorly Supported Lithologies.}
Figure~\ref{fig:failure-cases}(a) compares per-class recall in-distribution and on \textsc{pilbara-test}. In-distribution, the model performs well on most common cover and bedrock classes. Under cross-province transfer, however, errors are concentrated in lithologies with little source-region support. Well-represented classes such as CLAY, SAPL, and VEIN remain reasonably predictable, whereas every zero-recall class on \textsc{pilbara-test} has at most 71 training layers. This indicates a clear class-support shift: stratigraphic patterns may transfer, but lithologies that are rarely observed in the source region remain difficult to recover from sequence context alone. Future gains may therefore come more from larger and geologically broader training corpora, or limited target-region supervision, than from simply increasing model capacity.

\textbf{LLMs Perform Better OOD than In-Domain.} We additionally evaluate several recent LLMs under zero-shot prompting, with and without retrieved geological context. Surprisingly, without retrieved context their performance improves with increasing shift: GPT-6 rises from 30.5\% to 47.4\% under cross-province shift, slightly above our model (44.9\%), while our trained model shows the opposite trend. This suggests that LLMs may draw on broad geological knowledge that is less tied to the source region, although part of it may reflect exposure to the public DASC records during pretraining (Appendix~\ref{app:llm}). Although weaker than specialised models in-distribution, they may be particularly useful when target-region labels are scarce. A promising direction is to combine LLMs with our retrieval and autoregressive components in an agentic system, using the LLM for broad geological reasoning and the specialised model for local prediction.

\textbf{Case Studies Reveal Complementary Strengths and Remaining Failures.}
Figure~\ref{fig:failure-cases}(c--d) shows how these trends appear at the drillhole level. The autoregressive model already captures major stratigraphic transitions better than Markov and cell-level baselines, while retrieval helps when nearby holes provide relevant local structure and pretraining becomes more useful as the geological shift increases. Combining both gives the most consistent predictions across the four spatial settings, including substantial gains on the \textsc{OOD-D} and \textsc{pilbara-test} examples. The free-running cases show similar progress, but also expose remaining errors in lithology identity and layer boundaries as predictions accumulate. Together, these examples reinforce the main result: autoregressive modelling provides the transferable foundation, retrieval contributes local evidence, and pretraining supplies broader geological priors. Further gains are likely to require broader geological coverage and better use of sparse target-region evidence (e.g. LLMs), rather than simply larger backbones.

\section{Conclusion}
We study autoregressive modelling for mineral-exploration drillholes, a domain where predictions must remain useful across strong spatial and geological shifts. We introduce DrillBench to evaluate next-layer prediction and free-running generation, and show that location- and assay-based conditioning can create a clear transfer boundary. To address this domain-specific challenge, we develop a transfer-safe recipe that combines layer-level autoregression, large-scale geological pretraining, and spatial retrieval: retrieval exploits local continuity, while pretraining supports broader geological transfer. Our results show that effective subsurface modelling requires not only stronger sequence models, but mechanisms designed around geological structure, data sparsity, and regional shift. 

\subsection*{AI use statement}

Generative AI tools were used to assist with language editing, translation, manuscript organisation, literature search, and the refinement of methodological descriptions. They were also used to discuss experimental design and interpretation of results. All AI-assisted suggestions were critically reviewed by the authors and checked against the underlying data, experiments, code, and cited literature. The authors made all final methodological and scientific decisions and take full responsibility for the content, results, and claims of this work.

\subsection*{Ethics statement} This work uses only public mineral-exploration records from the DEMIRS DASC archive (CC-BY-4.0), which contain no personal data and involve no human subjects. We foresee no ethical concerns beyond standard caution in using model predictions for exploration decisions.

\subsection*{Reproducibility statement} All implementation details needed to reproduce our results are provided in the appendix, including the full data curation pipeline and lithology vocabulary (Appendices~\ref{app:dataset_curation} and~\ref{app:litho_detail}) together with the split construction (\S\ref{subsec:splits}), architectures, hyperparameters, and training schedules for every baseline and for our model (Appendix~\ref{app:baselines}), the evaluation protocol and metrics (Appendix~\ref{app:evaluation}), and the exact prompts and inference settings for the LLM baselines (Appendix~\ref{app:llm}). We further report extensive ablations on every component of the recipe (Appendix~\ref{app:ptrk}; all settings are summarised in Table~\ref{tab:app-hparams}), together with seeds and hole-cluster bootstrap confidence intervals (Appendices~\ref{app:results_task1} and~\ref{app:results_task2}). The DrillBench dataset, the WA-DASC-full pretraining and retrieving corpus, all code, pretrained checkpoints, and raw evaluation outputs will be publicly released upon acceptance.



\bibliography{iclr2027_conference}

@article{meyer2019importance,
  title={Importance of spatial predictor variable selection in machine learning applications--Moving from data reproduction to spatial prediction},
  author={Meyer, Hanna and Reudenbach, Christoph and W{\"o}llauer, Stephan and Nauss, Thomas},
  journal={Ecological Modelling},
  volume={411},
  pages={108815},
  year={2019},
  publisher={Elsevier}
}

@article{ploton2020spatial,
  title={Spatial validation reveals poor predictive performance of large-scale ecological mapping models},
  author={Ploton, Pierre and Mortier, Fr{\'e}d{\'e}ric and R{\'e}jou-M{\'e}chain, Maxime and Barbier, Nicolas and Picard, Nicolas and Rossi, Vivien and Dormann, Carsten and Cornu, Guillaume and Viennois, Ga{\"e}lle and Bayol, Nicolas and others},
  journal={Nature communications},
  volume={11},
  number={1},
  pages={4540},
  year={2020},
  publisher={Nature Publishing Group UK London}
}

@inproceedings{holtzman2021surface,
  title={Surface form competition: Why the highest probability answer isn’t always right},
  author={Holtzman, Ari and West, Peter and Shwartz, Vered and Choi, Yejin and Zettlemoyer, Luke},
  booktitle={Proceedings of the 2021 conference on empirical methods in natural language processing},
  pages={7038--7051},
  year={2021}
}

@article{perron2026universat,
  title={UniverSat: Resolution-and Modality-Agnostic Transformers for Earth Observation},
  author={Perron, Yohann and Astruc, Guillaume and Gonthier, Nicolas and Mallet, Clement and Landrieu, Loic},
  journal={arXiv preprint arXiv:2606.23503},
  year={2026}
}

@inproceedings{danish2026terrafm,
  title={TerraFM: A scalable foundation model for unified multisensor earth observation},
  author={Danish, Muhammad Sohail and Munir, Muhammad Akhtar and Shah, Syed and Khan, Muhammad Haris and Anwer, Rao and Laaksonen, Jorma and Khan, Fahad and Khan, Salman},
  booktitle={International Conference on Learning Representations},
  volume={2026},
  pages={12787--12809},
  year={2026}
}

@article{xiao2026geommbench,
  title={Geommbench and geommagent: Toward expert-level multimodal intelligence in geoscience and remote sensing},
  author={Xiao, Aoran and Cheng, Shihao and Xu, Yonghao and Ren, Yexian and Chen, Hongruixuan and Yokoya, Naoto},
  journal={arXiv preprint arXiv:2604.08896},
  year={2026}
}

@article{li2026reobench,
  title={Reobench: Benchmarking robustness of earth observation foundation models},
  author={Li, Xiang and Tao, Yong and Zhang, Siyuan and Liu, Siwei and Xiong, Zhitong and Luo, Chunbo and Liu, Lu and Pechenizkiy, Mykola and Zhu, Xiaoxiang and Huang, Tianjin},
  journal={Advances in Neural Information Processing Systems},
  volume={38},
  year={2026}
}

@inproceedings{ji2026cf,
  title={CF-IPT: Cross-Modal Fusion Interactive Prompt Tuning of Vision-Language Pre-Trained Model for Multisource Remote Sensing Data Classification},
  author={Ji, Jinheng and Qu, Jiahui and Dong, Wenqian and Li, Yunsong},
  booktitle={Proceedings of the IEEE/CVF Conference on Computer Vision and Pattern Recognition},
  pages={23021--23030},
  year={2026}
}

@article{bai2026geolink,
  title={Geolink: Empowering remote sensing foundation model with openstreetmap data},
  author={Bai, Lubin and Zhang, Xiuyuan and Zhang, Siqi and Zhang, Zepeng and Wang, Haoyu and Qin, Wei and Du, Shihong},
  journal={Advances in Neural Information Processing Systems},
  volume={38},
  pages={168557--168587},
  year={2026}
}

@article{anand2010guide,
  title={A guide for mineral exploration through the regolith in the Yilgarn Craton, Western Australia},
  author={Anand, RR and Butt, CRM},
  journal={Australian Journal of Earth Sciences},
  volume={57},
  number={8},
  pages={1015--1114},
  year={2010},
  publisher={Taylor \& Francis}
}

@String{Computing = "Computing" }

@String{Computer = "{IEEE} Computer" }

@String{Springer = "Springer-Verlag" }

@article{achiam2023gpt,
  title={Gpt-4 technical report},
  author={Achiam, Josh and Adler, Steven and Agarwal, Sandhini and Ahmad, Lama and Akkaya, Ilge and Aleman, Florencia Leoni and Almeida, Diogo and Altenschmidt, Janko and Altman, Sam and Anadkat, Shyamal and others},
  journal={arXiv preprint arXiv:2303.08774},
  year={2023}
}

@article{carle1996transition,
  title={Transition probability-based indicator geostatistics},
  author={Carle, Steven F and Fogg, Graham E},
  journal={Mathematical geology},
  volume={28},
  number={4},
  pages={453--476},
  year={1996},
  publisher={Springer}
}

@article{li2026generalized,
  title={A Generalized Borehole Autoregressive Neural Network for 3D Geological Modeling},
  author={Li, Hao and Zhang, Chi and He, Zhenwen},
  journal={Algorithms},
  volume={19},
  number={2},
  pages={128},
  year={2026},
  publisher={MDPI}
}

@article{brown2020language,
  title={Language models are few-shot learners},
  author={Brown, Tom and Mann, Benjamin and Ryder, Nick and Subbiah, Melanie and Kaplan, Jared D and Dhariwal, Prafulla and Neelakantan, Arvind and Shyam, Pranav and Sastry, Girish and Askell, Amanda and others},
  journal={Advances in neural information processing systems},
  volume={33},
  pages={1877--1901},
  year={2020}
}

@article{nguyen2024sequence,
  title={Sequence modeling and design from molecular to genome scale with evo. bioRxiv},
  author={Nguyen, Eric and Poli, Michael and Durrant, Matthew G and Thomas, Armin W and Kang, Brian and Sullivan, Jeremy and Ng, Madelena Y and Lewis, Ashley and Patel, Aman and Lou, Aaron and others},
  journal={Published online March},
  volume={6},
  pages={2024--02},
  year={2024}
}

@article{ren2025deep,
  title={A deep learning method for 3D geological modeling using ET4DD with offset-attention mechanism},
  author={Ren, Anjing and Wu, Liang and Xu, Jianglong and Xing, Yanjie and Qiu, Qinjun and Xie, Zhong},
  journal={Computers \& Geosciences},
  volume={200},
  pages={105929},
  year={2025},
  publisher={Elsevier}
}

@article{garza2023timegpt,
  title={TimeGPT-1},
  author={Garza, Azul and Challu, Cristian and Mergenthaler-Canseco, Max},
  journal={arXiv preprint arXiv:2310.03589},
  year={2023}
}

@article{madani2023large,
  title={Large language models generate functional protein sequences across diverse families},
  author={Madani, Ali and Krause, Ben and Greene, Eric R and Subramanian, Subu and Mohr, Benjamin P and Holton, James M and Olmos Jr, Jose Luis and Xiong, Caiming and Sun, Zachary Z and Socher, Richard and others},
  journal={Nature biotechnology},
  volume={41},
  number={8},
  pages={1099--1106},
  year={2023},
  publisher={Nature Publishing Group US New York}
}

@article{bormann2020force,
  title={FORCE 2020 Well well log and lithofacies dataset for machine learning competition},
  author={Bormann, Peter and Aursand, Peder and Dilib, Fahad and Manral, Surrender and Dischington, Peter},
  journal={Zenodo: Geneva, Switzerland},
  year={2020}
}

@article{sun2023cross,
  title={Cross-well lithology identification based on wavelet transform and adversarial learning},
  author={Sun, Longxiang and Li, Zerui and Li, Kun and Liu, Haining and Liu, Ge and Lv, Wenjun},
  journal={Energies},
  volume={16},
  number={3},
  pages={1475},
  year={2023},
  publisher={MDPI}
}

@article{hall2016facies,
  title={Facies classification using machine learning},
  author={Hall, Brendon},
  journal={The Leading Edge},
  volume={35},
  number={10},
  pages={906--909},
  year={2016},
  publisher={Society of Exploration Geophysicists}
}

@article{carvalho2026lithology,
  title={Lithology Classification Based on Well Log Data: A Benchmark for Machine Learning Models},
  author={Carvalho, Henrique MO and Daniel, Henrique and Korenchendler, Andr{\'e} and Sobreira, Matheus CA and Machado, Alexei MC},
  journal={Mathematical Geosciences},
  pages={1--23},
  year={2026},
  publisher={Springer}
}

@article{wedge2019machine,
  title={Machine learning assisted geological interpretation of drillhole data: Examples from the Pilbara Region, Western Australia},
  author={Wedge, Daniel and Hartley, Owen and McMickan, Andrew and Green, Thomas and Holden, Eun-Jung},
  journal={Ore Geology Reviews},
  volume={114},
  pages={103118},
  year={2019},
  publisher={Elsevier}
}

@article{ghorbanfekr2025classification,
  title={Classification of geological borehole descriptions using a domain adapted large language model},
  author={Ghorbanfekr, Hossein and Kerstens, Pieter Jan and Dirix, Katrijn},
  journal={Applied Computing and Geosciences},
  volume={25},
  pages={100229},
  year={2025},
  publisher={Elsevier}
}

@article{qi2025wlfm,
  title={WLFM: A Well-Logs Foundation Model for Multi-Task and Cross-Well Geological Interpretation},
  author={Qi, Zhenyu and Yu, Qing and Wang, Jichen and Zhao, Yun-Bo and Li, Zerui and Lv, Wenjun},
  journal={arXiv preprint arXiv:2509.18152},
  year={2025}
}

@article{liu2020well,
  title={Well logging based lithology identification model establishment under data drift: A transfer learning method},
  author={Liu, Haining and Wu, Yuping and Cao, Yingchang and Lv, Wenjun and Han, Hongwei and Li, Zerui and Chang, Ji},
  journal={Sensors},
  volume={20},
  number={13},
  pages={3643},
  year={2020},
  publisher={MDPI}
}

@book{caers2011modeling,
  title={Modeling uncertainty in the earth sciences},
  author={Caers, Jef},
  year={2011},
  publisher={John Wiley \& Sons}
}

@article{strebelle2002conditional,
  title={Conditional simulation of complex geological structures using multiple-point statistics},
  author={Strebelle, Sebastien},
  journal={Mathematical geology},
  volume={34},
  number={1},
  pages={1--21},
  year={2002},
  publisher={Springer}
}

@inproceedings{chen2016xgboost,
  title={Xgboost: A scalable tree boosting system},
  author={Chen, Tianqi and Guestrin, Carlos},
  booktitle={Proceedings of the 22nd acm sigkdd international conference on knowledge discovery and data mining},
  pages={785--794},
  year={2016}
}

@BOOK{test,
   author = "Donald E. Knuth",
   title = "Seminumerical Algorithms",
   volume = 2,
   series = "The Art of Computer Programming",
   publisher = "Addison-Wesley",
   address = "Reading, MA",
   edition = "2nd",
   month = "10~" # jan,
   year = "1981",
}

@ArtifactSoftware{R,
    title = {R: A Language and Environment for Statistical Computing},
    author = {{R Core Team}},
    organization = {R Foundation for Statistical Computing},
    address = {Vienna, Austria},
    year = {2019},
    url = {https://www.R-project.org/},
}
\bibliographystyle{iclr2027_conference}

\newpage
\appendix

\section{Scope and Limitations}\label{app:limitations}
The aim of this paper is to establish autoregressive next-stratum modelling as a paradigm for drillholes: we formulate the task, build a benchmark and protocol that test it under graded spatial and cross-province shift, and show with a transfer-safe recipe that sequence structure learned from historical archives transfers where location-based conditioning does not. Several scope choices serve this aim. We use company logs as they are: depths and thicknesses are measured along the hole without correcting for its orientation (dip and azimuth), so thicknesses are apparent rather than true, and lithology comes from a rule-based harmonisation of free-text descriptions (App.~\ref{app:dataset_curation}); the formulation must work on the records that exploration actually produces. We evaluate cross-province transfer on the Pilbara, which we harmonised to benchmark quality alongside the Eastern Goldfields; extending the benchmark to the other provinces of WA-DASC-full, where only about 15\% of intervals map to our vocabulary (\S\ref{subsec:drillmar-pretrain}), requires reconciling each company's logging vocabulary, a curation effort that we leave to future releases. We keep the models small (Table~\ref{tab:compute}) and their inputs to lithology (App.~\ref{app:ptrk-components}), so that the gains can be attributed to the formulation rather than to scale or to region-specific signals. Orientation-corrected depths, larger backbones, per-hole likelihood evaluation of generation (App.~\ref{app:results_task2}), and transferable encodings of location and chemistry are natural next steps, and the benchmark is built to support them.

\label{app:related_work}

\begin{table*}[t]
\centering
\caption{Comparison with representative studies on subsurface modelling.}
\label{tab:related}
\footnotesize
\setlength{\tabcolsep}{5pt}
\renewcommand{\arraystretch}{1.12}
\begin{adjustbox}{width=\textwidth}
\begin{tabular}{l l l l l l c}
\toprule
\textbf{Work} &
\textbf{Object} &
\textbf{Input} &
\textbf{Task formulation} &
\textbf{Evaluation setting} &
\textbf{Scale} &
\textbf{Public data} \\
\midrule

\rowcolor{groupgray}
\multicolumn{7}{l}{\textit{Petroleum well logs}} \\

Lithology benchmark~\citet{carvalho2026lithology}
& well log
& curves and facies
& point-wise classification
& within-basin CV
& 341 wells
& yes \\

SEG'16 contest~\citet{hall2016facies}
& well log
& curves and facies
& point-wise classification
& within-basin
& 10 wells
& yes \\

TL under data drift~\citet{liu2020well}
& well log
& continuous curves
& classification and adaptation
& labelled target wells
& 7 wells
& no \\

FORCE'20 contest~\citet{bormann2020force}
& well log
& curves and facies
& point-wise classification
& within-basin
& 118 wells
& yes \\

Adversarial cross-well~\citet{sun2023cross}
& well log
& continuous curves
& classification and adaptation
& labelled target wells
& ${\sim}$32k pts
& no \\

TimeGPT~\citet{garza2023timegpt}
& well log
& single curve
& zero-shot forecasting
& cross-basin
& 100B pts$^{\ddagger}$
& no \\

WLFM~\citet{qi2025wlfm}
& well log
& multiple curves
& masked pretraining
& cross-well, same basin
& 1.2k wells
& no \\
\rowcolor{groupgray}
\multicolumn{7}{l}{\textit{Mineral drillholes}} \\

ML-assisted logging~\citet{wedge2019machine}
& drillhole
& lithology, assays
& stratigraphic interpretation
& single province
& 396 holes
& no \\

ET4DD~\citet{ren2025deep}
& drillhole
& coordinates
& 3D point classification
& single area
& 719 holes
& no \\

GBARNN~\citet{li2026generalized}
& drillhole
& coordinates
& spatial autoregression
& single area
& 64 holes
& no \\

GEOBERTje~\citet{ghorbanfekr2025classification}
& drillhole text
& free text
& lithology classification
& not evaluated
& 283k desc.
& partial \\

\midrule

\rowcolor{proposedbg}
\textbf{Ours}
& \textbf{drillhole}
& \textbf{lithology}
& \textbf{causal prediction and generation}
& \textbf{cross-province zero-shot}
& \textbf{50k holes}
& \textbf{yes} \\

\bottomrule
\end{tabular}
\end{adjustbox}

\vspace{2pt}
\raggedright
\end{table*}
\section{Dataset Details}

\subsection{Dataset Curation Details}
\label{app:dataset_curation}

Table~\ref{tab:curation-funnel} reports how many holes and intervals each step removes. Steps are grouped by what they act on rather than listed in execution order. The vocabulary mapping (4) is computed before the quality filter (2), whose condition (iv) depends on it, and value-level masking (5) removes no rows, so the counts are unaffected by ordering.

\begin{table}[t]
\centering
\caption{Curation funnel. Steps 4--5 change values, not rows; step 7 moves holes out of the benchmark, it does not delete them.}
\label{tab:curation-funnel}
\small
\setlength{\tabcolsep}{4pt}
\begin{tabular}{@{}l l r r@{}}
\toprule
\textbf{Step} & \textbf{Removed} & \textbf{Holes} & \textbf{Intervals} \\
\midrule
(0) Logged holes in region & --         & 706{,}198   & 16{,}008{,}692 \\
(1) Corrupt records      & 26 records   & --          & -- \\
(2) Quality filter       & 655{,}085 holes & 51{,}113 & 1{,}533{,}265 \\
(3) No valid coordinates & 27 holes     & 51{,}086    & 1{,}533{,}004 \\
(4) Lithology mapping    & 0            & \multicolumn{2}{r}{68.2\% intervals mapped} \\
(5) Bounds masking       & 0            & \multicolumn{2}{r}{409{,}821 values nulled} \\
(6) Twin de-duplication (\textsc{val}, \textsc{test}) & 1{,}051 holes & 50{,}035 & 1{,}511{,}299 \\
(7) Pilbara context holes \(\to\) retrieval archive & 364 holes & \textbf{49{,}671} & \textbf{1{,}499{,}181} \\
\bottomrule
\end{tabular}
\end{table}

\paragraph{(1) Corrupt-record removal.}
The source is distributed as native BCP exports, and a small fraction of rows are byte-desynchronised. We detect these by values no drillhole can take: a maximum depth above \(10\,\mathrm{km}\), or a \texttt{hole\_id} longer than 500 characters. The 26 affected collar records (941{,}143 \(\to\) 941{,}117 in the two study regions) are dropped, not repaired, because the remaining fields of a desynchronised row cannot be trusted.

\paragraph{(2) Hole-level quality filter.}
Missing \texttt{to\_depth} values are first filled from the next interval's \texttt{from\_depth}. Intervals that still lack a base or have non-positive thickness are then discarded. A hole is admitted only if it satisfies all four conditions:
(i) maximum depth \(\ge 30\,\mathrm{m}\);
(ii) at least five logged intervals;
(iii) substantive lithology text on at least \(50\%\) of intervals;
(iv) at least \(30\%\) of intervals mapped to the vocabulary.
Conditions (i)--(ii) exclude shallow reconnaissance drilling that is too short to form a stratigraphic sequence and gives an autoregressive model insufficient context. For (iii), text is non-substantive if it is empty, purely numeric (e.g.\ recovery percentages entered in the lithology column), a single character, or a bare modifier such as \texttt{Fg}. Such text passes a naive non-null check but carries no lithology. Condition (iv) keeps logs that are mostly unreadable by the vocabulary out of the benchmark. The filter is the dominant reduction (Table~\ref{tab:curation-funnel}) and is not uniform across regions: Pilbara iron-ore exploration relies heavily on shallow rotary-air-blast drilling, so a larger share of its holes fails (i)--(ii).

\paragraph{(3) Coordinate harmonisation and unlocatable holes.}
Every DrillBench split is defined by spatial distance, so a hole without valid collar coordinates cannot be assigned to a split; we exclude 27 such holes (261 intervals), all in the Eastern Goldfields. For the remaining holes, the coordinates as reported are not mutually comparable: companies file eastings and northings in MGA zone 50 or 51 and in either GDA94 or the older AGD84 datum, whose grids differ by about \(211\,\mathrm{m}\) (24{,}802 holes, \(49\%\), are filed in AGD84). We therefore re-derive eastings and northings from the GDA94 longitude and latitude held by the source database, in one zone per region (Eastern Goldfields EPSG:28351, Pilbara EPSG:28350), and keep the reported values in separate columns. All distances used for splitting, de-duplication and retrieval are computed in this frame.

\paragraph{(4) Lithology vocabulary.}
Free-text descriptions are mapped to the 37-code vocabulary (Table~\ref{tab:lith-vocab}) by exact lookup followed by ordered regular-expression rules. On the 1{,}533{,}265 retained intervals, \(68.2\%\) receive a code. The remaining \(31.8\%\) break down as follows:
unmapped raw text \(28.5\%\), spread over 9{,}988 distinct strings, dominated by project-specific shorthand such as \texttt{TAC}, \texttt{OTA}, \texttt{CPZ};
modifier-only text \(2.2\%\);
blank \(0.9\%\);
explicit ``unknown'' \(0.3\%\).
Unmapped intervals keep their raw text in \texttt{lith\_raw} with a null code and are treated as unobserved in every model input and metric. The per-split lithology coverage in Table~\ref{tab:drillbench_splits} therefore measures available supervision, not logging completeness.

\paragraph{(5) Physical-bounds masking.}
Assays are harmonised to 25 elements: Au in ppb, trace elements in ppm, major oxides in wt\%, with Al reported as \(0.5293\times\mathrm{Al_2O_3}\). A value whose magnitude exceeds its element's stoichiometric bound (Table~\ref{tab:phys-bounds}) is set to \texttt{null}; the rest of the record is kept, so masking removes no interval. Over the 70.96M source assay rows, 409{,}821 values are nulled. Negative values within bounds are kept. By Western Australian reporting convention a negative encodes a below-detection-limit result: \texttt{As\_ppm}\(=-3000\) means \(<3000\,\mathrm{ppm}\). Downstream code decodes this convention at use time, and users of the geochemistry channel must do the same.

\begin{table}[h]
\centering
\caption{Per-element physical upper bounds and number of values nulled.}
\label{tab:phys-bounds}
\small
\setlength{\tabcolsep}{4pt}
\begin{tabular}{@{}l r r @{\hspace{10pt}} l r r@{}}
\toprule
Element & Bound & Nulled & Element & Bound & Nulled \\
\midrule
Au (ppb) & \(10^{6}\) & 22{,}090 & Fe (\%) & 72 & 16{,}663 \\
Ag (ppm) & 50{,}000 & 11{,}107 & Al (\%) & 30 & 29{,}843 \\
As (ppm) & 200{,}000 & 643 & SiO\(_2\) (\%) & 100 & 53{,}650 \\
Sb (ppm) & 50{,}000 & 546 & MgO (\%) & 60 & 32{,}736 \\
Bi (ppm) & 50{,}000 & 4{,}146 & CaO (\%) & 60 & 28{,}301 \\
Cu (ppm) & 300{,}000 & 1{,}093 & Na\(_2\)O (\%) & 20 & 30{,}536 \\
Pb (ppm) & 300{,}000 & 712 & K\(_2\)O (\%) & 20 & 29{,}588 \\
Zn (ppm) & 300{,}000 & 2{,}845 & TiO\(_2\) (\%) & 30 & 28{,}153 \\
Ni (ppm) & 100{,}000 & 19{,}058 & P\(_2\)O\(_5\) (\%) & 50 & 24{,}359 \\
Co (ppm) & 50{,}000 & 4{,}275 & LOI (\%) & 50 & 30{,}745 \\
Cr (ppm) & 200{,}000 & 1{,}678 & S (\%) & 60 & 18{,}395 \\
V (ppm) & 10{,}000 & 782 & & & \\
Ba (ppm) & 100{,}000 & 3{,}341 & & & \\
Sr (ppm) & 10{,}000 & 14{,}536 & \multicolumn{2}{r}{\textbf{Total}} & \textbf{409{,}821} \\
\bottomrule
\end{tabular}
\end{table}

\paragraph{(6) Twin-hole de-duplication.}
Statutory reporting means the same physical hole is often filed more than once, each time under a different \texttt{hole\_id} and project name. A hole-level split would then place near-copies on both sides and flatter any method that predicts from neighbouring holes. From each evaluation split we remove every hole that has a \emph{twin} in its reference pool: a hole within \(25\,\mathrm{m}\) collar distance whose full lithology sequence is identical, or whose interval count matches and whose depth grid agrees within \(0.11\,\mathrm{m}\). Table~\ref{tab:dedup} breaks the removals down by split. \textsc{Train}, both spatial OOD splits, and \textsc{pilbara-test} are unchanged; a separate scan against \textsc{train} finds no twins in the OOD splits. Distinct holes within \(25\,\mathrm{m}\) of each other are deliberately kept, since they are legitimate evidence for interpolating methods. The twin search is run on the harmonised coordinates: in the coordinates as reported, a hole filed once in AGD84 and once in GDA94 lies about \(211\,\mathrm{m}\) from its own copy and escapes a \(25\,\mathrm{m}\) radius, which hid 198 verbatim copies (69 in \textsc{val}, 129 in \textsc{test}). Three further \textsc{test} holes are removed because an identical log of at least three layers, on the same depth grid, is filed under a \emph{different} report within \(500\,\mathrm{m}\) of a training hole (re-surveyed collars); they are counted under ``same lith.'' in Table~\ref{tab:dedup}.

\paragraph{Twins in the archive.}
Exclusion by \texttt{hole\_id} does not catch copies filed under a different identifier, and the 25\,m retrieval guard does not catch re-surveyed collars. We therefore scan every evaluation split against the full WA-DASC-full archive, whose collars are re-derived from GDA94 longitude and latitude in the same frame as DrillBench. An archive hole within 500\,m is a twin if its raw lithology text sequence or its mapped sequence is identical over at least three layers, or if its interval count matches and its depth grid agrees within 0.11\,m; raw text is compared because only about 15\% of archive intervals map to our vocabulary. Table~\ref{tab:dedup} reports the matches. All matched archive holes are removed from both the pretraining corpus and the retrieval pool before any result in this paper is computed; the benchmark splits are unchanged.

\begin{table}[h]
\centering
\caption{Twin-hole de-duplication by evaluation split.}
\label{tab:dedup}
\small
\setlength{\tabcolsep}{4pt}
\begin{tabular}{@{}l l r r r r@{}}
\toprule
& & \multicolumn{2}{c}{Holes} & \multicolumn{2}{c}{Removed by rule} \\
\cmidrule(lr){3-4}\cmidrule(l){5-6}
Split & Reference pool & before \(\to\) after & removed & same lith. & same grid \\
\midrule
\textsc{val} & \textsc{train} & 2{,}286 \(\to\) 1{,}951 & 335 & 292 & 43 \\
\textsc{test} & \textsc{train} & 4{,}572 \(\to\) 3{,}856 & 716 & 654 & 62 \\
\midrule
Total & & & 1{,}051 & 946 & 105 \\
\bottomrule
\end{tabular}
\end{table}

\definecolor{lcframe}{HTML}{333333}
\definecolor{lcALLU}{HTML}{E8D8A8}\definecolor{lcCOLL}{HTML}{DFCF96}
\definecolor{lcGRVL}{HTML}{D9C27E}\definecolor{lcREGL}{HTML}{E3D3B4}
\definecolor{lcSAND}{HTML}{F2E2A0}\definecolor{lcTOPS}{HTML}{E5D5B5}
\definecolor{lcLATR}{HTML}{A94A26}\definecolor{lcSAPL}{HTML}{D08A45}
\definecolor{lcCLAY}{HTML}{F4D75E}\definecolor{lcBSLT}{HTML}{2F7355}
\definecolor{lcDOLR}{HTML}{6B8E4E}\definecolor{lcKOMA}{HTML}{7A9E52}
\definecolor{lcANDS}{HTML}{7A9A7A}\definecolor{lcDACI}{HTML}{C9B8D8}
\definecolor{lcPORP}{HTML}{D8A8C8}\definecolor{lcRHYO}{HTML}{E8C8D8}
\definecolor{lcDIOR}{HTML}{B8A8A8}\definecolor{lcGABR}{HTML}{3E7A6A}
\definecolor{lcGRAN}{HTML}{E8A8A8}\definecolor{lcGRND}{HTML}{DFB0A0}
\definecolor{lcPEGM}{HTML}{E9C7E0}\definecolor{lcAMPH}{HTML}{356B50}
\definecolor{lcGNSS}{HTML}{B0A090}\definecolor{lcGRNS}{HTML}{63A075}
\definecolor{lcPHYL}{HTML}{988878}\definecolor{lcQRTZ}{HTML}{EDEDED}
\definecolor{lcSCHI}{HTML}{9A9070}\definecolor{lcSERP}{HTML}{8FBC8F}
\definecolor{lcSKAR}{HTML}{C89878}\definecolor{lcCGLM}{HTML}{C6AD85}
\definecolor{lcSDST}{HTML}{E0A93E}\definecolor{lcSHLE}{HTML}{7189A6}
\definecolor{lcCARB}{HTML}{A5D8EC}\definecolor{lcCHRT}{HTML}{C8C8C0}
\definecolor{lcIRON}{HTML}{6E3413}\definecolor{lcVEIN}{HTML}{FFFFFF}
\definecolor{lcUNKN}{HTML}{CCCCCC}
\newcommand{\sw}[1]{%
  \begingroup
  \setlength{\fboxsep}{0pt}\setlength{\fboxrule}{0.3pt}%
  \fcolorbox{lcframe}{lc#1}{\rule{0pt}{5pt}\rule{9pt}{0pt}}%
  \endgroup}

\subsection{Lithology Vocabulary Details}
\label{app:litho_detail}
\begin{table}[t]
\centering
\caption{\textbf{Shared lithology vocabulary.}
The 37 fine-grained codes (ids $0$--$36$) form the prediction target for both tasks and are shared across data providers. For the cover/bedrock analysis (\S\ref{sec:results}, Table~\ref{tab:app-task1-zones}) each code is additionally assigned to one of two zones: \textsc{cover} groups transported and weathered material above fresh rock, while \textsc{bedrock} groups the in-situ units relevant to primary mineralisation. \textbf{Fill} denotes the colour used in the case-study lithology strips; hatching (\texttt{///}) marks incorrect predictions. Two ids are reserved: \texttt{<UNK>} ($37$) for missing or out-of-vocabulary lithology and \texttt{<CONT>} ($38$) for layer continuation. \texttt{UNKN} ($35$) is a valid fine code for ``undifferentiated'' and is distinct from \texttt{<UNK>}; both are shown in grey and excluded from scoring.}
\label{tab:lith-vocab}
\small
\setlength{\tabcolsep}{4pt}
\begin{tabular}{@{}r c l l @{\hspace{14pt}} r c l l@{}}
\toprule
\textbf{id} & \textbf{Fill} & \textbf{Code} & \textbf{Description} &
\textbf{id} & \textbf{Fill} & \textbf{Code} & \textbf{Description} \\
\midrule
\rowcolor{black!4}
\multicolumn{8}{@{}l}{\textsc{cover} --- transported and weathered material (9 codes)} \\
 0 & \sw{ALLU} & ALLU & Alluvium              & 25 & \sw{REGL} & REGL & Regolith / overburden \\
 7 & \sw{CLAY} & CLAY & Clay                  & 27 & \sw{SAND} & SAND & Sand \\
 8 & \sw{COLL} & COLL & Colluvium             & 28 & \sw{SAPL} & SAPL & Saprolite / saprock \\
17 & \sw{GRVL} & GRVL & Gravel                & 34 & \sw{TOPS} & TOPS & Transported overburden \\
20 & \sw{LATR} & LATR & Laterite / ferricrete  &    &           &      & \\
\midrule
\rowcolor{black!4}
\multicolumn{8}{@{}l}{\textsc{bedrock} --- in-situ rock (27 codes)} \\
 1 & \sw{AMPH} & AMPH & Amphibolite           & 18 & \sw{IRON} & IRON & Ironstone / BIF \\
 2 & \sw{ANDS} & ANDS & Andesite              & 19 & \sw{KOMA} & KOMA & Komatiite \\
 3 & \sw{BSLT} & BSLT & Basalt                & 21 & \sw{PEGM} & PEGM & Pegmatite \\
 4 & \sw{CARB} & CARB & Limestone / dolomite  & 22 & \sw{PHYL} & PHYL & Phyllite \\
 5 & \sw{CGLM} & CGLM & Conglomerate / breccia& 23 & \sw{PORP} & PORP & Porphyry \\
 6 & \sw{CHRT} & CHRT & Chert                 & 24 & \sw{QRTZ} & QRTZ & Quartzite \\
 9 & \sw{DACI} & DACI & Dacite                & 26 & \sw{RHYO} & RHYO & Rhyolite \\
10 & \sw{DIOR} & DIOR & Diorite               & 29 & \sw{SCHI} & SCHI & Schist \\
11 & \sw{DOLR} & DOLR & Dolerite / dyke       & 30 & \sw{SDST} & SDST & Sandstone \\
12 & \sw{GABR} & GABR & Gabbro                & 31 & \sw{SERP} & SERP & Serpentinite / talc \\
13 & \sw{GNSS} & GNSS & Gneiss                & 32 & \sw{SHLE} & SHLE & Shale / siltstone \\
14 & \sw{GRAN} & GRAN & Granite               & 33 & \sw{SKAR} & SKAR & Skarn \\
15 & \sw{GRND} & GRND & Granodiorite          & 36 & \sw{VEIN} & VEIN & Quartz vein \\
16 & \sw{GRNS} & GRNS & Greenstone            &    &           &      & \\
\midrule
\rowcolor{black!4}
\multicolumn{8}{@{}l}{\textit{Unzoned and reserved}} \\
35 & \sw{UNKN} & UNKN & Unknown / undifferentiated & 38 & -- & \texttt{<CONT>} & layer continuation \\
37 & \sw{UNKN} & \texttt{<UNK>} & out-of-vocab / missing & & & & \\
\bottomrule
\end{tabular}
\end{table}

Lithology labels are standardised to a single vocabulary of 37 codes (Table~\ref{tab:lith-vocab}), which is used for training, prediction, and evaluation throughout. Free-text descriptions from all data providers are mapped into these codes by exact matching followed by regular-expression rules (\S\ref{app:dataset_curation}), so that holes logged by different companies and geologists become directly comparable. Intervals whose text cannot be mapped keep their original wording in \texttt{lith\_raw} but carry no code, and are treated as unobserved rather than assigned a fallback class.
\texttt{<UNK>} denotes missing or unmapped lithology and is excluded from training and evaluation; it is distinct from \texttt{UNKN}, which is a valid code meaning ``undifferentiated.'' Cell-level tokenisation additionally uses \texttt{<CONT>} for layer continuation (\S\ref{subsec:tokenization}), and layer-level pretraining uses \texttt{GAP} for unlogged runs (\S\ref{subsec:drillmar-pretrain}).
The \textsc{cover}/\textsc{bedrock} assignment is used only for analysis, not for training or prediction. It follows the regolith--bedrock distinction of \citet{anand2010guide}: \textsc{cover} spans transported material (alluvium, colluvium, gravel, sand) and the in-situ weathering profile (laterite, saprolite, clay), whereas \textsc{bedrock} covers fresh lithologies. Saprolite and clay sit at the boundary, since both are weathering products of the underlying rock; we place them in \textsc{cover} because they are logged by weathering state rather than protolith. \texttt{UNKN} carries no zone and is excluded from the zone analysis.

\section{Implementation Details}
\label{app:implementation_details}

\subsection{Baseline Implementation Details}
\label{app:baselines}

\definecolor{cellbg}{RGB}{230,240,250}    
\definecolor{layerbg}{RGB}{252,238,220}   

\newcommand{\cellmethod}[1]{%
  {\setlength{\fboxsep}{1pt}\colorbox{cellbg}{\emph{#1}}}}

\newcommand{\layermethod}[1]{%
  {\setlength{\fboxsep}{1pt}\colorbox{layerbg}{\emph{#1}}}}
  
We include both cell-level and layer-level statistical baselines. Cell-level methods are shown in \cellmethod{\strut light blue}, while layer-level methods are shown in \layermethod{\strut light orange}. The first five methods operate on the 1\,m cell sequence, while the final two operate directly on stratum-level sequences.

\subsubsection{Frequency and Markov Models}

We include both cell-level and layer-level frequency- and Markov-based baselines. Cell-level methods operate on the 1\,m sequence and are marked in \cellmethod{\strut light blue}, while layer-level methods operate on stratum sequences and are marked in \layermethod{\strut light orange}. Unless stated otherwise, count tables use add-one smoothing. \texttt{UNK} is excluded from count estimation and receives only smoothing mass.

\cellmethod{Unigram} predicts the marginal distribution of the 37 lithology classes in \textsc{train}, without using sequence context.

\cellmethod{$n$-gram} uses an order-3 model,
\[
P(\ell_t \mid \ell_{t-1},\ell_{t-2}),
\]
with interpolated back-off. For a context observed $N_{\mathrm{ctx}}$ times, the higher-order estimate receives weight
\[
\lambda_{\mathrm{ctx}} =
\frac{N_{\mathrm{ctx}}}{N_{\mathrm{ctx}}+\gamma},
\qquad \gamma=8,
\]
and the remaining weight is assigned to the next lower-order model. This is applied recursively from trigram to bigram to unigram, so rare contexts rely more on shorter histories.

\cellmethod{Depth-Markov} uses a bigram additionally conditioned on depth,
\[
P(\ell_t \mid d_t,\ell_{t-1}),
\]
where $d_t$ is the 10\,m depth band containing the interval top. We use 20 bands from 0--200\,m and one overflow band for deeper intervals. If a $(d_t,\ell_{t-1})$ context is unseen, the model backs off as
\[
P(\ell_t\mid d_t,\ell_{t-1})
\rightarrow
P(\ell_t\mid d_t)
\rightarrow
P(\ell_t).
\]

\cellmethod{Persistence} assigns probability $1-\epsilon$ to the previous cell's lithology and distributes the remaining probability $\epsilon$ over the other classes, with $\epsilon=0.01$. At the first cell of a hole, or when the previous cell is \texttt{UNK}, it falls back to the training marginal. Since only about 1.5--2.4\% of cells correspond to stratum boundaries, this provides a simple reference for prediction on the 1\,m grid.

\cellmethod{HMM} is a discrete-emission hidden Markov model with $K=24$ latent states. It is trained with Baum--Welch on a fixed-seed subsample of 8{,}000 training holes for at most 25 EM iterations, with convergence tolerance $10^{-3}$ and a self-transition bias of 5.0. \texttt{UNK} observations are treated as missing by assigning them uniform emission likelihood across states, so they do not contribute to emission-count updates. Evaluation uses all holes in each split.

\layermethod{Layer Markov} operates directly on stratum lithologies. It uses an order-2 transition model with hierarchical $2\!\to\!1\!\to\!0$ back-off. Each higher-order distribution is smoothed toward the next lower-order distribution using a pseudo-count prior of weight $\kappa=5$, while the context-free marginal uses add-one smoothing. To convert layer predictions to the common 1\,m grid, the model also estimates a discrete thickness hazard,
\[
P(T=r \mid T\ge r),
\]
where $T$ is stratum thickness in cells and $r$ is the current run length. A separate hazard is estimated for each lithology and shrunk toward the global hazard with weight $\kappa=5$.

\layermethod{Layer marginal} removes sequence dependence. Each new stratum lithology is sampled from the training marginal, while its thickness is generated from the global thickness hazard. It serves as a Task~2 reference for distinguishing sequence generation quality from simply matching overall lithology frequencies.

\subsubsection{Geostatistical Models}

The geostatistical baselines fall into two groups according to where their information comes from. \emph{Neighbour-based methods} retrieve nearby logged drillholes at prediction time, whereas \emph{region-level methods} estimate transition or pattern statistics from \textsc{train} and use no query-side neighbours.

\textbf{Retrieval pool.}
To make the fair comparison, all neighbour-based baselines use the same retrieval pool as our model (\S\ref{subsec:retrieval}): DrillBench \textsc{train} and WA-DASC-full, with the archive twins of every evaluation hole removed. Retrieval is restricted to the same geological region: Eastern Goldfields queries use \textsc{train} plus archive holes (pretraining corpus). Thus, these methods differ from our retriving approach in how they use retrieved evidence, not in which drillholes are available.

\emph{Coordinate frame.}
DrillBench collars are stored in local MGA zones: \textsc{train}, \textsc{val}, \textsc{test}, and \textsc{ood-regional} use zone~51, whereas \textsc{ood-distal} and the Pilbara splits use zone~50. All collars are converted to a common metric coordinate system before computing distances.\footnote{The zone assignment was verified using 5{,}038 \textsc{train} holes shared with the archive and by checking nearest-archive distances for the remaining splits. The correct transformation places holes in their expected provinces, whereas using the wrong zone produces displacements of tens to hundreds of kilometres.}


\emph{Duplicate guard and missing evidence.}
As in our receip, any pool hole within $25\,\mathrm{m}$ of the query collar is excluded to avoid re-reported copies of the same physical drillhole. At a queried depth, neighbours that do not reach that depth provide no vote. If no retrieved neighbour covers the queried position, the prediction falls back to the \textsc{train} marginal.

\textbf{Neighbour-based methods.}

\cellmethod{Spatial $k$-NN} retrieves the $k=10$ nearest eligible collars in the same region. At each query depth, the lithology of every available neighbour is obtained by interval lookup, and class probabilities are formed using inverse-distance weighting.

\cellmethod{Indicator kriging} uses the same depth-wise lookup with $k=20$ neighbours, but replaces inverse-distance weighting with simple indicator kriging. We use a spherical variogram with range $1\,\mathrm{km}$, nugget $0.05$, and sill $0.95$, and blend the resulting estimate with the \textsc{train} marginal using $\alpha=0.05$. As neighbours move beyond the variogram range, their contribution decreases and the prediction approaches the marginal.

\cellmethod{Sequential indicator simulation (SIS)} uses the same kriging conditional distribution but draws $M$ realisations and reports their empirical class distribution. It therefore provides a stochastic counterpart to indicator kriging, suitable for Task~2 where complete stratigraphic continuations must be generated.

\cellmethod{Retrieval-only readout} isolates the information contained in our retrieval vector $r_i$ by removing the autoregressive sequence model. It uses the same retrieval pool, $k$, duplicate guard, and depth alignment as ours. We evaluate both a parameter-free readout, which predicts the arg-max of $r_i$, and a readout learned on \textsc{train} layer-start cells. Their accuracies differ by less than one percentage point, indicating that performance is mainly limited by the retrieved evidence rather than the readout.

When collar elevation is available, neighbour lithology is aligned by elevation rather than down-hole depth. Because the public archive does not provide collar elevation, this adjustment has no effect when all retrieved evidence comes from archive holes; the same limitation applies to our receip.

\textbf{Region-level methods.}

These methods use only \textsc{train} sequences when fitting their statistics and do not access the retrieval pool at prediction time.

\cellmethod{TPROGS} separates persistence within a lithology from transitions between lithologies. For each region, it estimates
\[
p_{\mathrm{self}}(c),
\]
the probability that lithology $c$ continues into the next cell, and an embedded transition matrix
\[
T_e(c\!\to\!c'),
\]
which gives the probability of transitioning from $c$ to a different lithology $c'$ once a boundary occurs. This separates stratum run length from lithology-transition structure.

\cellmethod{MPS-lite} is a vertical approximation of multi-point statistics. The regional \textsc{train} sequences form the training patterns. At position $t$, the model matches the preceding $W=5$ tokens and uses the empirical distribution of the following token as its prediction. Sparse patterns progressively back off to shorter contexts using
\[
\lambda_{\mathrm{ctx}}
=
\frac{N_{\mathrm{ctx}}}{N_{\mathrm{ctx}}+\gamma},
\qquad
\gamma=8,
\]
where $N_{\mathrm{ctx}}$ is the number of training occurrences of the current pattern. This tests whether a longer local pattern provides useful information beyond the shorter contexts captured by the $n$-gram baseline.

\subsubsection{Feature-based Machine Learning}
\cellmethod{XGBoost} represents the feature-based machine-learning baseline. It uses gradient-boosted decision trees with only information available at prediction time: z-scored collar coordinates, region one-hot features, depth, current run length, and previous lithologies. We train 200 boosting rounds with maximum tree depth 6 and learning rate 0.1. Training uses early stopping if validation performance does not improve for 20 consecutive rounds.

\subsubsection{Cell-level autoregressive models.}

All five neural baselines operate directly on the 1 meter token sequence and use only lithology history as input; collar information, geochemistry, and auxiliary objectives are excluded. They are trained with teacher forcing and cross-entropy, so the main difference between them is how past tokens are represented and combined.

\cellmethod{LSTM} uses a 64-dimensional token embedding, 256 hidden units, 2 recurrent layers, and dropout 0.1. It is trained for 20 epochs with AdamW using learning rate $2\times10^{-3}$, weight decay $10^{-4}$, and batch size 64. The recurrent hidden state summarises the sequence observed above the current cell.

\cellmethod{GRU} uses the same embedding size, hidden size, number of layers, dropout, optimiser, and training schedule as the LSTM, replacing only the LSTM recurrent unit with a GRU. This provides a controlled comparison between the two recurrent architectures.

\cellmethod{Transformer} replaces recurrence with causal self-attention. It uses $d_{\text{model}}=192$, 6 attention heads, 6 layers, feed-forward width 768, sinusoidal positional encodings, context length 2048, and dropout 0.1. It is trained for 30 epochs with AdamW using learning rate $3\times10^{-4}$, batch size 32, and a warm-up followed by inverse-square-root decay.

\cellmethod{TCN} uses six residual blocks, each containing two dilated causal convolutions with kernel size 3. The dilation factors are $1,2,4,8,16,32$, giving a receptive field of 253 cells. Each block uses GELU, dropout, a residual connection, and LayerNorm. The hidden width and optimisation schedule match the Transformer.

\cellmethod{Mamba} uses $d_{\text{model}}=192$ and 8 selective state-space blocks, with state size 16, convolution width 4, and expansion factor 2. The blocks are arranged in a pre-norm residual stack and trained with the same optimisation schedule as the Transformer.

For all five models, we retain the checkpoint with the lowest negative log-likelihood on \textsc{val}.

\subsubsection{Our layer-level model.}
The body is deliberately small: $d_{\text{model}} = 192$ with a 2-layer GRU or a
3-layer, 4-head Transformer, at most 128 strata per hole, 0.65--1.6\,M
parameters. Thickness and depth are not fed back as inputs, since both are
province fingerprints; the inputs are the lithology embedding and, when
retrieval is enabled, $\phi(r_i)$. Training runs for a fixed 20 epochs with
AdamW (learning rate $3\times10^{-4}$, weight decay 0.01, batch size 256) and
label smoothing 0.3 on the lithology head; the thickness head is trained by
Gaussian negative log-likelihood on $\log D$ with weight 0.3. No checkpoint
selection is performed, so \textsc{val} is never used for model choice.
Retrieval draws $k=10$ neighbours from the pool, excludes neighbours closer than
25\,m to the query collar (a duplicate-report guard), aligns the look-up by
elevation rather than hole depth, and summarises the neighbours into a 37-class
inverse-distance-weighted lithology distribution; the whole retrieval
vector is dropped for 30\% of training holes.
Pretraining runs on the deduplicated multi-province corpus,
layerised with an explicit \texttt{GAP} token for unlogged runs, over a
lithology vocabulary plus \texttt{GAP} and \texttt{UNK}; batch size 512,
learning rate $3\times10^{-4}$, label smoothing 0.3, 24 epochs for the
checkpoint used in the main tables, with auxiliary thickness, masked-token, and
chemistry-anchor losses weighted 0.3, 0.5 and 0.5.

\paragraph{Evaluation and reporting.} Layer-level rows of Table~\ref{tab:main-tasks} are run with ten seeds, other trained arms with three, and single-run baselines once; we report the mean. Task~1 reports macro-F1 at layer-start cells, with top-1 as a secondary metric, and 95\% confidence intervals from a hole-level cluster bootstrap; Task~2 reports the joint KL of free-running rollouts under the protocol of \S\ref{app:fair-eval}.

\paragraph{Compute.}
All runs are single-GPU on a SLURM cluster (V100, A6000 or H100 depending on
availability). A layer-level fine-tuning run takes 3--4 minutes, a cell-level
run with retrieval about 17 minutes, and corpus pretraining 20 minutes (GRU) to
7 hours (12-layer Transformer). The complete set of results in this paper is
roughly 400 GPU-hours, dominated by the pretraining checkpoints and the
data-size sweep. Cross-province numbers are sensitive to the CPU/GPU split on
\textsc{ood-distal}, so every paired comparison is run on the same device type.
\providecolor{unitgray}{HTML}{E5E7EB}
\begin{table}[t]
\centering
\footnotesize
\setlength{\tabcolsep}{4.5pt}
\renewcommand{\arraystretch}{1.08}
\caption{\textbf{Model sizes and compute.} Wall-clock is one run on one GPU
(V100, A6000 or H100, whichever was free); classical baselines are fitted on
CPU. Deep arms are trained with three seeds, so the cost of a reported row is
three times the figure shown. Pretraining is paid once per checkpoint and is
reused by every arm that starts from it.}
\label{tab:compute}
\begin{adjustbox}{width=\columnwidth}
\begin{tabular}{@{}l l r r@{}}
\toprule
\textbf{Method} & \textbf{Inputs} & \textbf{Params} & \textbf{Wall-clock} \\
\midrule
\rowcolor{unitgray}
\multicolumn{4}{@{}l}{\textbf{Classical and geostatistical} (fit\,+\,score, CPU)} \\
Unigram / per-region unigram & lith. & -- & $<1$\,min \\
Bigram / $3$-gram / depth-Markov & lith. & -- & 1\,min \\
HMM ($K{=}24$) & lith. & -- & 1\,min \\
Layer Markov (order 2) & lith. & -- & 1\,min \\
TPROGS & lith.\,+\,loc. & -- & 1\,min \\
Spatial $k$-NN / kriging / SIS & loc. & -- & 1--2\,min \\
MPS-lite & lith.\,+\,loc. & -- & 2\,min \\
XGBoost & lith.\,+\,loc. & -- & 47\,min \\
\midrule
\rowcolor{unitgray}
\multicolumn{4}{@{}l}{\textbf{Cell-level autoregressive} (1\,m token, 20--30 epochs)} \\
GRU & lith. & 0.65\,M & 4\,min \\
LSTM & lith. & 0.87\,M & 7\,min \\
TCN & lith. & 1.35\,M & 11\,min \\
Transformer & lith. & 2.68\,M & 16\,min \\
Mamba & lith. & 2.03\,M & 4.7\,h \\
\addlinespace[2pt]
GRU + loc.\,/ + chem. & \multicolumn{1}{c}{--} & 0.73\,M & 5--7\,min \\
LSTM + chem.\,/ + chem.\,+\,loc. & \multicolumn{1}{c}{--} & 0.88--1.01\,M & 10\,min \\
TCN + chem.\,+\,loc. & \multicolumn{1}{c}{--} & 1.40\,M & 11\,min \\
Transformer + chem.\,+\,loc. & \multicolumn{1}{c}{--} & 2.74\,M & 23\,min \\
Mamba + chem.\,+\,loc. & \multicolumn{1}{c}{--} & 2.08\,M & 4.7\,h \\
\midrule
\rowcolor{unitgray}
\multicolumn{4}{@{}l}{\textbf{Layer level} (stratum token, 20 epochs)} \\
GRU & lith. & 0.65\,M & 4\,min \\
Transformer & lith. & 1.6\,M & 6\,min \\
\rowcolor{proposedbg}
GRU + retrieval ($\pm$ pretraining) & lith.\,+\,retr. & 0.68\,M & 4\,min \\
\rowcolor{proposedbg}
\quad + Task 2 rollouts & \multicolumn{1}{c}{--} & \multicolumn{1}{c}{--} & 17\,min \\
\midrule
\rowcolor{unitgray}
\multicolumn{4}{@{}l}{\textbf{Cell level, ours} (10 epochs, archive retrieval)} \\
\rowcolor{proposedbg}
GRU + retrieval ($\pm$ pretraining) & lith.\,+\,retr. & 3.4\,M & 10\,min \\
\rowcolor{proposedbg}
Transformer + retrieval ($\pm$ pretr.) & lith.\,+\,retr. & 38.8\,M & 27\,min \\
\rowcolor{proposedbg}
\quad + Task 2 rollouts & \multicolumn{1}{c}{--} & \multicolumn{1}{c}{--} & 53\,min \\
\midrule
\rowcolor{unitgray}
\multicolumn{4}{@{}l}{\textbf{Corpus pretraining}} \\
Layer-level GRU, 24 epochs & lith.\,(+\,chem.) & 0.65\,M & 20\,min \\
Cell-level GRU, $d{=}512$, 2 layers & lith. & 3.4\,M & 22\,min \\
Cell-level Transformer, $d{=}512$, 12 layers & lith. & 38.8\,M & 7\,h \\
\midrule
\rowcolor{unitgray}
\multicolumn{4}{@{}l}{\textbf{Zero-shot LLMs} (150 holes per split, no training)} \\
Qwen3-8B / 14B / 32B, phi-4 & prompt & 8--32\,B & 2.0\,h per arm \\
GPT-6, Gemini-pro & prompt & -- & batch API \\
\bottomrule
\end{tabular}
\end{adjustbox}
\end{table}

\subsection{Evaluation Protocol Details}
\label{app:evaluation}


\paragraph{Unified output space.}
The evaluated methods predict different objects. Cell-level methods directly produce a distribution over the next 1\,m token, whereas layer-level methods predict the lithology and thickness of the next stratum. To make them directly comparable, every method is converted to the same per-cell distribution over $\{\texttt{CONT}\}\cup\{1,\ldots,37\}$.

For a layer-level model, let $q(k)$ denote the probability that the next stratum has lithology $k$, and let $\lambda$ denote the probability that the current stratum ends at the current cell. Its equivalent cell-level distribution is
\[
P(\texttt{CONT}) = 1-\lambda,
\qquad
P(k)=\lambda q(k).
\]
Cell-level methods already produce this distribution directly. All methods are therefore scored by the same evaluation code on the same 1\,m grid.

\paragraph{Task~1: next-layer lithology prediction.}
Task~1 asks which lithology begins at the next true stratum boundary. Given the true history above a boundary, we evaluate the predicted distribution at the first cell of the new stratum and report macro-F1, averaged with equal weight over the lithology classes that have at least 20 true next-stratum events in the split. \texttt{CONT} is excluded from the argmax because the boundary location is already given in this task.

We do not use ordinary per-cell accuracy as the primary metric because about 96\% of logged cells are continuations and new strata start at only 1.2--2.6\% of cells. A model that always predicts \texttt{CONT}, i.e.\ that the current stratum simply continues, therefore reaches about 96\% per-cell accuracy without learning any geological transition. Evaluating only true stratum starts instead isolates the intended question: \emph{which rock comes next?}

We additionally report top-1 accuracy, balanced accuracy, support-weighted F1 and Cohen's $\kappa$ as secondary metrics (App.~\ref{app:results_task1}); on OOD-R and OOD-D, where single exploration reports hold hundreds of holes, top-1 is macro-averaged over reports. Confidence intervals are 95\% bootstrap intervals ($B{=}1{,}000$) that resample drillholes rather than individual cells, because observations within a hole are strongly correlated; trained models additionally resample seeds.

\paragraph{Task~2: free-running generation.}
Task~2 evaluates whether a model can continue a partially observed drillhole using its own predictions. Each model receives the true upper 50\% of a hole, then samples one cell at a time until the known total depth is reached.

A generated sequence should not be compared with the true suffix cell by cell: two plausible stratigraphic columns may contain similar lithologies and thicknesses but place a boundary a few metres apart. We therefore evaluate the distribution of generated strata rather than exact cell alignment.

Our primary metric is the KL divergence $\mathrm{KL}(P_{\text{true}}\,\|\,P_{\text{gen}})$ between the true and generated joint distributions of
\[
(\text{lithology},\text{thickness bin}),
\]
pooled over all evaluated holes and rollouts. Lithology uses the 37 DrillBench classes. Thickness is discretised into seven log-spaced bins with edges
\[
1,\,2,\,3,\,5,\,8,\,13,\,21,\,31\ \text{m},
\]
with the final bin absorbing thicker strata. Before counting, both true and generated suffixes are converted from cells back into strata by run-length encoding, and strata of unknown lithology are dropped. KL divergence uses $\varepsilon=10^{-6}$ numerical smoothing.

The joint metric is used because lithology and thickness marginals can each be matched independently while still producing unrealistic stratigraphy. For example, a model may reproduce the correct lithology frequencies while fragmenting units into unrealistically thin layers, or match the thickness distribution while assigning it to the wrong rocks.

\paragraph{Generation protocol and calibration.}
All generators use the same true 50\% prefix, sample autoregressively from their full predicted distribution, and stop at the logged hole depth. We draw five rollouts per hole on a fixed, report-stratified sample: each exploration report contributes at most $c$ randomly drawn holes, with $c$ the smallest cap that yields at least 600 holes, giving 610--827 holes per split. The sample is drawn once, independently of any model, so every method is scored on the same holes. Confidence intervals resample whole reports ($B{=}1{,}000$), because holes of one report share a project and a logging vocabulary.

Because some generators naturally over- or under-segment, we fit one scalar calibration factor $s$ on \textsc{train} rollouts for every method:
\[
P(\texttt{CONT}) = 1-s\lambda,
\qquad
P(k)=s\lambda q(k),
\]
with $s\lambda$ clipped to $[10^{-4},\,0.99]$. The factor rescales the layer-start probability so that the generated boundary rate matches the training data. The same procedure is applied to all generators, and $s$ is never fitted on validation or test data. Calibration is used only for Task~2.

\paragraph{Fair comparison.}
\label{app:fair-eval}
All methods are converted to the same per-cell output space and scored by the same implementation. All generators receive the same prefix, stopping depth and hole sample, and the same train-only boundary calibration is applied to every generator. The layer-level neural models, whose lithology heads are trained with label smoothing, additionally sharpen $q$ with a temperature selected on \textsc{train} rollouts; the other methods receive no such temperature. Spatial $k$-NN is marked with $^{*}$ in Task~2 because it is a depth-wise lookup rather than an autoregressive generator.

\section{Additional Results}
\label{app:results}

\providecolor{unitgray}{HTML}{E5E7EB}
\subsection{Additional Results: Next-Layer Prediction}
\label{app:results_task1}

\subsubsection{Protocol and uncertainty}
All numbers are computed from per-event logs (hole, true next lithology, predicted next lithology) written by the shared
evaluation harness, so every method is scored on exactly the same layer starts. Macro-F1 is averaged over the lithology
classes with at least 20 events in the split. Top-1 is pooled on \textsc{test} and \textsc{pilbara-test} and
macro-averaged over exploration reports on the two OOD rings, where single reports hold up to several hundred holes.
Intervals are $95\%$ hole-cluster bootstrap intervals ($B{=}1000$): all layer starts of a hole are resampled together,
and rows trained with several seeds (10 for the five layer-level rows of Table~\ref{tab:main-tasks}, 3 for every other
trained row) additionally resample seeds. One resample per split is shared by all rows, so the difference between any two
rows is a \emph{paired} bootstrap. The unpaired intervals are wide on macro-F1 ($\pm3$--$5$ points, because rare classes
carry equal weight); the evidence for every claim in \S\ref{sec:results} is therefore the paired difference
(Table~\ref{tab:app-task1-paired}).

\subsubsection{All baselines}
\begin{table*}[t]
\centering
\footnotesize
\setlength{\tabcolsep}{3pt}
\caption{\textbf{Task~1, all baselines} on the four evaluation splits. Layer-level neural rows use label smoothing ($\varepsilon{=}0.3$). $\pm$: half-width of the 95\% hole-cluster bootstrap interval ($B{=}1000$); trained rows also resample seeds. macro-F1 over classes with $\geq$20 events; top-1 on OOD-R / OOD-D is macro-averaged over exploration reports.}
\label{tab:app-task1-baselines}
\begin{adjustbox}{width=\textwidth}
\begin{tabular}{@{}l c rrrr rrrr@{}}
\toprule
& & \multicolumn{4}{c}{\textbf{macro-F1 (\%)}} & \multicolumn{4}{c}{\textbf{top-1 accuracy (\%)}} \\
\cmidrule(lr){3-6}\cmidrule(lr){7-10}
\textbf{Method} & \textbf{runs} & Test & OOD-R & OOD-D & Pil-T & Test & OOD-R & OOD-D & Pil-T \\
\midrule
\rowcolor{unitgray}
\multicolumn{10}{@{}l}{\textbf{Frequency / Markov (cell level)}} \\
Unigram & 1 & 2.0{\tiny$\pm$0.2} & 0.8{\tiny$\pm$0.1} & 1.1{\tiny$\pm$0.2} & 1.7{\tiny$\pm$0.2} & 20.0{\tiny$\pm$1.0} & 12.0{\tiny$\pm$1.3} & 13.2{\tiny$\pm$2.0} & 23.7{\tiny$\pm$1.2} \\
Per-region unigram & 1 & 2.0{\tiny$\pm$0.2} & 0.8{\tiny$\pm$0.1} & 1.1{\tiny$\pm$0.2} & 1.7{\tiny$\pm$0.2} & 20.0{\tiny$\pm$1.0} & 12.0{\tiny$\pm$1.3} & 13.2{\tiny$\pm$2.0} & 23.7{\tiny$\pm$1.2} \\
Persistence & 1 & 2.5{\tiny$\pm$0.3} & 0.6{\tiny$\pm$0.1} & 0.7{\tiny$\pm$0.1} & 1.6{\tiny$\pm$0.2} & 11.6{\tiny$\pm$0.8} & 3.4{\tiny$\pm$1.3} & 4.8{\tiny$\pm$1.7} & 10.5{\tiny$\pm$1.0} \\
Bigram & 1 & 4.5{\tiny$\pm$0.7} & 1.1{\tiny$\pm$0.2} & 1.2{\tiny$\pm$0.2} & 1.7{\tiny$\pm$0.2} & 21.3{\tiny$\pm$1.0} & 12.2{\tiny$\pm$1.3} & 13.2{\tiny$\pm$2.1} & 22.7{\tiny$\pm$1.2} \\
$3$-gram & 1 & 9.4{\tiny$\pm$1.3} & 3.2{\tiny$\pm$1.0} & 4.0{\tiny$\pm$0.9} & 3.7{\tiny$\pm$0.8} & 23.3{\tiny$\pm$0.9} & 13.6{\tiny$\pm$1.4} & 15.9{\tiny$\pm$2.4} & 21.0{\tiny$\pm$1.0} \\
Depth-Markov & 1 & 9.0{\tiny$\pm$1.2} & 3.7{\tiny$\pm$0.5} & 3.4{\tiny$\pm$0.6} & 1.5{\tiny$\pm$0.2} & 28.0{\tiny$\pm$1.2} & 18.0{\tiny$\pm$1.6} & 15.6{\tiny$\pm$2.4} & 10.4{\tiny$\pm$0.7} \\
HMM ($K{=}24$) & 1 & 2.0{\tiny$\pm$0.2} & 0.8{\tiny$\pm$0.1} & 1.1{\tiny$\pm$0.2} & 1.7{\tiny$\pm$0.2} & 20.0{\tiny$\pm$1.0} & 12.0{\tiny$\pm$1.3} & 13.2{\tiny$\pm$2.0} & 23.7{\tiny$\pm$1.2} \\
\midrule
\rowcolor{unitgray}
\multicolumn{10}{@{}l}{\textbf{Geostatistical (cell level)}} \\
Spatial $k$-NN, train pool & 1 & 71.5{\tiny$\pm$3.7} & 0.9{\tiny$\pm$0.1} & 2.0{\tiny$\pm$0.3} & 2.1{\tiny$\pm$0.2} & 78.2{\tiny$\pm$1.1} & 12.5{\tiny$\pm$1.3} & 14.0{\tiny$\pm$2.1} & 24.2{\tiny$\pm$1.2} \\
Spatial $k$-NN, archive pool & 1 & 68.9{\tiny$\pm$3.8} & 11.5{\tiny$\pm$1.4} & 9.6{\tiny$\pm$1.3} & 8.0{\tiny$\pm$1.0} & 75.7{\tiny$\pm$1.1} & 24.9{\tiny$\pm$1.9} & 23.5{\tiny$\pm$2.6} & 19.8{\tiny$\pm$1.2} \\
Indicator kriging, train pool & 1 & 71.6{\tiny$\pm$3.8} & 0.9{\tiny$\pm$0.1} & 2.0{\tiny$\pm$0.3} & 2.1{\tiny$\pm$0.2} & 76.1{\tiny$\pm$1.1} & 12.5{\tiny$\pm$1.3} & 14.0{\tiny$\pm$2.1} & 24.2{\tiny$\pm$1.2} \\
Indicator kriging, archive pool & 1 & 69.5{\tiny$\pm$3.8} & 6.4{\tiny$\pm$0.8} & 7.6{\tiny$\pm$1.7} & 6.8{\tiny$\pm$1.1} & 74.5{\tiny$\pm$1.1} & 22.8{\tiny$\pm$1.8} & 24.3{\tiny$\pm$2.3} & 23.8{\tiny$\pm$1.2} \\
SIS, train pool & 1 & 68.3{\tiny$\pm$3.8} & 4.0{\tiny$\pm$0.5} & 4.6{\tiny$\pm$0.7} & 2.4{\tiny$\pm$0.3} & 74.8{\tiny$\pm$1.2} & 11.2{\tiny$\pm$1.5} & 8.5{\tiny$\pm$1.5} & 6.6{\tiny$\pm$0.5} \\
SIS, archive pool & 1 & 66.2{\tiny$\pm$3.7} & 8.3{\tiny$\pm$0.9} & 8.5{\tiny$\pm$1.1} & 5.6{\tiny$\pm$0.7} & 73.2{\tiny$\pm$1.2} & 19.6{\tiny$\pm$1.5} & 19.9{\tiny$\pm$2.6} & 14.0{\tiny$\pm$1.0} \\
MPS-lite & 1 & 18.8{\tiny$\pm$2.4} & 5.6{\tiny$\pm$1.1} & 6.4{\tiny$\pm$1.3} & 7.4{\tiny$\pm$1.4} & 28.7{\tiny$\pm$1.1} & 16.8{\tiny$\pm$1.3} & 16.2{\tiny$\pm$2.5} & 22.7{\tiny$\pm$1.1} \\
TPROGS & 1 & 4.5{\tiny$\pm$0.7} & 1.1{\tiny$\pm$0.2} & 1.2{\tiny$\pm$0.2} & 1.7{\tiny$\pm$0.2} & 21.3{\tiny$\pm$1.0} & 12.2{\tiny$\pm$1.3} & 13.2{\tiny$\pm$2.1} & 22.7{\tiny$\pm$1.2} \\
\midrule
\rowcolor{unitgray}
\multicolumn{10}{@{}l}{\textbf{Machine learning (cell level)}} \\
XGBoost & 1 & 64.8{\tiny$\pm$4.9} & 2.4{\tiny$\pm$0.3} & 2.5{\tiny$\pm$0.4} & 2.2{\tiny$\pm$0.3} & 76.1{\tiny$\pm$1.1} & 19.3{\tiny$\pm$1.7} & 11.5{\tiny$\pm$2.2} & 11.4{\tiny$\pm$0.8} \\
\midrule
\rowcolor{unitgray}
\multicolumn{10}{@{}l}{\textbf{Deep learning, cell level, lithology only}} \\
LSTM & 1 & 58.9{\tiny$\pm$5.1} & 25.8{\tiny$\pm$3.1} & 22.2{\tiny$\pm$3.4} & 22.3{\tiny$\pm$2.5} & 60.1{\tiny$\pm$1.2} & 39.1{\tiny$\pm$2.0} & 31.7{\tiny$\pm$2.7} & 38.5{\tiny$\pm$1.3} \\
GRU & 1 & 58.8{\tiny$\pm$5.3} & 21.4{\tiny$\pm$2.6} & 23.0{\tiny$\pm$3.4} & 23.2{\tiny$\pm$2.2} & 60.3{\tiny$\pm$1.1} & 37.1{\tiny$\pm$2.0} & 35.5{\tiny$\pm$3.0} & 36.3{\tiny$\pm$1.4} \\
Mamba & 1 & 59.2{\tiny$\pm$5.0} & 24.8{\tiny$\pm$2.6} & 22.2{\tiny$\pm$3.2} & 19.4{\tiny$\pm$1.6} & 60.1{\tiny$\pm$1.2} & 38.5{\tiny$\pm$1.9} & 34.9{\tiny$\pm$2.9} & 35.9{\tiny$\pm$1.2} \\
TCN & 1 & 51.6{\tiny$\pm$5.3} & 17.9{\tiny$\pm$2.2} & 17.5{\tiny$\pm$2.5} & 16.3{\tiny$\pm$1.7} & 58.6{\tiny$\pm$1.1} & 34.9{\tiny$\pm$2.1} & 32.4{\tiny$\pm$2.7} & 31.7{\tiny$\pm$1.3} \\
Transformer & 1 & 60.3{\tiny$\pm$5.1} & 26.5{\tiny$\pm$3.0} & 23.6{\tiny$\pm$3.4} & 22.5{\tiny$\pm$2.2} & 60.6{\tiny$\pm$1.2} & 39.7{\tiny$\pm$2.0} & 34.8{\tiny$\pm$3.0} & 38.2{\tiny$\pm$1.3} \\
\midrule
\rowcolor{unitgray}
\multicolumn{10}{@{}l}{\textbf{Deep learning, cell level, conditioned on location / geochemistry}} \\
LSTM + location & 1 & 73.4{\tiny$\pm$4.0} & 11.5{\tiny$\pm$1.5} & 10.5{\tiny$\pm$1.8} & 7.5{\tiny$\pm$1.2} & 80.5{\tiny$\pm$1.0} & 23.1{\tiny$\pm$1.7} & 20.4{\tiny$\pm$2.0} & 22.6{\tiny$\pm$1.1} \\
LSTM + geochem & 1 & 65.0{\tiny$\pm$5.3} & 17.8{\tiny$\pm$2.2} & 17.2{\tiny$\pm$2.5} & 9.2{\tiny$\pm$1.3} & 68.1{\tiny$\pm$1.0} & 34.0{\tiny$\pm$2.3} & 27.2{\tiny$\pm$2.7} & 19.9{\tiny$\pm$1.2} \\
LSTM + geochem + location & 1 & 75.6{\tiny$\pm$4.4} & 10.4{\tiny$\pm$1.4} & 7.8{\tiny$\pm$1.4} & 3.4{\tiny$\pm$0.5} & 81.1{\tiny$\pm$0.9} & 30.3{\tiny$\pm$2.0} & 18.0{\tiny$\pm$2.3} & 8.0{\tiny$\pm$0.8} \\
GRU + location & 1 & 73.6{\tiny$\pm$5.5} & 13.0{\tiny$\pm$1.6} & 3.1{\tiny$\pm$0.6} & 0.5{\tiny$\pm$0.1} & 79.6{\tiny$\pm$0.9} & 34.4{\tiny$\pm$2.4} & 12.1{\tiny$\pm$1.9} & 3.0{\tiny$\pm$0.4} \\
GRU + geochem & 1 & 65.6{\tiny$\pm$5.3} & 19.9{\tiny$\pm$2.2} & 17.7{\tiny$\pm$2.5} & 11.0{\tiny$\pm$1.5} & 67.7{\tiny$\pm$1.0} & 38.1{\tiny$\pm$2.3} & 30.8{\tiny$\pm$2.9} & 23.2{\tiny$\pm$1.2} \\
GRU + geochem + location & 1 & 75.1{\tiny$\pm$3.8} & 15.2{\tiny$\pm$2.0} & 3.7{\tiny$\pm$0.8} & 2.5{\tiny$\pm$0.4} & 80.4{\tiny$\pm$0.9} & 32.2{\tiny$\pm$2.2} & 12.4{\tiny$\pm$2.1} & 11.3{\tiny$\pm$1.2} \\
Transformer + geochem + location & 1 & \best{77.2}{\tiny$\pm$4.9} & 18.1{\tiny$\pm$2.2} & 11.0{\tiny$\pm$1.7} & 7.2{\tiny$\pm$1.0} & \best{81.8}{\tiny$\pm$0.9} & 32.8{\tiny$\pm$2.1} & 21.8{\tiny$\pm$2.6} & 22.5{\tiny$\pm$1.1} \\
Mamba + geochem + location & 1 & 74.2{\tiny$\pm$5.2} & 18.7{\tiny$\pm$2.3} & 10.6{\tiny$\pm$1.7} & 9.4{\tiny$\pm$1.2} & 80.8{\tiny$\pm$0.9} & 33.9{\tiny$\pm$2.2} & 24.6{\tiny$\pm$2.7} & 26.4{\tiny$\pm$1.2} \\
TCN + geochem + location & 1 & 66.5{\tiny$\pm$5.6} & 15.4{\tiny$\pm$1.8} & 13.0{\tiny$\pm$1.9} & 9.2{\tiny$\pm$1.5} & 76.0{\tiny$\pm$1.0} & 35.3{\tiny$\pm$2.3} & 20.6{\tiny$\pm$2.3} & 21.3{\tiny$\pm$1.1} \\
\midrule
\rowcolor{unitgray}
\multicolumn{10}{@{}l}{\textbf{Layer level}} \\
Layer marginal & 1 & 2.0{\tiny$\pm$0.2} & 0.8{\tiny$\pm$0.1} & 1.1{\tiny$\pm$0.2} & 1.7{\tiny$\pm$0.2} & 20.0{\tiny$\pm$1.0} & 12.0{\tiny$\pm$1.3} & 13.2{\tiny$\pm$2.0} & 23.7{\tiny$\pm$1.2} \\
Layer Markov & 1 & 56.7{\tiny$\pm$5.0} & 36.3{\tiny$\pm$4.0} & 26.1{\tiny$\pm$3.7} & 22.0{\tiny$\pm$3.3} & 57.0{\tiny$\pm$1.2} & 37.4{\tiny$\pm$1.8} & 32.8{\tiny$\pm$2.6} & 36.5{\tiny$\pm$1.4} \\
GRU & 10 & 59.4{\tiny$\pm$5.0} & 37.4{\tiny$\pm$3.5} & 28.3{\tiny$\pm$4.0} & 28.2{\tiny$\pm$3.1} & 58.6{\tiny$\pm$1.0} & 40.5{\tiny$\pm$1.8} & 34.7{\tiny$\pm$3.2} & 42.0{\tiny$\pm$1.3} \\
Transformer & 10 & 58.9{\tiny$\pm$4.9} & 36.6{\tiny$\pm$3.5} & 29.1{\tiny$\pm$3.9} & 26.0{\tiny$\pm$2.9} & 58.3{\tiny$\pm$1.1} & 40.2{\tiny$\pm$1.8} & 35.1{\tiny$\pm$3.0} & 39.7{\tiny$\pm$1.3} \\
GRU + R & 10 & 70.0{\tiny$\pm$5.0} & 37.4{\tiny$\pm$3.3} & 33.9{\tiny$\pm$4.5} & 26.9{\tiny$\pm$2.9} & 73.5{\tiny$\pm$0.9} & 52.5{\tiny$\pm$2.0} & 46.9{\tiny$\pm$3.2} & 42.8{\tiny$\pm$1.2} \\
GRU + PT & 10 & 58.5{\tiny$\pm$5.1} & \best{39.2}{\tiny$\pm$3.3} & 33.0{\tiny$\pm$4.2} & \best{31.7}{\tiny$\pm$2.8} & 58.6{\tiny$\pm$1.1} & 40.8{\tiny$\pm$1.8} & 35.4{\tiny$\pm$3.0} & 44.0{\tiny$\pm$1.2} \\
GRU + PT + R (ours) & 10 & 69.2{\tiny$\pm$5.0} & 38.8{\tiny$\pm$3.5} & \best{37.2}{\tiny$\pm$4.6} & 30.6{\tiny$\pm$2.9} & 73.7{\tiny$\pm$0.9} & \best{52.9}{\tiny$\pm$2.1} & \best{48.5}{\tiny$\pm$3.3} & \best{44.3}{\tiny$\pm$1.2} \\
\bottomrule
\end{tabular}
\end{adjustbox}
\end{table*}

Table~\ref{tab:app-task1-baselines} extends Table~\ref{tab:main-tasks} to all 37 methods. \emph{Frequency and Markov
models at the cell level} never exceed $10\%$ macro-F1: under the \texttt{CONT} tokenisation they cannot copy the previous
metre, and a first-order chain carries no stratigraphic memory. \emph{Spatial methods} are strong only where training
holes are within reach: $k$-NN, indicator kriging and SIS reach $66$--$72\%$ macro-F1 on \textsc{test} and fall to
$1$--$5\%$ on every shifted split when restricted to the training pool; giving them the same archive pool that our
retrieval uses lifts them only to $6$--$12\%$, so the advantage of the proposed recipe is not access to a larger pool.
\emph{Directly conditioned neural baselines} expose the transfer boundary: adding location to an LSTM raises \textsc{test}
top-1 from $60.1$ to $80.5$ and lowers \textsc{pilbara-test} from $38.5$ to $22.6$; for a GRU the same input gives
$79.6$ in distribution and $3.0$ across provinces. Geochemistry alone is milder ($67.7$ and $23.2$ for the GRU), and the
two together are the strongest in-distribution models of the benchmark (up to $81.8$ top-1) and among the weakest under
cross-province transfer. Spatial baselines and the collar-conditioned networks were re-fitted on the coordinates;
sequence-only baselines are unaffected by the coordinate revision because the training split is unchanged.

\subsubsection{Paired differences}
\begin{table*}[t]
\centering
\footnotesize
\setlength{\tabcolsep}{3pt}
\caption{\textbf{Task~1 paired differences} (percentage points) with 95\% intervals. Both methods are scored on the same hole-cluster bootstrap resample ($B{=}1000$), seeds resampled independently. Intervals that exclude zero are bold. GRU and layer Transformer: the plain models with label smoothing ($\varepsilon{=}0.3$).}
\label{tab:app-task1-paired}
\begin{adjustbox}{width=\textwidth}
\begin{tabular}{@{}l rrrr rrrr@{}}
\toprule
& \multicolumn{4}{c}{\textbf{$\Delta$ macro-F1}} & \multicolumn{4}{c}{\textbf{$\Delta$ top-1}} \\
\cmidrule(lr){2-5}\cmidrule(lr){6-9}
\textbf{Comparison} & Test & OOD-R & OOD-D & Pil-T & Test & OOD-R & OOD-D & Pil-T \\
\midrule
GRU + R $-$ GRU & \textbf{+10.7{\tiny\,[+8.9, +12.7]}} & -0.1{\tiny\,[-1.0, +2.2]} & \textbf{+5.6{\tiny\,[+4.3, +6.9]}} & -1.3{\tiny\,[-3.1, +0.6]} & \textbf{+15.0{\tiny\,[+14.2, +15.8]}} & \textbf{+12.0{\tiny\,[+10.9, +14.1]}} & \textbf{+12.2{\tiny\,[+8.4, +16.1]}} & +0.8{\tiny\,[-0.0, +1.7]} \\
GRU + PT $-$ GRU & -0.9{\tiny\,[-2.0, +0.3]} & \textbf{+1.8{\tiny\,[+1.1, +3.2]}} & \textbf{+4.7{\tiny\,[+3.4, +5.8]}} & \textbf{+3.5{\tiny\,[+2.2, +4.8]}} & +0.0{\tiny\,[-0.2, +0.3]} & +0.3{\tiny\,[-0.7, +1.5]} & +0.7{\tiny\,[-2.7, +3.6]} & \textbf{+2.0{\tiny\,[+1.0, +3.0]}} \\
Ours $-$ GRU & \textbf{+9.9{\tiny\,[+7.8, +13.1]}} & \textbf{+1.4{\tiny\,[+0.4, +4.0]}} & \textbf{+8.9{\tiny\,[+7.5, +10.6]}} & \textbf{+2.4{\tiny\,[+1.0, +4.3]}} & \textbf{+15.1{\tiny\,[+14.3, +16.0]}} & \textbf{+12.4{\tiny\,[+11.2, +14.4]}} & \textbf{+13.8{\tiny\,[+10.4, +17.6]}} & \textbf{+2.4{\tiny\,[+1.5, +3.3]}} \\
Ours $-$ GRU + R \quad(adds PT) & -0.8{\tiny\,[-1.8, +2.9]} & \textbf{+1.4{\tiny\,[+0.4, +2.6]}} & \textbf{+3.4{\tiny\,[+2.1, +4.6]}} & \textbf{+3.7{\tiny\,[+2.2, +5.4]}} & +0.2{\tiny\,[-0.2, +0.5]} & +0.4{\tiny\,[-0.8, +1.1]} & +1.6{\tiny\,[-0.9, +4.9]} & \textbf{+1.5{\tiny\,[+1.0, +2.1]}} \\
Ours $-$ GRU + PT \quad(adds R) & \textbf{+10.8{\tiny\,[+9.2, +13.1]}} & -0.4{\tiny\,[-1.8, +1.6]} & \textbf{+4.2{\tiny\,[+3.3, +5.5]}} & -1.1{\tiny\,[-2.2, +0.4]} & \textbf{+15.1{\tiny\,[+14.2, +16.0]}} & \textbf{+12.1{\tiny\,[+10.6, +14.1]}} & \textbf{+13.1{\tiny\,[+10.0, +17.1]}} & +0.3{\tiny\,[-0.5, +1.0]} \\
Ours $-$ Layer Transformer & \textbf{+10.4{\tiny\,[+8.4, +13.5]}} & \textbf{+2.2{\tiny\,[+0.9, +5.3]}} & \textbf{+8.1{\tiny\,[+6.5, +9.7]}} & \textbf{+4.6{\tiny\,[+3.4, +6.6]}} & \textbf{+15.4{\tiny\,[+14.5, +16.3]}} & \textbf{+12.7{\tiny\,[+11.6, +14.8]}} & \textbf{+13.5{\tiny\,[+10.1, +17.4]}} & \textbf{+4.6{\tiny\,[+3.7, +5.5]}} \\
Ours $-$ Layer Markov & \textbf{+12.5{\tiny\,[+10.1, +15.7]}} & \textbf{+2.5{\tiny\,[+1.8, +5.7]}} & \textbf{+11.1{\tiny\,[+9.0, +13.2]}} & \textbf{+8.7{\tiny\,[+4.9, +11.5]}} & \textbf{+16.7{\tiny\,[+15.7, +17.8]}} & \textbf{+15.5{\tiny\,[+13.8, +17.8]}} & \textbf{+15.8{\tiny\,[+13.8, +20.1]}} & \textbf{+7.9{\tiny\,[+6.7, +8.9]}} \\
Ours $-$ cell Transformer & \textbf{+8.9{\tiny\,[+6.9, +12.1]}} & \textbf{+12.3{\tiny\,[+8.9, +14.3]}} & \textbf{+13.6{\tiny\,[+11.0, +15.8]}} & \textbf{+8.1{\tiny\,[+6.1, +12.4]}} & \textbf{+13.0{\tiny\,[+12.1, +14.1]}} & \textbf{+13.3{\tiny\,[+11.3, +15.6]}} & \textbf{+13.7{\tiny\,[+11.1, +18.1]}} & \textbf{+6.2{\tiny\,[+5.2, +7.2]}} \\
Ours $-$ Spatial $k$-NN (archive) & +0.3{\tiny\,[-7.1, +4.0]} & \textbf{+27.3{\tiny\,[+24.4, +30.3]}} & \textbf{+27.6{\tiny\,[+23.2, +30.9]}} & \textbf{+22.6{\tiny\,[+20.7, +27.2]}} & \textbf{-2.0{\tiny\,[-3.1, -0.8]}} & \textbf{+28.0{\tiny\,[+25.1, +29.4]}} & \textbf{+25.1{\tiny\,[+22.0, +29.1]}} & \textbf{+24.5{\tiny\,[+23.2, +25.9]}} \\
Ours $-$ XGBoost & +4.4{\tiny\,[-7.2, +8.2]} & \textbf{+36.4{\tiny\,[+33.4, +40.2]}} & \textbf{+34.8{\tiny\,[+29.9, +38.7]}} & \textbf{+28.4{\tiny\,[+27.0, +32.9]}} & \textbf{-2.4{\tiny\,[-3.5, -1.3]}} & \textbf{+33.7{\tiny\,[+31.1, +35.9]}} & \textbf{+37.0{\tiny\,[+33.4, +42.1]}} & \textbf{+32.9{\tiny\,[+31.5, +34.0]}} \\
GRU + location $-$ GRU \quad(cell) & \textbf{+14.7{\tiny\,[+12.3, +17.7]}} & \textbf{-8.4{\tiny\,[-10.7, -7.2]}} & \textbf{-19.9{\tiny\,[-22.7, -16.4]}} & \textbf{-22.7{\tiny\,[-24.5, -20.1]}} & \textbf{+19.3{\tiny\,[+18.3, +20.5]}} & -2.7{\tiny\,[-5.0, +1.0]} & \textbf{-23.4{\tiny\,[-26.0, -20.0]}} & \textbf{-33.3{\tiny\,[-34.8, -31.8]}} \\
GRU + geochem. $-$ GRU \quad(cell) & \textbf{+6.8{\tiny\,[+4.2, +9.1]}} & -1.4{\tiny\,[-3.1, +0.0]} & \textbf{-5.3{\tiny\,[-6.9, -3.4]}} & \textbf{-12.2{\tiny\,[-13.5, -8.6]}} & \textbf{+7.4{\tiny\,[+6.5, +8.2]}} & +1.1{\tiny\,[-1.1, +3.8]} & \textbf{-4.7{\tiny\,[-7.3, -1.9]}} & \textbf{-13.1{\tiny\,[-14.4, -11.9]}} \\
GRU + both $-$ GRU \quad(cell) & \textbf{+16.3{\tiny\,[+13.3, +21.5]}} & \textbf{-6.1{\tiny\,[-8.4, -4.8]}} & \textbf{-19.3{\tiny\,[-22.0, -15.8]}} & \textbf{-20.7{\tiny\,[-22.3, -18.1]}} & \textbf{+20.1{\tiny\,[+19.0, +21.3]}} & \textbf{-4.9{\tiny\,[-7.6, -2.2]}} & \textbf{-23.1{\tiny\,[-25.6, -20.4]}} & \textbf{-25.0{\tiny\,[-26.7, -23.2]}} \\
LSTM + location $-$ LSTM \quad(cell) & \textbf{+14.5{\tiny\,[+12.9, +20.8]}} & \textbf{-14.2{\tiny\,[-17.3, -13.1]}} & \textbf{-11.7{\tiny\,[-13.6, -9.0]}} & \textbf{-14.8{\tiny\,[-18.6, -13.3]}} & \textbf{+20.4{\tiny\,[+19.3, +21.5]}} & \textbf{-16.0{\tiny\,[-17.7, -14.2]}} & \textbf{-11.3{\tiny\,[-12.5, -7.6]}} & \textbf{-16.0{\tiny\,[-17.3, -14.6]}} \\
\bottomrule
\end{tabular}
\end{adjustbox}
\end{table*}

Table~\ref{tab:app-task1-paired} scores both methods of each row on the same bootstrap resample of holes.
\emph{Components.} Relative to the plain layer-level GRU, which is trained with label smoothing like every arm, retrieval adds $+10.7$ macro-F1 points on \textsc{test} and $+5.6$ on OOD-D, and nothing
measurable on OOD-R ($-0.1$) or \textsc{pilbara-test} ($-1.3$); both intervals include zero.
Pretraining shows the complementary profile: no effect in distribution ($-0.9$, not significant) and $+1.8$, $+4.7$ and
$+3.5$ on the three shifted splits. The \textsc{pilbara-test} gain from pretraining does not come from Pilbara holes in
the archive: it persists when the whole province is removed from the pretraining corpus
(App.~\ref{app:province_holdout}). Combined, the full recipe improves on the plain GRU on all four splits, on layer
Markov, on the layer-level Transformer and on the cell-level Transformer, with every interval excluding zero; against the strongest spatial baselines it
is on par in distribution and $+23$ to $+36$ points ahead under shift. On OOD-R and \textsc{pilbara-test} the full
recipe is numerically below GRU\,+\,PT in macro-F1 ($38.8$ vs.\ $39.2$ and $30.6$ vs.\ $31.7$), but the paired
differences ($-0.4$ and $-1.1$) are not significant: once the model is pretrained, retrieval neither helps nor hurts
there. Under top-1, balanced accuracy, support-weighted F1 and Cohen's $\kappa$ the full recipe is the best row on all
three shifted splits (e.g.\ $\kappa{\times}100 = 45.8 / 40.3 / 37.6$ vs.\ $32.5 / 29.9 / 34.6$ for GRU\,+\,PT); retrieval
thus helps mostly on frequent lithologies, and pretraining is what moves the rare ones.
\emph{The transfer boundary.} The last rows make the wall a paired statement on identical holes: adding location to the
cell-level GRU changes macro-F1 by $+14.7$ on \textsc{test} and by $-8.4$, $-19.9$ and $-22.7$ on OOD-R, OOD-D and
\textsc{pilbara-test}; geochemistry by $+6.8$ and $-1.4$, $-5.3$, $-12.2$; an LSTM shows the same sign pattern. Every
interval except geochemistry on OOD-R excludes zero, so the reversal is not a property of one backbone or of seed noise.

\subsubsection{Backbones and token granularity}
\begin{table*}[t]
\centering
\footnotesize
\setlength{\tabcolsep}{3pt}
\caption{\textbf{Task~1 across layer-level backbones.} Each backbone is trained plain (lithology only), with retrieval (R), with fine-objective pretraining (PT), and with both; all four arms use label smoothing ($\varepsilon{=}0.3$). $\pm$: half-width of the 95\% hole-cluster bootstrap interval ($B{=}1000$); trained rows also resample seeds. macro-F1 over classes with $\geq$20 events; top-1 on OOD-R / OOD-D is macro-averaged over exploration reports.}
\label{tab:app-task1-backbones}
\begin{adjustbox}{width=\textwidth}
\begin{tabular}{@{}l c rrrr rrrr@{}}
\toprule
& & \multicolumn{4}{c}{\textbf{macro-F1 (\%)}} & \multicolumn{4}{c}{\textbf{top-1 accuracy (\%)}} \\
\cmidrule(lr){3-6}\cmidrule(lr){7-10}
\textbf{Method} & \textbf{runs} & Test & OOD-R & OOD-D & Pil-T & Test & OOD-R & OOD-D & Pil-T \\
\midrule
\rowcolor{unitgray}
\multicolumn{10}{@{}l}{\textbf{GRU}} \\
GRU & 10 & 59.4{\tiny$\pm$4.9} & 37.4{\tiny$\pm$3.5} & 28.3{\tiny$\pm$3.9} & 28.2{\tiny$\pm$3.0} & 58.6{\tiny$\pm$1.1} & 40.5{\tiny$\pm$1.9} & 34.7{\tiny$\pm$3.1} & 42.0{\tiny$\pm$1.4} \\
\quad + R & 10 & \best{70.0}{\tiny$\pm$5.0} & 37.4{\tiny$\pm$3.3} & 33.9{\tiny$\pm$4.5} & 26.9{\tiny$\pm$2.8} & 73.5{\tiny$\pm$1.0} & 52.5{\tiny$\pm$1.9} & 46.9{\tiny$\pm$3.4} & 42.8{\tiny$\pm$1.2} \\
\quad + PT & 10 & 58.5{\tiny$\pm$5.1} & 39.2{\tiny$\pm$3.2} & 33.0{\tiny$\pm$4.2} & \best{31.7}{\tiny$\pm$2.8} & 58.6{\tiny$\pm$1.1} & 40.8{\tiny$\pm$1.9} & 35.4{\tiny$\pm$3.0} & 44.0{\tiny$\pm$1.3} \\
\quad + PT + R & 10 & 69.2{\tiny$\pm$4.9} & 38.8{\tiny$\pm$3.5} & 37.2{\tiny$\pm$4.7} & 30.6{\tiny$\pm$2.9} & 73.7{\tiny$\pm$0.9} & \best{52.9}{\tiny$\pm$2.1} & 48.5{\tiny$\pm$3.3} & 44.3{\tiny$\pm$1.2} \\
\midrule
\rowcolor{unitgray}
\multicolumn{10}{@{}l}{\textbf{LSTM}} \\
LSTM & 3 & 59.1{\tiny$\pm$5.1} & 37.4{\tiny$\pm$4.0} & 28.1{\tiny$\pm$4.1} & 25.0{\tiny$\pm$3.7} & 58.5{\tiny$\pm$1.1} & 41.2{\tiny$\pm$2.2} & 36.0{\tiny$\pm$4.6} & 39.8{\tiny$\pm$1.8} \\
\quad + R & 3 & 69.5{\tiny$\pm$5.3} & 37.3{\tiny$\pm$3.9} & 33.8{\tiny$\pm$4.9} & 25.0{\tiny$\pm$3.8} & 73.6{\tiny$\pm$0.9} & 52.3{\tiny$\pm$2.0} & 48.1{\tiny$\pm$4.0} & 41.5{\tiny$\pm$1.8} \\
\quad + PT & 3 & 58.2{\tiny$\pm$5.3} & 38.9{\tiny$\pm$3.4} & 30.5{\tiny$\pm$4.2} & 27.6{\tiny$\pm$3.1} & 58.5{\tiny$\pm$1.1} & 41.5{\tiny$\pm$2.3} & 36.4{\tiny$\pm$4.3} & 43.2{\tiny$\pm$1.7} \\
\quad + PT + R & 3 & 67.9{\tiny$\pm$5.6} & 38.5{\tiny$\pm$3.2} & 34.6{\tiny$\pm$4.6} & 29.1{\tiny$\pm$3.5} & 73.6{\tiny$\pm$1.0} & 52.6{\tiny$\pm$2.1} & 47.7{\tiny$\pm$4.0} & \best{45.5}{\tiny$\pm$1.4} \\
\midrule
\rowcolor{unitgray}
\multicolumn{10}{@{}l}{\textbf{Transformer}} \\
Transformer & 10 & 58.9{\tiny$\pm$5.0} & 36.6{\tiny$\pm$3.6} & 29.1{\tiny$\pm$3.9} & 26.0{\tiny$\pm$3.0} & 58.3{\tiny$\pm$1.1} & 40.2{\tiny$\pm$1.8} & 35.1{\tiny$\pm$3.1} & 39.7{\tiny$\pm$1.3} \\
\quad + R & 3 & 69.8{\tiny$\pm$4.8} & 36.6{\tiny$\pm$3.8} & 34.8{\tiny$\pm$4.6} & 24.0{\tiny$\pm$3.5} & \best{73.7}{\tiny$\pm$1.0} & 52.0{\tiny$\pm$2.2} & 48.4{\tiny$\pm$4.5} & 39.3{\tiny$\pm$1.4} \\
\quad + PT & 3 & 57.5{\tiny$\pm$5.2} & 38.8{\tiny$\pm$3.3} & 30.8{\tiny$\pm$3.8} & 28.1{\tiny$\pm$2.9} & 58.1{\tiny$\pm$1.1} & 40.9{\tiny$\pm$2.1} & 36.4{\tiny$\pm$4.2} & 40.7{\tiny$\pm$1.3} \\
\quad + PT + R & 3 & 68.2{\tiny$\pm$5.3} & \best{39.6}{\tiny$\pm$3.3} & 34.9{\tiny$\pm$4.1} & 26.0{\tiny$\pm$3.8} & 73.4{\tiny$\pm$1.0} & 52.5{\tiny$\pm$2.3} & 48.3{\tiny$\pm$4.9} & 41.8{\tiny$\pm$1.3} \\
\midrule
\rowcolor{unitgray}
\multicolumn{10}{@{}l}{\textbf{TCN}} \\
TCN & 3 & 58.4{\tiny$\pm$4.9} & 32.0{\tiny$\pm$2.7} & 30.1{\tiny$\pm$4.1} & 27.9{\tiny$\pm$2.7} & 58.1{\tiny$\pm$1.1} & 40.3{\tiny$\pm$2.1} & 35.0{\tiny$\pm$4.6} & 43.5{\tiny$\pm$1.3} \\
\quad + R & 3 & 69.4{\tiny$\pm$4.9} & 33.0{\tiny$\pm$3.2} & 32.3{\tiny$\pm$4.3} & 26.4{\tiny$\pm$2.8} & 73.5{\tiny$\pm$1.0} & 52.6{\tiny$\pm$2.1} & 47.2{\tiny$\pm$4.4} & 42.8{\tiny$\pm$1.3} \\
\quad + PT & 3 & 57.7{\tiny$\pm$4.9} & 36.0{\tiny$\pm$3.5} & 33.2{\tiny$\pm$4.0} & 29.9{\tiny$\pm$2.5} & 58.0{\tiny$\pm$1.1} & 41.7{\tiny$\pm$2.2} & 37.7{\tiny$\pm$4.7} & 42.1{\tiny$\pm$1.4} \\
\quad + PT + R & 3 & 68.7{\tiny$\pm$4.9} & 37.7{\tiny$\pm$3.5} & \best{37.9}{\tiny$\pm$4.7} & 29.2{\tiny$\pm$2.3} & 73.3{\tiny$\pm$1.0} & 52.7{\tiny$\pm$2.2} & \best{48.7}{\tiny$\pm$4.3} & 43.7{\tiny$\pm$1.3} \\
\midrule
\rowcolor{unitgray}
\multicolumn{10}{@{}l}{\textbf{Mamba}} \\
Mamba & 3 & 58.0{\tiny$\pm$4.9} & 28.4{\tiny$\pm$3.5} & 25.5{\tiny$\pm$3.6} & 19.5{\tiny$\pm$3.2} & 57.9{\tiny$\pm$1.1} & 41.2{\tiny$\pm$2.2} & 36.9{\tiny$\pm$2.6} & 33.6{\tiny$\pm$2.4} \\
\quad + R & 3 & 68.2{\tiny$\pm$5.5} & 28.6{\tiny$\pm$3.7} & 29.6{\tiny$\pm$4.5} & 21.2{\tiny$\pm$2.1} & 73.2{\tiny$\pm$1.0} & 51.3{\tiny$\pm$2.2} & 46.4{\tiny$\pm$4.5} & 37.0{\tiny$\pm$3.0} \\
\quad + PT & 3 & 57.5{\tiny$\pm$4.9} & 29.3{\tiny$\pm$3.2} & 27.3{\tiny$\pm$3.8} & 23.7{\tiny$\pm$2.8} & 57.8{\tiny$\pm$1.1} & 40.3{\tiny$\pm$2.4} & 35.2{\tiny$\pm$4.6} & 38.1{\tiny$\pm$1.2} \\
\quad + PT + R & 3 & 67.6{\tiny$\pm$5.2} & 30.9{\tiny$\pm$3.3} & 30.5{\tiny$\pm$4.1} & 22.6{\tiny$\pm$2.7} & 73.1{\tiny$\pm$1.0} & 51.1{\tiny$\pm$2.3} & 46.7{\tiny$\pm$4.1} & 39.6{\tiny$\pm$1.4} \\
\bottomrule
\end{tabular}
\end{adjustbox}
\end{table*}

Table~\ref{tab:app-task1-backbones} repeats the four arms for five layer-level backbones, each with a label-smoothed
plain arm. The pattern of Figure~\ref{fig:transfer}(c) holds for each of them: in macro-F1, retrieval adds $10$--$11$
points on \textsc{test} and nothing measurable on \textsc{pilbara-test} ($-2.0$ to $+1.7$), while pretraining adds nothing
on \textsc{test} and $+2.1$ to $+4.2$ on \textsc{pilbara-test}. With the full
recipe GRU, LSTM, TCN and Transformer are within a few points of each other; Mamba is the weakest. We use the GRU in the
main text because it is the most even across the two tasks; the conclusions do not depend on this choice. All backbones
share one learning rate ($3{\times}10^{-4}$). The same recipe applied to a \emph{cell-level} Transformer behaves
differently: retrieval transfers ($+10.3$ macro-F1 on \textsc{test}), but pretraining lowers macro-F1 on three of the
four splits ($60.3 \to 53.3$ on \textsc{test}, $23.6 \to 19.4$ on OOD-D, $22.5 \to 19.1$ on \textsc{pilbara-test}), and
the full cell-level recipe is $10$--$14$ points below the layer-level one on the three shifted splits. At one-metre
resolution most targets are \texttt{CONT}, so the pretraining signal is dominated by thickness rather than by lithological
succession; the layer-level formulation removes that imbalance, which is why the recipe is built on it.

\subsubsection{Cover and bedrock}
\begin{table}[t]
\centering
\small
\setlength{\tabcolsep}{4pt}
\caption{\textbf{Cover vs.\ bedrock} (layer-level GRU, top-1 \%, pooled inside each zone by the \emph{true} next
lithology; 10 seeds). $\Delta$R: retrieval gain over the plain model (with label smoothing, as every arm); $\Delta$PT: pretraining gain without retrieval.
$n$: scored layer starts.}
\label{tab:app-task1-zones}
\begin{tabular}{@{}ll r rrrr rr@{}}
\toprule
Split & Zone & $n$ & plain & +R & +PT & +PT+R & $\Delta$R & $\Delta$PT \\
\midrule
\textsc{test} & cover & 5{,}870 & 59.7 & 75.7 & 59.7 & 76.0 & $+16.0$ & $+0.0$ \\
              & bedrock & 4{,}539 & 57.1 & 70.7 & 57.2 & 70.7 & $+13.6$ & $+0.1$ \\
OOD-R & cover & 26{,}510 & 43.2 & 58.6 & 43.9 & 59.3 & $+15.3$ & $+0.7$ \\
              & bedrock & 4{,}100 & 53.3 & 54.0 & 54.7 & 56.6 & $+0.7$ & $+1.4$ \\
OOD-D & cover & 7{,}380 & 36.7 & 47.6 & 41.0 & 52.3 & $+10.9$ & $+4.3$ \\
              & bedrock & 2{,}233 & 34.1 & 39.1 & 34.6 & 39.7 & $+4.9$ & $+0.5$ \\
\textsc{pilbara-test} & cover & 4{,}214 & 43.6 & 44.4 & 44.8 & 44.8 & $+0.8$ & $+1.2$ \\
              & bedrock & 5{,}724 & 40.7 & 41.6 & 43.4 & 44.0 & $+0.8$ & $+2.7$ \\
\bottomrule
\end{tabular}
\end{table}

Table~\ref{tab:app-task1-zones} gives the numbers behind Figure~\ref{fig:transfer}(d). Retrieval is a regolith
mechanism: within the Eastern Goldfields its gain in cover is $+16.0$, $+15.3$ and $+10.9$ points, against $+13.6$,
$+0.7$ and $+4.9$ in bedrock, and it all but vanishes across provinces ($+0.8$ in both zones). Pretraining is the mirror
image on the two splits where bedrock differs most from training: $+1.4$ vs.\ $+0.7$ on OOD-R and $+2.7$ vs.\ $+1.2$ on
\textsc{pilbara-test}, with no effect in distribution. OOD-D is the exception and we report it as such: there the
pretraining gain is larger in cover ($+4.3$) than in bedrock ($+0.5$); the distal ring is $77\%$ regolith by scored layer
starts, and a plausible reading is that its weathering profiles differ from those of the training camps, so that the
archive's regolith sequences are what transfers; we have not tested this directly.

\subsubsection{Error analysis}
\begin{figure*}[t]
\centering
\includegraphics[width=\textwidth]{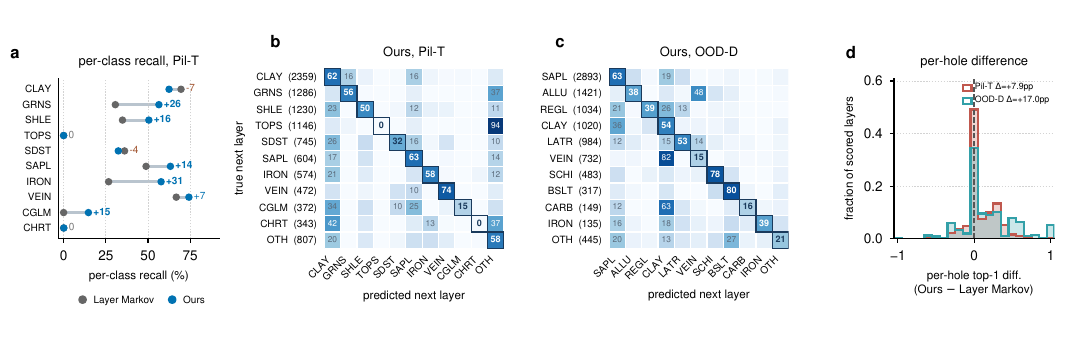}
\caption{\textbf{Error analysis} (full recipe, 10 seeds pooled; row counts are per run).
\textbf{(a)} Per-class recall on \textsc{pilbara-test} for the ten most frequent true classes, layer Markov
$\rightarrow$ ours. \textbf{(b, c)} Row-normalised confusion on \textsc{pilbara-test} and OOD-D.
\textbf{(d)} Distribution of the per-hole top-1 difference, ours minus layer Markov, weighted by scored layers.}
\label{fig:app-task1-error}
\end{figure*}
Figure~\ref{fig:app-task1-error}(a) shows that the \textsc{pilbara-test} margin over layer Markov is carried by bedrock
classes that are rare in the Eastern Goldfields training region: ironstone/BIF ($+31$ recall points), greenstone
($+26$), shale ($+16$) and conglomerate ($0 \to 15$), together with saprolite ($+14$). Two classes remain unsolved by
every method: transported overburden (\textsc{tops}) is $12\%$ of Pilbara layer starts and essentially absent from the
training region, and chert is absorbed by clay and the residual classes. On OOD-D (panel c) the dominant errors are
within the regolith profile (alluvium $\to$ laterite $48\%$, regolith $\to$ clay $26\%$) and quartz vein $\to$ clay
($82\%$), i.e.\ confusions between units that loggers themselves separate inconsistently, rather than failures of
stratigraphic order. Panel (d) shows that the gain is broad: on OOD-D $54\%$ of the scored layers sit in holes where the
recipe beats layer Markov and $14\%$ in holes where it loses; on \textsc{pilbara-test} the split is $41\%$ to $13\%$.
By exploration report, the natural unit of a project, a logging geologist and a local vocabulary, the recipe beats layer
Markov in 39 of 44 OOD-D reports (event-weighted margin $+17.3$ points); on \textsc{pilbara-test} it wins 20 and loses
18 of 43 reports, and the $+7.9$-point margin is carried by a few large reports (up to $+41$ points; $+23$ on the
largest, with 707 layer starts), while most losses are within $5$ points. Cross-province transfer therefore improves the
typical large project but is not yet uniform across reports.

\subsection{Additional Results: Free-Running Generation}
\label{app:results_task2}

\subsubsection{Protocol and uncertainty}
Every method generates the lower half of the same report-stratified sample of holes ($\approx$600 per split, drawn so that
every exploration report is represented in proportion to its size rather than by file order), from the true upper half,
with $M{=}5$ rollouts per hole. Generated and true suffixes are cut into layers and compared through their joint
(lithology, thickness) law over the 37 codes and seven log-spaced thickness bins; the pooled forward KL of this law is
the main-text metric (App.~\ref{app:evaluation} defines it and the secondary metrics). Because a divergence is a
function of sums of per-hole histograms, it can be bootstrapped: intervals are $95\%$ report-cluster bootstrap intervals
($B{=}1000$; all holes of a WAMEX report are resampled together, since they share a camp, a logger and a vocabulary), and
trained rows additionally resample seeds. Plug-in divergences are biased upward under resampling, so intervals are
re-centred on the full-sample estimate. One report resample per split is shared by all rows, so every difference in
Table~\ref{tab:app-task2-paired} is paired. The intervals are wide on OOD-D, where one camp holds more than half of the
holes and the resample often drops or doubles it.

\subsubsection{All baselines}
\begin{table*}[t]
\centering
\footnotesize
\setlength{\tabcolsep}{3pt}
\caption{\textbf{Task~2, all baselines.} Pooled joint KL, JSD and per-metre agreement. $^{*}$: lookup, not free-running; not highlighted. $\pm$: half-width of the 95\% report-cluster bootstrap interval ($B{=}1000$; holes of one exploration report are resampled together); trained rows also resample seeds. Plug-in divergences are biased upward under resampling, so intervals are re-centred on the full-sample estimate. Layer-level neural rows (and the plain arms of the backbone table) use label smoothing ($\varepsilon{=}0.3$).}
\label{tab:app-task2-baselines}
\begin{adjustbox}{max width=\textwidth}
\begin{tabular}{@{}l c rrrr rrrr rrrr@{}}
\toprule
& & \multicolumn{4}{c}{\textbf{KL} $\downarrow$} & \multicolumn{4}{c}{\textbf{JSD} $\downarrow$} & \multicolumn{4}{c}{\textbf{Acc/m} $\uparrow$} \\
\cmidrule(lr){3-6} \cmidrule(lr){7-10} \cmidrule(lr){11-14}
\textbf{Method} & \textbf{runs} & Test & OOD-R & OOD-D & Pil-T & Test & OOD-R & OOD-D & Pil-T & Test & OOD-R & OOD-D & Pil-T \\
\midrule
\rowcolor{unitgray}
\multicolumn{14}{@{}l}{\textbf{Frequency / Markov (cell level)}} \\
Unigram & 1 & 0.226{\tiny$\pm$0.077} & 0.599{\tiny$\pm$0.149} & 0.872{\tiny$\pm$0.476} & 1.111{\tiny$\pm$0.316} & 0.047{\tiny$\pm$0.013} & 0.130{\tiny$\pm$0.024} & 0.133{\tiny$\pm$0.050} & 0.203{\tiny$\pm$0.041} & 50.7{\tiny$\pm$2.6} & 48.2{\tiny$\pm$4.3} & 51.5{\tiny$\pm$5.3} & 41.9{\tiny$\pm$4.9} \\
Per-region unigram & 1 & 0.226{\tiny$\pm$0.077} & 0.599{\tiny$\pm$0.149} & 0.872{\tiny$\pm$0.476} & 1.111{\tiny$\pm$0.316} & 0.047{\tiny$\pm$0.013} & 0.130{\tiny$\pm$0.024} & 0.133{\tiny$\pm$0.050} & 0.203{\tiny$\pm$0.041} & 50.7{\tiny$\pm$2.6} & 48.2{\tiny$\pm$4.3} & 51.5{\tiny$\pm$5.3} & 41.9{\tiny$\pm$4.9} \\
Persistence & 1 & 1.318{\tiny$\pm$0.303} & 1.848{\tiny$\pm$0.487} & 2.959{\tiny$\pm$1.397} & 2.112{\tiny$\pm$0.551} & 0.213{\tiny$\pm$0.020} & 0.286{\tiny$\pm$0.034} & 0.301{\tiny$\pm$0.075} & 0.285{\tiny$\pm$0.032} & 67.9{\tiny$\pm$3.3} & \best{64.8}{\tiny$\pm$5.4} & \best{66.3}{\tiny$\pm$6.3} & \best{57.3}{\tiny$\pm$6.4} \\
Bigram & 1 & 0.184{\tiny$\pm$0.052} & 0.489{\tiny$\pm$0.149} & 0.639{\tiny$\pm$0.411} & 0.949{\tiny$\pm$0.298} & 0.042{\tiny$\pm$0.011} & 0.104{\tiny$\pm$0.022} & 0.111{\tiny$\pm$0.048} & 0.179{\tiny$\pm$0.037} & 46.2{\tiny$\pm$2.6} & 43.7{\tiny$\pm$4.3} & 47.5{\tiny$\pm$5.1} & 37.4{\tiny$\pm$5.1} \\
$3$-gram & 1 & 0.186{\tiny$\pm$0.067} & 0.533{\tiny$\pm$0.190} & 0.781{\tiny$\pm$0.498} & 0.960{\tiny$\pm$0.285} & 0.040{\tiny$\pm$0.012} & 0.110{\tiny$\pm$0.021} & 0.110{\tiny$\pm$0.046} & 0.175{\tiny$\pm$0.036} & 48.9{\tiny$\pm$2.6} & 47.5{\tiny$\pm$4.3} & 50.9{\tiny$\pm$5.1} & 38.8{\tiny$\pm$5.1} \\
Depth-Markov & 1 & 0.220{\tiny$\pm$0.062} & 0.518{\tiny$\pm$0.125} & 0.564{\tiny$\pm$0.335} & 0.797{\tiny$\pm$0.223} & 0.058{\tiny$\pm$0.012} & 0.122{\tiny$\pm$0.021} & 0.116{\tiny$\pm$0.045} & 0.176{\tiny$\pm$0.032} & 48.7{\tiny$\pm$2.6} & 46.4{\tiny$\pm$4.3} & 49.1{\tiny$\pm$5.0} & 38.2{\tiny$\pm$5.3} \\
HMM ($K{=}24$) & 1 & 0.226{\tiny$\pm$0.073} & 0.584{\tiny$\pm$0.143} & 0.774{\tiny$\pm$0.456} & 1.059{\tiny$\pm$0.299} & 0.047{\tiny$\pm$0.012} & 0.126{\tiny$\pm$0.025} & 0.130{\tiny$\pm$0.049} & 0.203{\tiny$\pm$0.041} & 49.9{\tiny$\pm$2.6} & 47.5{\tiny$\pm$4.4} & 50.5{\tiny$\pm$5.2} & 41.0{\tiny$\pm$5.0} \\
\midrule
\rowcolor{unitgray}
\multicolumn{14}{@{}l}{\textbf{Geostatistical (cell level)}} \\
Spatial $k$-NN, train pool$^{*}$ & 1 & 0.546{\tiny$\pm$0.052} & 1.445{\tiny$\pm$0.394} & 2.022{\tiny$\pm$0.942} & 2.028{\tiny$\pm$0.679} & 0.161{\tiny$\pm$0.013} & 0.226{\tiny$\pm$0.031} & 0.254{\tiny$\pm$0.068} & 0.260{\tiny$\pm$0.037} & 52.4{\tiny$\pm$2.3} & 63.0{\tiny$\pm$5.4} & 64.3{\tiny$\pm$6.3} & 55.9{\tiny$\pm$6.3} \\
Spatial $k$-NN, archive pool$^{*}$ & 1 & 0.553{\tiny$\pm$0.053} & 0.950{\tiny$\pm$0.105} & 0.852{\tiny$\pm$0.352} & 0.729{\tiny$\pm$0.124} & 0.162{\tiny$\pm$0.013} & 0.262{\tiny$\pm$0.022} & 0.219{\tiny$\pm$0.047} & 0.199{\tiny$\pm$0.028} & 54.6{\tiny$\pm$2.3} & 39.3{\tiny$\pm$4.5} & 39.8{\tiny$\pm$4.3} & 33.7{\tiny$\pm$4.8} \\
Indicator kriging, train pool & 1 & 0.192{\tiny$\pm$0.066} & 1.445{\tiny$\pm$0.394} & 2.022{\tiny$\pm$0.942} & 2.028{\tiny$\pm$0.679} & 0.039{\tiny$\pm$0.010} & 0.226{\tiny$\pm$0.031} & 0.254{\tiny$\pm$0.068} & 0.260{\tiny$\pm$0.037} & 70.7{\tiny$\pm$2.6} & 63.0{\tiny$\pm$5.4} & 64.3{\tiny$\pm$6.3} & 55.9{\tiny$\pm$6.3} \\
Indicator kriging, archive pool & 1 & 0.437{\tiny$\pm$0.119} & 1.205{\tiny$\pm$0.402} & 1.991{\tiny$\pm$0.892} & 1.475{\tiny$\pm$0.633} & 0.089{\tiny$\pm$0.016} & 0.205{\tiny$\pm$0.034} & 0.249{\tiny$\pm$0.067} & 0.218{\tiny$\pm$0.038} & \best{77.8}{\tiny$\pm$2.4} & 62.9{\tiny$\pm$5.3} & 64.9{\tiny$\pm$6.0} & 55.9{\tiny$\pm$5.9} \\
SIS, train pool & 1 & 0.387{\tiny$\pm$0.038} & 0.844{\tiny$\pm$0.125} & 0.902{\tiny$\pm$0.440} & 0.806{\tiny$\pm$0.169} & 0.117{\tiny$\pm$0.009} & 0.220{\tiny$\pm$0.023} & 0.205{\tiny$\pm$0.061} & 0.197{\tiny$\pm$0.026} & 58.5{\tiny$\pm$2.4} & 47.4{\tiny$\pm$4.5} & 52.0{\tiny$\pm$5.2} & 41.4{\tiny$\pm$5.1} \\
SIS, archive pool & 1 & 0.424{\tiny$\pm$0.049} & 0.811{\tiny$\pm$0.126} & 0.890{\tiny$\pm$0.431} & 0.742{\tiny$\pm$0.150} & 0.124{\tiny$\pm$0.010} & 0.212{\tiny$\pm$0.025} & 0.203{\tiny$\pm$0.061} & 0.186{\tiny$\pm$0.026} & 62.0{\tiny$\pm$2.4} & 47.8{\tiny$\pm$4.5} & 52.2{\tiny$\pm$5.1} & 42.5{\tiny$\pm$5.1} \\
MPS-lite & 1 & 0.206{\tiny$\pm$0.083} & 0.526{\tiny$\pm$0.188} & 0.641{\tiny$\pm$0.389} & 0.968{\tiny$\pm$0.292} & 0.037{\tiny$\pm$0.011} & 0.106{\tiny$\pm$0.019} & 0.108{\tiny$\pm$0.043} & 0.172{\tiny$\pm$0.035} & 51.0{\tiny$\pm$2.7} & 50.5{\tiny$\pm$4.5} & 51.9{\tiny$\pm$5.4} & 40.4{\tiny$\pm$5.1} \\
TPROGS & 1 & 0.183{\tiny$\pm$0.051} & 0.492{\tiny$\pm$0.151} & 0.716{\tiny$\pm$0.380} & 0.885{\tiny$\pm$0.278} & 0.042{\tiny$\pm$0.011} & 0.103{\tiny$\pm$0.022} & 0.112{\tiny$\pm$0.048} & 0.177{\tiny$\pm$0.037} & 46.2{\tiny$\pm$2.6} & 43.7{\tiny$\pm$4.3} & 47.5{\tiny$\pm$5.1} & 37.4{\tiny$\pm$5.1} \\
\midrule
\rowcolor{unitgray}
\multicolumn{14}{@{}l}{\textbf{Machine learning (cell level)}} \\
XGBoost & 1 & 0.206{\tiny$\pm$0.076} & 0.518{\tiny$\pm$0.164} & 0.950{\tiny$\pm$0.475} & 1.776{\tiny$\pm$0.320} & 0.045{\tiny$\pm$0.011} & 0.119{\tiny$\pm$0.037} & 0.185{\tiny$\pm$0.059} & 0.343{\tiny$\pm$0.043} & 62.7{\tiny$\pm$2.6} & 43.2{\tiny$\pm$4.6} & 33.7{\tiny$\pm$4.1} & 27.0{\tiny$\pm$4.8} \\
\midrule
\rowcolor{unitgray}
\multicolumn{14}{@{}l}{\textbf{Deep learning, cell level, lithology only}} \\
LSTM & 1 & 0.341{\tiny$\pm$0.089} & 0.547{\tiny$\pm$0.127} & 0.529{\tiny$\pm$0.341} & 1.054{\tiny$\pm$0.398} & 0.078{\tiny$\pm$0.015} & 0.123{\tiny$\pm$0.027} & 0.096{\tiny$\pm$0.047} & 0.191{\tiny$\pm$0.045} & 69.1{\tiny$\pm$2.7} & 62.4{\tiny$\pm$5.1} & 59.0{\tiny$\pm$5.8} & 51.7{\tiny$\pm$6.1} \\
GRU & 1 & 0.340{\tiny$\pm$0.081} & 0.612{\tiny$\pm$0.136} & 0.559{\tiny$\pm$0.342} & 1.003{\tiny$\pm$0.449} & 0.078{\tiny$\pm$0.016} & 0.132{\tiny$\pm$0.025} & 0.103{\tiny$\pm$0.030} & 0.173{\tiny$\pm$0.043} & 69.4{\tiny$\pm$2.7} & 61.8{\tiny$\pm$5.2} & 60.6{\tiny$\pm$5.6} & 49.9{\tiny$\pm$6.3} \\
Mamba & 1 & 0.345{\tiny$\pm$0.077} & 0.580{\tiny$\pm$0.118} & 0.536{\tiny$\pm$0.358} & 0.900{\tiny$\pm$0.245} & 0.083{\tiny$\pm$0.014} & 0.132{\tiny$\pm$0.025} & 0.098{\tiny$\pm$0.038} & 0.185{\tiny$\pm$0.035} & 69.4{\tiny$\pm$2.7} & 64.2{\tiny$\pm$4.7} & 61.1{\tiny$\pm$5.5} & 51.7{\tiny$\pm$6.0} \\
TCN & 1 & 0.348{\tiny$\pm$0.099} & 0.683{\tiny$\pm$0.193} & 0.668{\tiny$\pm$0.397} & 0.899{\tiny$\pm$0.294} & 0.071{\tiny$\pm$0.013} & 0.143{\tiny$\pm$0.025} & 0.114{\tiny$\pm$0.036} & 0.160{\tiny$\pm$0.029} & 70.7{\tiny$\pm$2.7} & 64.5{\tiny$\pm$4.7} & 62.1{\tiny$\pm$5.7} & 54.9{\tiny$\pm$6.5} \\
Transformer & 1 & 0.356{\tiny$\pm$0.087} & 0.585{\tiny$\pm$0.114} & 0.533{\tiny$\pm$0.332} & 0.995{\tiny$\pm$0.373} & 0.079{\tiny$\pm$0.016} & 0.131{\tiny$\pm$0.022} & 0.103{\tiny$\pm$0.029} & 0.178{\tiny$\pm$0.033} & 69.7{\tiny$\pm$2.8} & 62.8{\tiny$\pm$5.1} & 59.5{\tiny$\pm$5.6} & 52.6{\tiny$\pm$6.1} \\
\midrule
\rowcolor{unitgray}
\multicolumn{14}{@{}l}{\textbf{Deep learning, cell level, conditioned on location / geochemistry}} \\
LSTM + location & 1 & 0.283{\tiny$\pm$0.083} & 0.801{\tiny$\pm$0.237} & 1.250{\tiny$\pm$0.613} & 1.511{\tiny$\pm$0.439} & 0.064{\tiny$\pm$0.015} & 0.166{\tiny$\pm$0.047} & 0.240{\tiny$\pm$0.084} & 0.252{\tiny$\pm$0.040} & 76.2{\tiny$\pm$2.4} & 55.1{\tiny$\pm$6.1} & 50.3{\tiny$\pm$5.7} & 42.8{\tiny$\pm$5.9} \\
LSTM + geochem & 1 & 0.595{\tiny$\pm$0.121} & 0.703{\tiny$\pm$0.159} & 0.783{\tiny$\pm$0.420} & 1.519{\tiny$\pm$0.469} & 0.131{\tiny$\pm$0.018} & 0.152{\tiny$\pm$0.027} & 0.136{\tiny$\pm$0.054} & 0.259{\tiny$\pm$0.038} & 65.4{\tiny$\pm$2.9} & 60.4{\tiny$\pm$5.5} & 56.6{\tiny$\pm$6.0} & 43.1{\tiny$\pm$6.3} \\
LSTM + geochem + location & 1 & 0.379{\tiny$\pm$0.095} & 0.977{\tiny$\pm$0.276} & 1.434{\tiny$\pm$0.631} & 3.082{\tiny$\pm$0.639} & 0.086{\tiny$\pm$0.016} & 0.208{\tiny$\pm$0.058} & 0.253{\tiny$\pm$0.075} & 0.442{\tiny$\pm$0.037} & 68.6{\tiny$\pm$2.9} & 51.3{\tiny$\pm$5.6} & 43.2{\tiny$\pm$6.4} & 27.9{\tiny$\pm$5.1} \\
GRU + location & 1 & 0.283{\tiny$\pm$0.075} & 0.540{\tiny$\pm$0.232} & 2.638{\tiny$\pm$0.649} & 6.016{\tiny$\pm$0.978} & 0.065{\tiny$\pm$0.016} & 0.116{\tiny$\pm$0.027} & 0.389{\tiny$\pm$0.067} & 0.499{\tiny$\pm$0.061} & 75.4{\tiny$\pm$2.6} & 56.0{\tiny$\pm$6.1} & 24.5{\tiny$\pm$6.2} & 5.9{\tiny$\pm$3.2} \\
GRU + geochem & 1 & 0.384{\tiny$\pm$0.102} & 0.772{\tiny$\pm$0.174} & 0.689{\tiny$\pm$0.424} & 1.081{\tiny$\pm$0.417} & 0.078{\tiny$\pm$0.014} & 0.161{\tiny$\pm$0.028} & 0.112{\tiny$\pm$0.046} & 0.202{\tiny$\pm$0.051} & 69.0{\tiny$\pm$3.0} & 61.4{\tiny$\pm$5.5} & 59.6{\tiny$\pm$6.0} & 50.9{\tiny$\pm$6.1} \\
GRU + geochem + location & 1 & 0.448{\tiny$\pm$0.104} & 1.350{\tiny$\pm$0.403} & 3.882{\tiny$\pm$0.904} & 4.537{\tiny$\pm$1.022} & 0.104{\tiny$\pm$0.017} & 0.274{\tiny$\pm$0.069} & 0.485{\tiny$\pm$0.049} & 0.398{\tiny$\pm$0.056} & 73.5{\tiny$\pm$2.6} & 54.5{\tiny$\pm$6.0} & 10.7{\tiny$\pm$4.4} & 18.9{\tiny$\pm$6.7} \\
Transformer + geochem + location & 1 & 0.472{\tiny$\pm$0.130} & 0.664{\tiny$\pm$0.159} & 1.360{\tiny$\pm$0.448} & 2.110{\tiny$\pm$0.396} & 0.104{\tiny$\pm$0.024} & 0.137{\tiny$\pm$0.027} & 0.247{\tiny$\pm$0.040} & 0.340{\tiny$\pm$0.034} & 72.3{\tiny$\pm$2.8} & 58.0{\tiny$\pm$6.4} & 45.3{\tiny$\pm$5.4} & 31.1{\tiny$\pm$5.0} \\
Mamba + geochem + location & 1 & 0.384{\tiny$\pm$0.099} & 1.046{\tiny$\pm$0.276} & 1.157{\tiny$\pm$0.490} & 1.894{\tiny$\pm$0.419} & 0.086{\tiny$\pm$0.016} & 0.204{\tiny$\pm$0.045} & 0.225{\tiny$\pm$0.065} & 0.324{\tiny$\pm$0.045} & 70.5{\tiny$\pm$2.9} & 61.5{\tiny$\pm$5.6} & 52.7{\tiny$\pm$5.5} & 39.8{\tiny$\pm$5.5} \\
TCN + geochem + location & 1 & 0.285{\tiny$\pm$0.105} & 0.621{\tiny$\pm$0.251} & 0.978{\tiny$\pm$0.632} & 1.084{\tiny$\pm$0.313} & 0.053{\tiny$\pm$0.011} & 0.117{\tiny$\pm$0.030} & 0.133{\tiny$\pm$0.076} & 0.196{\tiny$\pm$0.038} & 71.5{\tiny$\pm$2.7} & 59.6{\tiny$\pm$5.1} & 59.3{\tiny$\pm$5.4} & 47.6{\tiny$\pm$5.9} \\
\midrule
\rowcolor{unitgray}
\multicolumn{14}{@{}l}{\textbf{Layer level}} \\
Layer marginal & 1 & 0.211{\tiny$\pm$0.086} & 0.563{\tiny$\pm$0.146} & 0.653{\tiny$\pm$0.341} & 1.072{\tiny$\pm$0.303} & 0.043{\tiny$\pm$0.013} & 0.118{\tiny$\pm$0.024} & 0.113{\tiny$\pm$0.046} & 0.202{\tiny$\pm$0.041} & 51.5{\tiny$\pm$2.6} & 49.2{\tiny$\pm$4.3} & 51.1{\tiny$\pm$5.2} & 43.0{\tiny$\pm$5.0} \\
Layer Markov & 1 & 0.195{\tiny$\pm$0.063} & 0.385{\tiny$\pm$0.089} & 0.509{\tiny$\pm$0.330} & 0.849{\tiny$\pm$0.301} & \best{0.035}{\tiny$\pm$0.011} & 0.096{\tiny$\pm$0.021} & 0.093{\tiny$\pm$0.033} & 0.152{\tiny$\pm$0.034} & 56.9{\tiny$\pm$2.4} & 53.4{\tiny$\pm$3.9} & 54.6{\tiny$\pm$5.1} & 47.0{\tiny$\pm$4.9} \\
GRU & 10 & 0.183{\tiny$\pm$0.047} & 0.356{\tiny$\pm$0.076} & 0.464{\tiny$\pm$0.288} & 0.550{\tiny$\pm$0.212} & 0.047{\tiny$\pm$0.012} & 0.085{\tiny$\pm$0.017} & 0.093{\tiny$\pm$0.028} & 0.118{\tiny$\pm$0.029} & 63.1{\tiny$\pm$2.5} & 57.3{\tiny$\pm$4.3} & 56.7{\tiny$\pm$5.3} & 51.3{\tiny$\pm$4.6} \\
Transformer & 10 & \best{0.172}{\tiny$\pm$0.043} & 0.320{\tiny$\pm$0.073} & 0.495{\tiny$\pm$0.327} & 0.544{\tiny$\pm$0.212} & 0.045{\tiny$\pm$0.012} & 0.077{\tiny$\pm$0.017} & 0.095{\tiny$\pm$0.030} & 0.114{\tiny$\pm$0.029} & 62.9{\tiny$\pm$2.5} & 57.4{\tiny$\pm$4.4} & 56.4{\tiny$\pm$5.3} & 50.5{\tiny$\pm$4.7} \\
GRU + R & 10 & 0.192{\tiny$\pm$0.058} & 0.341{\tiny$\pm$0.076} & 0.479{\tiny$\pm$0.311} & 0.561{\tiny$\pm$0.211} & 0.048{\tiny$\pm$0.014} & 0.083{\tiny$\pm$0.020} & 0.094{\tiny$\pm$0.029} & 0.118{\tiny$\pm$0.029} & 64.7{\tiny$\pm$2.5} & 57.4{\tiny$\pm$4.3} & 56.8{\tiny$\pm$5.2} & 51.5{\tiny$\pm$4.6} \\
GRU + PT & 10 & 0.185{\tiny$\pm$0.048} & \best{0.303}{\tiny$\pm$0.077} & 0.488{\tiny$\pm$0.347} & 0.551{\tiny$\pm$0.223} & 0.046{\tiny$\pm$0.010} & \best{0.072}{\tiny$\pm$0.016} & \best{0.090}{\tiny$\pm$0.035} & 0.113{\tiny$\pm$0.031} & 63.3{\tiny$\pm$2.5} & 57.6{\tiny$\pm$4.1} & 56.2{\tiny$\pm$5.3} & 52.0{\tiny$\pm$4.6} \\
GRU + PT + R (ours) & 10 & 0.186{\tiny$\pm$0.046} & 0.322{\tiny$\pm$0.076} & \best{0.408}{\tiny$\pm$0.259} & \best{0.458}{\tiny$\pm$0.172} & 0.052{\tiny$\pm$0.013} & 0.081{\tiny$\pm$0.020} & 0.090{\tiny$\pm$0.029} & \best{0.106}{\tiny$\pm$0.027} & 63.8{\tiny$\pm$2.3} & 56.3{\tiny$\pm$4.2} & 56.1{\tiny$\pm$5.2} & 51.1{\tiny$\pm$4.8} \\
\bottomrule
\end{tabular}
\end{adjustbox}
\end{table*}

Table~\ref{tab:app-task2-baselines} extends Table~\ref{tab:main-tasks} to all 34 fitted baselines and reports, next to
the joint KL, the bounded JSD of the same law and the per-metre agreement of the generated suffix with the hole's own
log. Three observations. \emph{Per-metre agreement rewards copying, not generation}: persistence (repeat the last
lithology to the bottom) has the highest Acc/m on every shifted split ($64.8 / 66.3 / 57.3$) and the worst joint KL of
the table ($1.3$--$3.0$), because it produces a single layer of the wrong thickness; indicator kriging on the archive
pool likewise tops Acc/m on \textsc{test} ($77.8$) while its KL is $0.437$. This is why the joint law, not agreement,
is the headline metric. \emph{Spatial lookups do not roll a hole forward}: $k$-NN and kriging read the neighbours'
lithology at each depth and reach KL $1.4$--$2.0$ on the two rings, above every sequence model; they are marked
$^{*}$ and never highlighted. \emph{The transfer boundary appears in generation as it does in prediction}: cell-level
networks conditioned on location produce the most implausible rollouts of the benchmark once the coordinates are far
from training (GRU\,+\,location KL $2.64$ on OOD-D and $6.02$ on \textsc{pilbara-test}, against $0.56$ and $1.00$ for
the same network without coordinates), whereas the unconditioned layer-level models degrade gradually
($0.18 \to 0.55$ for the GRU). Among fitted baselines the best rows are layer Markov on \textsc{test} ($0.150$) and the
$3$-gram / TPROGS pair on the cell level; on the shifted splits the layer-level neural rows lead.

\subsubsection{Paired differences}
\begin{table*}[t]
\centering
\footnotesize
\setlength{\tabcolsep}{3pt}
\caption{\textbf{Task~2 paired differences} in pooled joint KL and JSD (negative = first method better). Both methods are scored on the same report-cluster bootstrap resample ($B{=}1000$), seeds resampled independently. Intervals that exclude zero are bold. GRU and layer Transformer: the plain models with label smoothing ($\varepsilon{=}0.3$).}
\label{tab:app-task2-paired}
\begin{adjustbox}{max width=\textwidth}
\begin{tabular}{@{}l rrrr rrrr@{}}
\toprule
& \multicolumn{4}{c}{\textbf{$\Delta$ KL}} & \multicolumn{4}{c}{\textbf{$\Delta$ JSD}} \\
\cmidrule(lr){2-5}\cmidrule(lr){6-9}
\textbf{Comparison} & Test & OOD-R & OOD-D & Pil-T & Test & OOD-R & OOD-D & Pil-T \\
\midrule
GRU + R $-$ GRU & +0.008{\tiny\,[-0.023, +0.045]} & -0.015{\tiny\,[-0.055, +0.024]} & +0.014{\tiny\,[-0.088, +0.157]} & +0.011{\tiny\,[-0.057, +0.093]} & +0.001{\tiny\,[-0.010, +0.014]} & -0.002{\tiny\,[-0.015, +0.013]} & +0.001{\tiny\,[-0.006, +0.010]} & -0.001{\tiny\,[-0.010, +0.009]} \\
GRU + PT $-$ GRU & +0.001{\tiny\,[-0.022, +0.026]} & \textbf{-0.053{\tiny\,[-0.088, -0.019]}} & +0.023{\tiny\,[-0.065, +0.211]} & +0.000{\tiny\,[-0.085, +0.100]} & -0.001{\tiny\,[-0.009, +0.006]} & \textbf{-0.013{\tiny\,[-0.022, -0.004]}} & -0.003{\tiny\,[-0.014, +0.011]} & -0.005{\tiny\,[-0.016, +0.005]} \\
Ours $-$ GRU & +0.003{\tiny\,[-0.028, +0.042]} & -0.034{\tiny\,[-0.072, +0.013]} & -0.057{\tiny\,[-0.157, +0.038]} & \textbf{-0.092{\tiny\,[-0.202, -0.016]}} & +0.005{\tiny\,[-0.005, +0.017]} & -0.004{\tiny\,[-0.015, +0.011]} & -0.003{\tiny\,[-0.013, +0.010]} & \textbf{-0.013{\tiny\,[-0.023, -0.001]}} \\
Ours $-$ GRU + R \quad(adds PT) & -0.005{\tiny\,[-0.050, +0.040]} & -0.019{\tiny\,[-0.067, +0.032]} & -0.071{\tiny\,[-0.225, +0.022]} & \textbf{-0.103{\tiny\,[-0.207, -0.023]}} & +0.004{\tiny\,[-0.011, +0.017]} & -0.002{\tiny\,[-0.019, +0.015]} & -0.004{\tiny\,[-0.014, +0.009]} & -0.012{\tiny\,[-0.023, +0.001]} \\
Ours $-$ GRU + PT \quad(adds R) & +0.002{\tiny\,[-0.030, +0.039]} & +0.019{\tiny\,[-0.025, +0.072]} & -0.080{\tiny\,[-0.253, +0.018]} & \textbf{-0.093{\tiny\,[-0.191, -0.019]}} & +0.006{\tiny\,[-0.003, +0.017]} & +0.009{\tiny\,[-0.002, +0.023]} & +0.000{\tiny\,[-0.012, +0.016]} & -0.007{\tiny\,[-0.019, +0.005]} \\
Ours $-$ Layer Transformer & +0.015{\tiny\,[-0.017, +0.049]} & +0.003{\tiny\,[-0.037, +0.052]} & \textbf{-0.087{\tiny\,[-0.254, -0.002]}} & \textbf{-0.086{\tiny\,[-0.182, -0.016]}} & +0.007{\tiny\,[-0.003, +0.019]} & +0.004{\tiny\,[-0.007, +0.019]} & -0.004{\tiny\,[-0.015, +0.008]} & -0.008{\tiny\,[-0.022, +0.005]} \\
Ours $-$ Layer Markov & +0.036{\tiny\,[-0.014, +0.085]} & -0.063{\tiny\,[-0.123, +0.001]} & -0.102{\tiny\,[-0.312, +0.058]} & \textbf{-0.391{\tiny\,[-0.586, -0.205]}} & \textbf{+0.016{\tiny\,[+0.006, +0.028]}} & -0.015{\tiny\,[-0.029, +0.002]} & -0.003{\tiny\,[-0.051, +0.020]} & \textbf{-0.046{\tiny\,[-0.070, -0.022]}} \\
Ours $-$ cell Transformer & \textbf{-0.170{\tiny\,[-0.255, -0.093]}} & \textbf{-0.263{\tiny\,[-0.371, -0.152]}} & \textbf{-0.125{\tiny\,[-0.294, -0.004]}} & \textbf{-0.537{\tiny\,[-0.821, -0.310]}} & \textbf{-0.028{\tiny\,[-0.045, -0.011]}} & \textbf{-0.051{\tiny\,[-0.072, -0.027]}} & -0.012{\tiny\,[-0.035, +0.005]} & \textbf{-0.073{\tiny\,[-0.093, -0.048]}} \\
Ours $-$ XGBoost & -0.019{\tiny\,[-0.098, +0.042]} & \textbf{-0.195{\tiny\,[-0.390, -0.047]}} & \textbf{-0.542{\tiny\,[-0.874, -0.313]}} & \textbf{-1.318{\tiny\,[-1.612, -1.089]}} & +0.007{\tiny\,[-0.006, +0.022]} & \textbf{-0.038{\tiny\,[-0.083, -0.002]}} & \textbf{-0.095{\tiny\,[-0.158, -0.057]}} & \textbf{-0.238{\tiny\,[-0.273, -0.202]}} \\
Ours $-$ $3$-gram & +0.000{\tiny\,[-0.060, +0.054]} & \textbf{-0.211{\tiny\,[-0.449, -0.113]}} & \textbf{-0.373{\tiny\,[-0.729, -0.064]}} & \textbf{-0.502{\tiny\,[-0.681, -0.311]}} & +0.011{\tiny\,[-0.000, +0.024]} & \textbf{-0.029{\tiny\,[-0.047, -0.011]}} & -0.019{\tiny\,[-0.082, +0.015]} & \textbf{-0.069{\tiny\,[-0.097, -0.042]}} \\
\bottomrule
\end{tabular}
\end{adjustbox}
\end{table*}

The paired differences separate what the recipe changes from sampling noise. The plain layer-level GRU is trained with
label smoothing like every arm. Relative to it, neither component moves the joint KL in
distribution ($|\Delta| \le 0.01$, intervals include zero), and on OOD-R only pretraining does ($-0.053$): the generated
law of the Eastern Goldfields core is already matched by a sequence model trained on it. On the two hardest splits
neither component alone is distinguishable from the plain GRU. The full recipe is, on \textsc{pilbara-test} ($-0.092$),
but not on OOD-D ($-0.057$, interval includes zero). On \textsc{pilbara-test} adding pretraining to the retrieval model
($-0.103$) and adding retrieval to the pretrained model ($-0.093$) are each significant, so the two components are
complementary there: each helps only in the presence of the other. Against the layer-level Transformer the recipe is
better on OOD-D ($-0.087$) and \textsc{pilbara-test} ($-0.086$). Against layer Markov the recipe is $0.39$ better on \textsc{pilbara-test} and not distinguishable on
the three Eastern Goldfields splits ($+0.036$, $-0.063$, $-0.102$, all intervals include zero; on JSD layer Markov is
in fact significantly better in distribution, $+0.016$). Against the cell-level Transformer, XGBoost and the $3$-gram the
recipe is ahead on every shifted split with intervals that exclude zero. Two entries of Table~\ref{tab:main-tasks}
should be read with this: the best \textsc{test} row (layer Markov, $0.150$) and the best OOD-R row (GRU\,+\,PT,
$0.303$) are within noise of the full recipe ($0.186$ and $0.322$). The recipe's advantage is confined to OOD-D and
\textsc{pilbara-test}. There it is large against the fitted and cell-level baselines and smaller, but significant on
\textsc{pilbara-test}, against the label-smoothed layer-level networks.

\subsubsection{Backbones}
\begin{table*}[t]
\centering
\footnotesize
\setlength{\tabcolsep}{3pt}
\caption{\textbf{Task~2 across layer-level backbones.} Each backbone is trained plain, with retrieval (R), with fine-objective pretraining (PT), and with both. Pooled joint KL, JSD and per-metre agreement. $\pm$: half-width of the 95\% report-cluster bootstrap interval ($B{=}1000$; holes of one exploration report are resampled together); trained rows also resample seeds. Plug-in divergences are biased upward under resampling, so intervals are re-centred on the full-sample estimate. Layer-level neural rows (and the plain arms of the backbone table) use label smoothing ($\varepsilon{=}0.3$).}
\label{tab:app-task2-backbones}
\begin{adjustbox}{max width=\textwidth}
\begin{tabular}{@{}l c rrrr rrrr rrrr@{}}
\toprule
& & \multicolumn{4}{c}{\textbf{KL} $\downarrow$} & \multicolumn{4}{c}{\textbf{JSD} $\downarrow$} & \multicolumn{4}{c}{\textbf{Acc/m} $\uparrow$} \\
\cmidrule(lr){3-6} \cmidrule(lr){7-10} \cmidrule(lr){11-14}
\textbf{Method} & \textbf{runs} & Test & OOD-R & OOD-D & Pil-T & Test & OOD-R & OOD-D & Pil-T & Test & OOD-R & OOD-D & Pil-T \\
\midrule
\rowcolor{unitgray}
\multicolumn{14}{@{}l}{\textbf{GRU}} \\
GRU & 10 & 0.183{\tiny$\pm$0.050} & 0.356{\tiny$\pm$0.076} & 0.464{\tiny$\pm$0.278} & 0.550{\tiny$\pm$0.207} & 0.047{\tiny$\pm$0.011} & 0.085{\tiny$\pm$0.017} & 0.093{\tiny$\pm$0.028} & 0.118{\tiny$\pm$0.029} & 63.1{\tiny$\pm$2.4} & 57.3{\tiny$\pm$4.4} & 56.7{\tiny$\pm$5.2} & 51.3{\tiny$\pm$4.6} \\
\quad + R & 10 & 0.192{\tiny$\pm$0.059} & 0.341{\tiny$\pm$0.081} & 0.479{\tiny$\pm$0.306} & 0.561{\tiny$\pm$0.212} & 0.048{\tiny$\pm$0.014} & 0.083{\tiny$\pm$0.020} & 0.094{\tiny$\pm$0.029} & 0.118{\tiny$\pm$0.029} & \best{64.7}{\tiny$\pm$2.5} & 57.4{\tiny$\pm$4.4} & 56.8{\tiny$\pm$5.3} & 51.5{\tiny$\pm$4.7} \\
\quad + PT & 10 & 0.185{\tiny$\pm$0.050} & 0.303{\tiny$\pm$0.078} & 0.488{\tiny$\pm$0.349} & 0.551{\tiny$\pm$0.224} & 0.046{\tiny$\pm$0.010} & 0.072{\tiny$\pm$0.017} & 0.090{\tiny$\pm$0.035} & 0.113{\tiny$\pm$0.030} & 63.3{\tiny$\pm$2.4} & 57.6{\tiny$\pm$4.2} & 56.2{\tiny$\pm$5.2} & \best{52.0}{\tiny$\pm$4.6} \\
\quad + PT + R & 10 & 0.186{\tiny$\pm$0.047} & 0.322{\tiny$\pm$0.074} & 0.408{\tiny$\pm$0.264} & \best{0.458}{\tiny$\pm$0.169} & 0.052{\tiny$\pm$0.013} & 0.081{\tiny$\pm$0.020} & 0.090{\tiny$\pm$0.030} & \best{0.106}{\tiny$\pm$0.027} & 63.8{\tiny$\pm$2.4} & 56.3{\tiny$\pm$4.4} & 56.1{\tiny$\pm$5.2} & 51.1{\tiny$\pm$4.5} \\
\midrule
\rowcolor{unitgray}
\multicolumn{14}{@{}l}{\textbf{LSTM}} \\
LSTM & 3 & 0.164{\tiny$\pm$0.044} & 0.338{\tiny$\pm$0.079} & 0.439{\tiny$\pm$0.275} & 0.549{\tiny$\pm$0.217} & 0.045{\tiny$\pm$0.011} & 0.081{\tiny$\pm$0.018} & 0.092{\tiny$\pm$0.030} & 0.121{\tiny$\pm$0.030} & 62.7{\tiny$\pm$2.3} & 57.8{\tiny$\pm$4.3} & 55.9{\tiny$\pm$5.4} & 51.0{\tiny$\pm$4.6} \\
\quad + R & 3 & 0.170{\tiny$\pm$0.048} & 0.328{\tiny$\pm$0.080} & 0.486{\tiny$\pm$0.306} & 0.571{\tiny$\pm$0.224} & 0.044{\tiny$\pm$0.013} & 0.078{\tiny$\pm$0.021} & 0.093{\tiny$\pm$0.028} & 0.119{\tiny$\pm$0.030} & 64.5{\tiny$\pm$2.6} & 58.0{\tiny$\pm$4.8} & 56.7{\tiny$\pm$5.6} & 51.2{\tiny$\pm$4.7} \\
\quad + PT & 3 & 0.173{\tiny$\pm$0.041} & 0.321{\tiny$\pm$0.072} & 0.403{\tiny$\pm$0.236} & 0.494{\tiny$\pm$0.172} & 0.049{\tiny$\pm$0.012} & 0.083{\tiny$\pm$0.019} & 0.092{\tiny$\pm$0.027} & 0.118{\tiny$\pm$0.029} & 62.3{\tiny$\pm$2.5} & 57.2{\tiny$\pm$4.4} & 55.6{\tiny$\pm$5.2} & 50.1{\tiny$\pm$4.5} \\
\quad + PT + R & 3 & 0.178{\tiny$\pm$0.050} & 0.303{\tiny$\pm$0.076} & 0.459{\tiny$\pm$0.309} & 0.543{\tiny$\pm$0.204} & 0.048{\tiny$\pm$0.012} & 0.075{\tiny$\pm$0.021} & 0.092{\tiny$\pm$0.028} & 0.119{\tiny$\pm$0.031} & 63.6{\tiny$\pm$2.5} & 56.8{\tiny$\pm$4.4} & 55.6{\tiny$\pm$5.2} & 50.5{\tiny$\pm$4.9} \\
\midrule
\rowcolor{unitgray}
\multicolumn{14}{@{}l}{\textbf{Transformer}} \\
Transformer & 10 & 0.172{\tiny$\pm$0.045} & 0.320{\tiny$\pm$0.072} & 0.495{\tiny$\pm$0.333} & 0.544{\tiny$\pm$0.210} & 0.045{\tiny$\pm$0.012} & 0.077{\tiny$\pm$0.017} & 0.095{\tiny$\pm$0.030} & 0.114{\tiny$\pm$0.027} & 62.9{\tiny$\pm$2.6} & 57.4{\tiny$\pm$4.2} & 56.4{\tiny$\pm$5.3} & 50.5{\tiny$\pm$4.6} \\
\quad + R & 3 & 0.182{\tiny$\pm$0.045} & 0.321{\tiny$\pm$0.070} & 0.441{\tiny$\pm$0.263} & 0.509{\tiny$\pm$0.173} & 0.049{\tiny$\pm$0.014} & 0.080{\tiny$\pm$0.019} & 0.092{\tiny$\pm$0.026} & 0.118{\tiny$\pm$0.029} & 63.9{\tiny$\pm$2.4} & 57.6{\tiny$\pm$4.6} & 56.4{\tiny$\pm$5.5} & 49.8{\tiny$\pm$4.9} \\
\quad + PT & 3 & 0.168{\tiny$\pm$0.045} & 0.285{\tiny$\pm$0.085} & 0.473{\tiny$\pm$0.368} & 0.541{\tiny$\pm$0.219} & 0.045{\tiny$\pm$0.011} & 0.072{\tiny$\pm$0.020} & 0.088{\tiny$\pm$0.030} & 0.120{\tiny$\pm$0.032} & 63.0{\tiny$\pm$2.4} & 56.9{\tiny$\pm$4.4} & 55.7{\tiny$\pm$5.2} & 51.0{\tiny$\pm$4.7} \\
\quad + PT + R & 3 & 0.175{\tiny$\pm$0.047} & \best{0.275}{\tiny$\pm$0.074} & 0.451{\tiny$\pm$0.308} & 0.521{\tiny$\pm$0.189} & 0.047{\tiny$\pm$0.015} & 0.071{\tiny$\pm$0.021} & 0.090{\tiny$\pm$0.030} & 0.113{\tiny$\pm$0.030} & 64.1{\tiny$\pm$2.5} & 57.2{\tiny$\pm$4.3} & 56.2{\tiny$\pm$5.6} & 49.9{\tiny$\pm$5.0} \\
\midrule
\rowcolor{unitgray}
\multicolumn{14}{@{}l}{\textbf{TCN}} \\
TCN & 3 & \best{0.161}{\tiny$\pm$0.042} & 0.328{\tiny$\pm$0.081} & 0.450{\tiny$\pm$0.317} & 0.596{\tiny$\pm$0.228} & 0.042{\tiny$\pm$0.010} & 0.077{\tiny$\pm$0.017} & 0.089{\tiny$\pm$0.031} & 0.131{\tiny$\pm$0.033} & 63.7{\tiny$\pm$2.3} & \best{58.5}{\tiny$\pm$4.3} & \best{56.9}{\tiny$\pm$5.6} & 51.2{\tiny$\pm$4.6} \\
\quad + R & 3 & 0.171{\tiny$\pm$0.047} & 0.324{\tiny$\pm$0.073} & 0.409{\tiny$\pm$0.240} & 0.574{\tiny$\pm$0.205} & 0.048{\tiny$\pm$0.014} & 0.080{\tiny$\pm$0.020} & 0.091{\tiny$\pm$0.025} & 0.134{\tiny$\pm$0.034} & 64.0{\tiny$\pm$2.6} & 58.3{\tiny$\pm$4.7} & 56.7{\tiny$\pm$5.4} & 49.9{\tiny$\pm$5.2} \\
\quad + PT & 3 & 0.204{\tiny$\pm$0.049} & 0.349{\tiny$\pm$0.078} & \best{0.345}{\tiny$\pm$0.163} & 0.531{\tiny$\pm$0.163} & 0.059{\tiny$\pm$0.015} & 0.092{\tiny$\pm$0.020} & 0.089{\tiny$\pm$0.026} & 0.131{\tiny$\pm$0.030} & 62.0{\tiny$\pm$2.4} & 57.9{\tiny$\pm$4.4} & 55.3{\tiny$\pm$5.2} & 48.8{\tiny$\pm$4.6} \\
\quad + PT + R & 3 & 0.169{\tiny$\pm$0.044} & 0.282{\tiny$\pm$0.073} & 0.426{\tiny$\pm$0.258} & 0.580{\tiny$\pm$0.221} & 0.045{\tiny$\pm$0.011} & \best{0.068}{\tiny$\pm$0.016} & 0.090{\tiny$\pm$0.030} & 0.124{\tiny$\pm$0.032} & 64.6{\tiny$\pm$2.6} & 57.6{\tiny$\pm$4.4} & 56.4{\tiny$\pm$5.4} & 50.3{\tiny$\pm$4.8} \\
\midrule
\rowcolor{unitgray}
\multicolumn{14}{@{}l}{\textbf{Mamba}} \\
Mamba & 3 & 0.175{\tiny$\pm$0.043} & 0.346{\tiny$\pm$0.074} & 0.451{\tiny$\pm$0.257} & 0.587{\tiny$\pm$0.195} & 0.048{\tiny$\pm$0.012} & 0.088{\tiny$\pm$0.019} & 0.096{\tiny$\pm$0.027} & 0.133{\tiny$\pm$0.034} & 62.3{\tiny$\pm$2.4} & 57.1{\tiny$\pm$4.5} & 55.5{\tiny$\pm$5.3} & 48.7{\tiny$\pm$5.3} \\
\quad + R & 3 & 0.167{\tiny$\pm$0.042} & 0.316{\tiny$\pm$0.073} & 0.467{\tiny$\pm$0.283} & 0.599{\tiny$\pm$0.216} & 0.045{\tiny$\pm$0.011} & 0.076{\tiny$\pm$0.017} & 0.095{\tiny$\pm$0.028} & 0.128{\tiny$\pm$0.032} & 63.7{\tiny$\pm$2.4} & 57.6{\tiny$\pm$4.5} & 55.7{\tiny$\pm$5.4} & 49.3{\tiny$\pm$5.2} \\
\quad + PT & 3 & 0.193{\tiny$\pm$0.043} & 0.345{\tiny$\pm$0.078} & 0.390{\tiny$\pm$0.220} & 0.544{\tiny$\pm$0.160} & 0.054{\tiny$\pm$0.014} & 0.089{\tiny$\pm$0.023} & 0.088{\tiny$\pm$0.029} & 0.130{\tiny$\pm$0.028} & 62.0{\tiny$\pm$2.4} & 56.4{\tiny$\pm$4.5} & 55.2{\tiny$\pm$5.3} & 49.0{\tiny$\pm$5.1} \\
\quad + PT + R & 3 & 0.162{\tiny$\pm$0.051} & 0.297{\tiny$\pm$0.084} & 0.456{\tiny$\pm$0.326} & 0.597{\tiny$\pm$0.219} & \best{0.042}{\tiny$\pm$0.010} & 0.069{\tiny$\pm$0.017} & \best{0.086}{\tiny$\pm$0.029} & 0.126{\tiny$\pm$0.033} & 64.3{\tiny$\pm$2.4} & 57.4{\tiny$\pm$4.2} & 56.6{\tiny$\pm$5.3} & 49.7{\tiny$\pm$4.7} \\
\bottomrule
\end{tabular}
\end{adjustbox}
\end{table*}

Table~\ref{tab:app-task2-backbones} repeats the four arms for five layer-level backbones (GRU and plain Transformer with
10 seeds, the rest with 3, so their intervals are wider). Every plain arm is trained with label smoothing. On
\textsc{test} and OOD-R the four arms of every backbone lie within $0.16$--$0.36$ and nothing separates them. On
\textsc{pilbara-test} the best augmented arm of every backbone is below its plain arm (plain $0.54$--$0.60$, best arm
$0.46$--$0.54$), but which arm is best varies, and retrieval alone does not lower the KL for every backbone; against the
label-smoothed plain arm most of these differences are within noise. With three
seeds the ordering of the two augmented arms is not stable: LSTM\,+\,PT ($0.403 / 0.494$ on
OOD-D / \textsc{pilbara-test}) and TCN\,+\,PT ($0.345$ on OOD-D) beat their own full recipe, and Transformer\,+\,PT\,+\,R
is the best OOD-R row of the table ($0.275$). For generation we therefore claim no per-component gain that holds for
every backbone. We use the GRU in the main text because it is the most even across the two tasks. All backbones share
one learning rate ($3{\times}10^{-4}$); Mamba at its own larger rate diverged for seed-determined runs, and the shared
rate removes the divergence without changing its scores.

\subsubsection{Other metrics}
Re-scoring the main-table methods under the joint law at 13 coarse classes, under JSD, under a report-balanced KL in
which every report contributes one normalised histogram, and under three per-hole metrics (Acc/m, normalised
layer-sequence edit distance, boundary F1 at $\pm1$\,m) changes no conclusion. The full recipe is the best row on
OOD-D and \textsc{pilbara-test} under KL, KL-13, JSD and the balanced KL (e.g.\ KL-13 $0.225 / 0.196$ vs.\
$0.267 / 0.385$ for layer Markov); layer Markov keeps \textsc{test} under every divergence, and OOD-R goes to
GRU\,+\,PT under KL, JSD and the balanced KL and to the full recipe under KL-13. The per-hole metrics are flat across all layer-level rows (Acc/m $56$--$65$, edit
distance $0.58$--$0.63$, boundary F1 $10$--$12\%$) and are not what separates methods: a free-running rollout is not
expected to reproduce the exact boundaries of one hole, only the layer statistics of the split, which is what the
divergences measure. The price is that the pooled law does not score the fit of each continuation to its own prefix: in distribution a prefix-agnostic generator comes close (layer marginal 0.211 vs.\ 0.186), so the metric discriminates mainly under shift, and all generators are also given the logged stopping depth; a per-hole likelihood of the true suffix would complement it.

\lstdefinestyle{prompt}{basicstyle=\ttfamily\scriptsize,breaklines=true,breakindent=0pt,
  columns=fullflexible,keepspaces=true,frame=single,framesep=3pt,rulecolor=\color{black!35},
  aboveskip=4pt,belowskip=4pt,xleftmargin=3pt,xrightmargin=3pt}

\subsection{Zero-Shot LLM Baselines: Implementation and Prompts}
\label{app:llm}

This appendix documents how the zero-shot large-language-model (LLM) rows are produced: which points are
scored, the exact text every model receives, how the answer is read out, and how each model was run. All
providers import one prompt builder and one scoring routine, so every model sees byte-identical text for
the same prediction point and is scored by the same rule as every trained method.

\subsubsection{Pipeline}

\begin{enumerate}
\item \textbf{Hole sampling.} For each split we draw 150 holes with at least two intervals by a seeded random
permutation (seed 0). Holes are stored project by project, so a prefix slice would cover only a few prospects.
\item \textbf{Prediction points.} A point is a cell that starts a layer and whose own token is a real lithology,
i.e.\ neither \texttt{CONT} nor unmapped. This is the Task~1 mask used for every other method. The 750 sampled
holes yield $2{,}161$ points (Table~\ref{tab:llm-points}).
\item \textbf{Context.} The model sees only the layers that start strictly above the point, top to bottom.
Unmapped layers are shown as \texttt{UNLOGGED}; $46.8\%$ of context layers carry this placeholder.
Points with no layer above them ($215$ of $2{,}161$) are cold starts and receive a dedicated sentence.
\item \textbf{One request per point.} Showing a whole hole and asking for every layer would leak labels, because
the later layers of a hole are the targets of its earlier points. Each point is therefore an independent request.
\item \textbf{Read-out.} Open-weight models are scored by candidate likelihood; hosted models return a
ranked top-3 in JSON.
\item \textbf{Scoring.} Top-1 and top-3 over the 37 lithologies, with per-hole hit/point tallies so that any
LLM row can be paired with any other method on the same holes by the hole-cluster bootstrap. Layer-start AUROC
and per-cell top-1 are left empty: the models are never asked where a layer ends, and deriving a boundary
probability from the true layer starts would score them on the answer key.
\end{enumerate}

\begin{table}[h]
\centering\small
\begin{tabular}{lrrr}
\toprule
\textbf{Split} & \textbf{Points} & \textbf{w/ retrieval} & \textbf{median dist.} \\
\midrule
\texttt{val}           & 466 & 46.6\% & 115\,m \\
\texttt{test}          & 361 & 43.8\% & 137\,m \\
\texttt{ood\_near}     & 471 & 15.7\% & 1{,}503\,m \\
\texttt{ood\_far}      & 401 & 18.7\% & 498\,m \\
\texttt{pilbara\_test} & 462 & 32.0\% & 1{,}202\,m \\
\midrule
Total                  & 2{,}161 & & \\
\bottomrule
\end{tabular}
\caption{Prediction points per split (150 holes each) and, for the \textbf{retrieval} arm, the share of
points with any neighbour evidence at that depth and the median weighted neighbour distance among them.}
\label{tab:llm-points}
\end{table}

\subsubsection{Prompts}

\paragraph{System prompt.} Shared by every point of an arm. The full text of the \textbf{strict} arm is reproduced below; the code list is one code per line in the actual prompt and is abbreviated here.

\begin{promptbox}[title=System prompt (strict arm)]
You are logging a mineral exploration drill hole. Given the lithological units already logged from the surface downward, name the single most likely lithology of the NEXT unit below.

Answer with codes from this list only (37 options):\\
\pcode{ALLU}{alluvium}\\
\pcode{AMPH}{amphibolite}\\
\pcode{ANDS}{andesite}\\
\pcode{...}{(one line per code, 37 lines)}\\
\pcode{UNKN}{unknown / unlogged}\\
\pcode{VEIN}{vein}

Units shown as UNLOGGED are intervals whose driller's code could not be mapped to this list. They are gaps in the record, not a rock type: reason across them to the rock that most likely lies below.

Return your three best guesses, most likely first.
\end{promptbox}

The 37 codes and their glosses are: ALLU alluvium, AMPH amphibolite, ANDS andesite, BSLT basalt, CARB carbonate, CGLM conglomerate, CHRT chert, CLAY clay, COLL colluvium, DACI dacite, DIOR diorite, DOLR dolerite, GABR gabbro, GNSS gneiss, GRAN granite, GRND granodiorite, GRNS greenstone, GRVL gravel, IRON iron formation / BIF, KOMA komatiite, LATR laterite, PEGM pegmatite, PHYL phyllite, PORP porphyry, QRTZ quartz, REGL regolith, RHYO rhyolite, SAND sand, SAPL saprolite, SCHI schist, SDST sandstone, SERP serpentinite, SHLE shale, SKAR skarn, TOPS topsoil, UNKN unknown / unlogged, VEIN vein. Spelling out each code removes the possibility that a model fails only because it does not recognise an abbreviation. The \texttt{UNLOGGED} clause separates a gap in the record from \texttt{UNKN}, which is a logged category in the schema.

The other two arms insert one paragraph after the first sentence of the system prompt:

\begin{promptbox}[title=Inserted paragraph (geo and retrieval arms)]
{[geo]} Each layer also carries its top depth and thickness in metres, and you are told which province the hole is in.

{[retrieval]} You are also given what NEARBY drill holes logged at the same depth. These are real observations from the same area, inverse-distance weighted, grouped into broad rock classes. They are strong evidence but not ground truth: when few neighbours reach the depth, or they sit far away, rely more on the sequence itself.
\end{promptbox}

\paragraph{User turn.} The three arms differ only here. Below is the same \texttt{test} point under each arm: the fifth layer of a hole in the Eastern Goldfields, whose true lithology is \texttt{ALLU}.

\begin{tcbraster}[raster columns=3, raster equal height, raster valign=top, raster column skip=4pt, raster force size=false]
\begin{promptbox}[title=strict, fontupper=\ttfamily\scriptsize, width=0.26\linewidth]
Units logged so far, top to bottom:{\setlength{\parskip}{0pt}
\pline{1. CLAY}
\pline{2. UNLOGGED}
\pline{3. CLAY}
\pline{4. UNLOGGED}\par}

What is the next unit's lithology?
\end{promptbox}
\begin{promptbox}[title=geo, fontupper=\ttfamily\scriptsize, width=0.41\linewidth]
Province: the Eastern Goldfields, Yilgarn Craton, Western Australia.

Units logged so far, top to bottom:{\setlength{\parskip}{0pt}
\pline{1. CLAY (top 0 m, 1 m thick)}
\pline{2. UNLOGGED (top 1 m, 2 m thick)}
\pline{3. CLAY (top 3 m, 45 m thick)}
\pline{4. UNLOGGED (top 48 m, 10 m thick)}\par}

The next unit starts at 58 m.

What is the next unit's lithology?
\end{promptbox}
\begin{promptbox}[title=retrieval, fontupper=\ttfamily\scriptsize, width=0.33\linewidth-8pt]
Units logged so far, top to bottom:{\setlength{\parskip}{0pt}
\pline{1. CLAY}
\pline{2. UNLOGGED}
\pline{3. CLAY}
\pline{4. UNLOGGED}\par}

Nearby holes logged at this depth: cover/alluvium 100\%.\\
(10\% of the retrieved holes reach this depth; mean distance 100 m.)

What is the next unit's lithology?
\end{promptbox}
\end{tcbraster}

{\sloppy Cold starts replace the layer list with \texttt{No units have been logged yet --- this is the first unit, at the surface.} (\textbf{geo}: \texttt{No units logged yet; this is the first unit, at 0 m.}, after the province line). The \textbf{strict} arm matches the inputs of our model's lithology head, which also sees neither depth nor thickness. The \textbf{geo} arm measures how much of a score comes from a regional prior. The Pilbara province line reads \texttt{the Pilbara Craton, Western Australia}.\par}

\textbf{Retrieval evidence.} The neighbour sentence is rendered from the same retrieval context as our model (\S\ref{subsec:retrieval}): \textsc{train} and WA-DASC-full, $k{=}10$, elevation-aligned where possible, a 25\,m minimum distance against re-reported twin holes, and a query hole never retrieves itself. For readability, the 37-class inverse-distance-weighted distribution at the query depth is grouped into 13 broad classes, and the five largest classes above 1\% are shown, together with the share of the $k$ neighbours that reach this depth and their mean distance, so that the model can judge how much evidence is available. When no neighbour reaches the depth, the sentence is \texttt{No nearby hole reaches this depth}. Neighbour bed thickness is not shown, so the retrieval arm adds only neighbour lithology, coverage and distance to the strict arm.

\subsubsection{Open-weight models: likelihood scoring}

Every candidate code is appended to the chat-templated prompt and scored by teacher-forced joint
log-likelihood; the prediction is the ranking of the candidates. The user turn gets one extra line,
\texttt{Answer with the single code only.}, so that the answer begins at the next token. Qwen3 is run with
\texttt{enable\_thinking=False} for the same reason. The model never has to emit a parseable string, so there
are no parse failures by construction.

Raw likelihood is biased here. On the Qwen3 tokeniser the 36 candidates are one token (1 code), two tokens
(27) or three tokens (8), and the model places almost all next-token mass on \texttt{UN}, echoing the
\texttt{UNLOGGED} placeholders. Two corrections are applied by default:
\begin{itemize}
\item \textbf{Domain-conditional PMI}~\citep{holtzman2021surface}:
$s(c) = \log P(c \mid \text{prompt}) - \log P(c \mid \text{neutral})$. The neutral prompt keeps the system prompt
and the question but removes all hole evidence, so the correction cancels each code's surface preference.
\item \textbf{\texttt{UNKN} is not a candidate.} It is the target on 2 of 4{,}258 points in the scored splits of
the pilot, all in \texttt{ood\_far}; those points count as misses.
\end{itemize}
Candidates are appended with no leading space, because the chat template ends on a newline: \texttt{" GRAN"}
tokenises as \texttt{[' GR','AN']} and scored 16.5 nats below \texttt{"GRAN"} on a diagnostic point.
The implementation repeats the prompt once per candidate in chunks of 8 and requests logits for the last
$L{+}1$ positions only, which keeps a 32B model inside one GPU.

\paragraph{Generative bridge.} Because hosted models cannot be likelihood-scored, Qwen3-8B and Qwen3-32B are also
run in generation mode: the exact hosted prompt with no extra line, greedy decoding with at most 96 new tokens,
thinking disabled, and the same parser as the hosted models. These rows isolate the effect of the read-out
instrument from the effect of the model.

\subsubsection{Hosted models: generative scoring}

Hosted models return \texttt{\{"top3": [c1, c2, c3]\}} under a strict JSON schema with exactly three strings.
The parser reads the JSON and falls back to any four-capital-letter tokens in the text, truncates each entry to
four characters, keeps only valid codes, and removes duplicates. A point whose response cannot be retrieved or
parsed is excluded from that row's denominator and counted in the parse-failure column of
Table~\ref{tab:llm-results}. Such rows are compared with our full recipe (GRU\,+\,PT\,+\,R) on the same quantity, top-1 under the shared mask, but not directly with likelihood-scored rows.
\begin{itemize}
\item \textbf{GPT-6-astra}: OpenAI Batch API, \texttt{/v1/chat/completions} with a 24-hour completion window;
system and user messages as above; \texttt{reasoning\_effort=medium}.
\item \textbf{Gemini-pro-latest}: Gemini batch API with \texttt{response\_mime\_type=application/\\json}, the same
schema, and \texttt{thinking\_level=medium}. The batch payload has no system field, so the system prompt is
placed in the user turn followed by a blank line; the text the model reads is otherwise identical.
\end{itemize}

\subsubsection{Models and configuration}

\begin{table}[h]
\centering\small
\begin{adjustbox}{max width=\columnwidth}
\begin{tabular}{lllll}
\toprule
\textbf{Model} & \textbf{Access} & \textbf{Read-out} & \textbf{Precision} & \textbf{Hardware / setting} \\
\midrule
Qwen3-8B    & open & likelihood+PMI & fp16 & V100 32\,GB \\
Qwen3-14B   & open & likelihood+PMI & bf16 & H100 \\
Qwen3-32B   & open & likelihood+PMI & bf16 & H100 \\
phi-4 (14B) & open & likelihood+PMI & bf16 & H100 \\
Qwen3-8B, 32B & open & generative bridge & bf16 & A6000, greedy \\
GPT-6-astra & hosted & generative & -- & batch, effort medium \\
Gemini-pro-latest & hosted & generative & -- & batch, thinking medium \\
\bottomrule
\end{tabular}
\end{adjustbox}
\caption{Zero-shot configurations. Llama and Gemma were not run because their weights require manual approval.}
\label{tab:llm-config}
\end{table}

\begin{table*}[t]
\centering\small
\begin{tabular}{llccccc r}
\toprule
\textbf{Model} & \textbf{Arm} & \texttt{val} & \texttt{test} & \texttt{ood\_r} & \texttt{ood\_d} & \texttt{pilbara\_test} & \textbf{parse fail.} \\
\midrule
\multicolumn{8}{l}{\emph{Likelihood + PMI}} \\
Qwen3-8B  & strict    & 14.6 / 36.9 & 14.4 / 32.7 & 20.6 / 41.4 & 24.2 / 50.1 & 18.8 / 32.7 & 0 \\
          & geo       & 15.4 / 30.3 & 17.2 / 26.6 & 21.7 / 25.5 & 29.7 / 31.9 & 8.2 / 26.6 & 0 \\
          & retrieval & 29.2 / 54.5 & 33.0 / 52.1 & 25.5 / 52.2 & 37.7 / 51.1 & 23.4 / 44.6 & 0 \\
Qwen3-14B & strict    & 29.0 / 44.2 & 24.4 / 33.2 & 31.2 / 39.5 & 38.4 / 51.4 & 42.4 / 60.2 & 0 \\
          & geo       & 21.0 / 35.4 & 16.6 / 25.2 & 22.9 / 35.2 & 29.7 / 43.1 & 14.5 / 38.5 & 0 \\
          & retrieval & 48.7 / 63.7 & 51.5 / 58.7 & 36.1 / 46.7 & 49.1 / 56.6 & 40.5 / 59.7 & 0 \\
Qwen3-32B & strict    & 22.8 / 46.6 & 20.8 / 37.4 & 29.3 / 52.6 & 32.7 / 41.1 & 35.9 / 50.4 & 0 \\
          & geo       & 14.4 / 34.3 & 10.8 / 26.9 & 21.0 / 27.0 & 12.2 / 29.9 & 28.1 / 43.7 & 0 \\
          & retrieval & 44.6 / 67.0 & 48.5 / 64.0 & 36.5 / 43.7 & 39.6 / 49.1 & 32.9 / 55.4 & 0 \\
phi-4     & strict    & 26.0 / 38.4 & 21.1 / 31.3 & 29.3 / 48.0 & 36.2 / 48.9 & 30.3 / 41.3 & 0 \\
          & geo       & 6.2 / 17.4  & 2.2 / 14.1  & 5.1 / 14.2  & 12.5 / 33.4 & 18.2 / 23.4 & 0 \\
          & retrieval & 44.2 / 54.5 & 47.4 / 55.4 & 29.5 / 51.2 & 37.4 / 54.6 & 34.2 / 44.6 & 0 \\
\midrule
\multicolumn{8}{l}{\emph{Generative (bridge)}} \\
Qwen3-8B  & strict    & 29.0 / 49.6 & 23.3 / 44.6 & 30.1 / 48.8 & 35.4 / 53.9 & 45.5 / 63.4 & 0 \\
          & geo       & 25.1 / 34.3 & 20.2 / 29.1 & 24.6 / 31.4 & 29.7 / 35.4 & 32.7 / 44.2 & 0 \\
          & retrieval & 39.7 / 54.3 & 35.2 / 44.9 & 32.1 / 40.8 & 41.9 / 54.1 & 44.4 / 59.1 & 0 \\
Qwen3-32B & strict    & 23.4 / 33.9 & 16.1 / 25.8 & 7.9 / 10.2  & 17.2 / 20.2 & 26.8 / 40.7 & 0 \\
          & geo       & 17.8 / 28.5 & 9.7 / 14.7  & 14.9 / 20.0 & 15.5 / 23.9 & 27.1 / 43.1 & 0 \\
          & retrieval & 42.9 / 56.4 & 47.4 / 55.1 & 24.6 / 32.7 & 30.4 / 37.9 & 29.4 / 47.0 & 0 \\
\midrule
\multicolumn{8}{l}{\emph{Generative (hosted)}} \\
GPT-6-astra & strict    & 38.6 / 56.9 & 30.5 / 46.5 & 30.8 / 46.3 & 43.6 / 61.1 & 47.4 / 59.3 & 0 \\
            & geo       & 41.6 / 60.9 & 35.2 / 56.2 & 27.2 / 53.5 & 44.4 / 70.1 & 40.3 / 53.7 & 0 \\
            & retrieval & 55.4 / 71.9 & 53.2 / 67.0 & 38.0 / 56.9 & 51.1 / 68.1 & 45.9 / 64.1 & 0 \\
Gemini-pro-latest & strict    & 9.7 / 21.5  & 8.9 / 21.6  & 11.2 / 22.3 & 20.7 / 31.9 & 5.4 / 14.9  & 0 \\
                  & geo       & 13.7 / 30.0 & 14.7 / 30.5 & 15.7 / 25.3 & 16.3 / 31.0 & 4.1 / 14.1  & 26 \\
                  & retrieval$^\dagger$ & 55.1 / 74.5 & 59.7 / 73.7 & 32.0 / 51.8 & 37.2 / 55.5 & 50.0 / 58.3 & 1{,}409 \\
\bottomrule
\end{tabular}
\caption{Zero-shot LLM next-layer top-1 / top-3 (\%) on the 150-hole samples of the original five splits;
$2{,}161$ points per row. Parse failures are excluded from the denominator. $^\dagger$Scored on the 752 points
that parsed; not a capability estimate.}
\label{tab:llm-results}
\end{table*}

\subsubsection{LLM Limitations}

\paragraph{Training-data contamination.} The WA DASC drill-hole records are public. A hosted model may have seen
them in pre-training, in which case part of its cross-province score reflects recall rather than inference. We
cannot audit closed training corpora, and this caveat applies to every hosted row.

\paragraph{Structured-output reliability.} GPT-6-astra and all open-weight runs parsed every point. Gemini-pro-latest
failed on 1{,}409 of 2{,}161 points ($65.2\%$) in the \textbf{retrieval} arm, whose prompts are longest, and on up
to $97\%$ of \texttt{pilbara\_test}. The surviving points are a biased subset.

\paragraph{Sampling.} Each cell is one run over 150 holes per split. The read-out is deterministic given the
prompt for the likelihood and greedy rows, so there is no seed variance, but hole-sampling variance remains:
the hole-cluster bootstrap 95\% interval is about $\pm3$--$4$ points, and smaller differences should not be read
as an ordering.

\paragraph{Cost.} The three GPT-6-astra arms consumed 3.36\,M input and 0.53\,M output tokens, $65\%$ of them
reasoning tokens, for one evaluation pass. A likelihood-scored open-weight arm takes about 2\,h on one GPU.
Our full recipe (GRU\,+\,PT\,+\,R) has 0.68\,M parameters and runs locally.

\providecolor{unitgray}{HTML}{E5E7EB}
\subsection{Pretraining, Retrieval and Hyperparameters}
\label{app:ptrk}

\begin{figure*}[t]
\centering
\includegraphics[width=\textwidth]{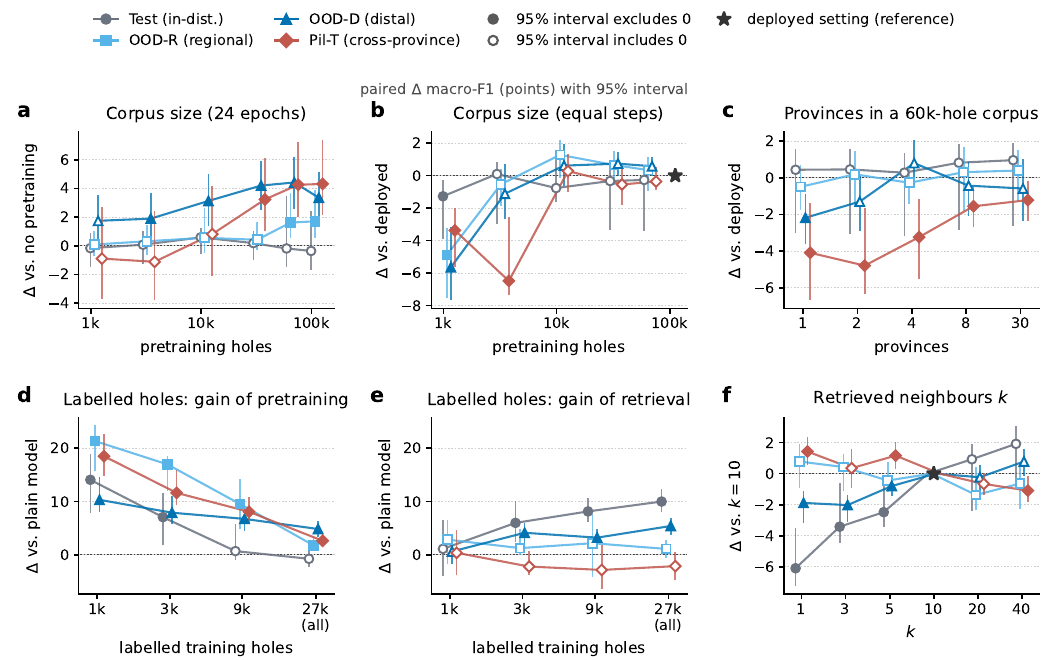}
\caption{\textbf{Pretraining and retrieval ablations} on Task~1: paired $\Delta$ macro-F1 (points) with $95\%$
intervals; filled markers are differences whose interval excludes zero, the star marks the deployed reference.
Panels a, b, c and f use the full recipe. \textbf{(a)} Pretraining-corpus size at a fixed 24 epochs, against no
pretraining. \textbf{(b)} The same corpus sizes with the number of optimisation steps held at that of the full corpus,
against the deployed checkpoint. \textbf{(c)} Number of provinces in a 60k-hole corpus, against the deployed checkpoint.
\textbf{(d, e)} Labelled training holes: gain of pretraining (d) and of retrieval (e) over the plain model trained on
the same holes. \textbf{(f)} Number of retrieved neighbours $k$, against the deployed $k{=}10$.}
\label{fig:app-ablation}
\end{figure*}

This section collects every experiment on the two components of the recipe and on the settings around them. Unless
stated otherwise, experiments use the layer-level GRU of the main text on Task~1 and are scored as in
App.~\ref{app:results_task1}: macro-F1 over classes with at least 20 events, and \emph{paired} differences against a
stated reference, with both arms scored on the same hole-cluster bootstrap resample and seeds resampled. Arms use three
fine-tuning seeds, and the deployed configuration uses ten. Each pretraining variant is a single checkpoint. ``Full
recipe'' means pretraining plus retrieval. Table~\ref{tab:app-hparams} lists every setting, its deployed value and where
it is varied, and Figure~\ref{fig:app-ablation} summarises the quantitative axes.

\begin{table}[t]
\centering
\footnotesize
\setlength{\tabcolsep}{3pt}
\caption{\textbf{Settings of the recipe}: deployed value, the alternatives tested, and where. Settings without a table
entry were fixed a priori and not tuned.}
\label{tab:app-hparams}
\begin{tabular}{@{}l l l l@{}}
\toprule
\textbf{Setting} & \textbf{Deployed} & \textbf{Tested} & \textbf{Where} \\
\midrule
\multicolumn{4}{@{}l}{\textit{Pretraining}} \\
Corpus & all 112k holes & 1.1k--67k holes; 1--30 provinces; & Tab.~\ref{tab:abl-ptscale}, \ref{tab:abl-ptprov}, \\
 & & region held out & \ref{tab:province_holdout} \\
Optimisation length & 24 epochs & compute-matched fractions & Tab.~\ref{tab:abl-ptcmatch} \\
Objective & fine + chemistry & fine only; coarse $\pm$ chemistry & Tab.~\ref{tab:abl-ptobj} \\
\midrule
\multicolumn{4}{@{}l}{\textit{Retrieval}} \\
Neighbours $k$ & 10 & 1--40; 1\,km radius; consensus & Tab.~\ref{tab:abl-topk} \\
Pool & train + archive & none; train only; + target context & Tab.~\ref{tab:abl-pool}, \ref{tab:abl-poolcurve} \\
Neighbour dropout & 0.3 & 0, 0.6 & Tab.~\ref{tab:abl-rkknob} \\
Self-exclusion radius & 25\,m & none & Tab.~\ref{tab:abl-rkknob} \\
Depth alignment & elevation & hole depth & Tab.~\ref{tab:abl-rkknob} \\
Fusion & additive & learned gate & Tab.~\ref{tab:abl-rkknob} \\
Neighbour votes & at bed starts & over whole beds & Tab.~\ref{tab:abl-rkknob} \\
\midrule
\multicolumn{4}{@{}l}{\textit{Model and data}} \\
Token & stratum (layer) & 1\,m cell & Tab.~\ref{tab:app-granularity} \\
Backbone & GRU & LSTM, Transformer, TCN, Mamba & Tab.~\ref{tab:app-task1-backbones}, \ref{tab:app-task2-backbones} \\
Inputs & lithology & + thickness, depth & Tab.~\ref{tab:abl-ladder} \\
Label smoothing & 0.3 & none & Tab.~\ref{tab:abl-ladder} \\
Grid & 1\,m & 0.5\,m, 2\,m & Tab.~\ref{tab:abl-grid} \\
Labelled holes & all of \textsc{train} & 1k, 3k, 9k & Tab.~\ref{tab:abl-dsize} \\
Learning rate & $3{\times}10^{-4}$ & shared by all backbones & -- \\
Fine-tuning length & 20 epochs & no checkpoint selection & -- \\
\midrule
\multicolumn{4}{@{}l}{\textit{Generation (Task~2 only)}} \\
Boundary-rate scale $s$ & fitted on \textsc{train} & every generator & App.~\ref{app:evaluation} \\
Lithology temperature $\tau$ & 0.7, set on \textsc{train} & -- & App.~\ref{app:evaluation} \\
\bottomrule
\end{tabular}
\end{table}

\subsubsection{What each component contributes}
\label{app:ptrk-components}

Table~\ref{tab:abl-ladder} adds the components one at a time. Label smoothing, applied to the lithology-only model,
helps on all three shifted splits at no cost in distribution. Retrieval then gives the largest single gain in
distribution and a clear gain on the distal ring, and nothing measurable across provinces. Pretraining shows the
opposite profile: nothing in distribution and a significant gain on every shifted split. Feeding thickness and depth back
as inputs helps slightly in distribution, not significantly, and costs a significant amount on \textsc{pilbara-test}.
This is the transfer boundary of \S\ref{sec:results} at a smaller scale, and the reason the deployed model reads lithology
only.

\begin{table*}[t]
\centering
\footnotesize
\setlength{\tabcolsep}{3pt}
\caption{\textbf{Cumulative components.} Layer-level GRU, 10 seeds per row; $\Delta$ vs.\ the previous row. $\pm$: half-width of the 95\% hole-cluster bootstrap interval ($B{=}1000$); trained rows also resample seeds. macro-F1 over classes with $\geq$20 events; top-1 on OOD-R / OOD-D is macro-averaged over exploration reports. $\Delta$: paired difference in macro-F1 (points) with its 95\% interval; bold when the interval excludes zero.}
\label{tab:abl-ladder}
\begin{adjustbox}{max width=\textwidth}
\begin{tabular}{@{}l c rrrr rrrr rrrr@{}}
\toprule
& & \multicolumn{4}{c}{\textbf{macro-F1 (\%)}} & \multicolumn{4}{c}{\textbf{top-1 (\%)}} & \multicolumn{4}{c}{\textbf{$\Delta$ macro-F1 vs.\ reference}} \\
\cmidrule(lr){3-6}\cmidrule(lr){7-10}\cmidrule(lr){11-14}
\textbf{Arm} & \textbf{runs} & Test & OOD-R & OOD-D & Pil-T & Test & OOD-R & OOD-D & Pil-T & Test & OOD-R & OOD-D & Pil-T \\
\midrule
A0: layer GRU, lithology only & 10 & 59.5{\tiny$\pm$4.8} & 34.9{\tiny$\pm$3.3} & 26.1{\tiny$\pm$3.8} & 26.5{\tiny$\pm$2.6} & 58.6{\tiny$\pm$1.1} & 40.2{\tiny$\pm$2.1} & 34.5{\tiny$\pm$3.1} & 40.2{\tiny$\pm$1.4} & -- & -- & -- & -- \\
A1: + label smoothing & 10 & 59.4{\tiny$\pm$5.0} & 37.4{\tiny$\pm$3.5} & 28.3{\tiny$\pm$4.0} & 28.2{\tiny$\pm$3.2} & 58.6{\tiny$\pm$1.1} & 40.5{\tiny$\pm$2.0} & 34.7{\tiny$\pm$3.2} & 42.0{\tiny$\pm$1.3} & -0.1{\tiny\,[-0.8, +0.5]} & \textbf{+2.6{\tiny\,[+1.3, +3.6]}} & \textbf{+2.2{\tiny\,[+1.2, +3.1]}} & \textbf{+1.7{\tiny\,[+0.2, +4.4]}} \\
A2: + retrieval & 10 & 70.0{\tiny$\pm$5.0} & 37.4{\tiny$\pm$3.3} & 33.9{\tiny$\pm$4.6} & 26.9{\tiny$\pm$3.0} & 73.5{\tiny$\pm$0.9} & 52.5{\tiny$\pm$2.2} & 46.9{\tiny$\pm$3.3} & 42.8{\tiny$\pm$1.1} & \textbf{+10.7{\tiny\,[+8.9, +12.6]}} & -0.1{\tiny\,[-1.1, +2.2]} & \textbf{+5.6{\tiny\,[+4.4, +7.0]}} & -1.3{\tiny\,[-3.0, +0.6]} \\
A3: + pretraining (full recipe) & 10 & 69.2{\tiny$\pm$4.9} & 38.8{\tiny$\pm$3.5} & 37.2{\tiny$\pm$4.8} & 30.6{\tiny$\pm$2.9} & 73.7{\tiny$\pm$0.9} & 52.9{\tiny$\pm$2.2} & 48.5{\tiny$\pm$3.4} & 44.3{\tiny$\pm$1.1} & -0.8{\tiny\,[-1.8, +2.8]} & \textbf{+1.4{\tiny\,[+0.5, +2.5]}} & \textbf{+3.4{\tiny\,[+2.1, +4.6]}} & \textbf{+3.7{\tiny\,[+2.3, +5.3]}} \\
A4: + thickness / depth inputs & 10 & 71.3{\tiny$\pm$4.9} & 39.6{\tiny$\pm$4.2} & 37.3{\tiny$\pm$5.0} & 29.0{\tiny$\pm$3.0} & 75.1{\tiny$\pm$0.9} & 53.1{\tiny$\pm$2.4} & 47.9{\tiny$\pm$3.0} & 44.2{\tiny$\pm$1.1} & +2.1{\tiny\,[-1.3, +3.1]} & +0.8{\tiny\,[-0.6, +4.1]} & +0.1{\tiny\,[-1.9, +1.3]} & \textbf{-1.6{\tiny\,[-3.2, -0.5]}} \\
\bottomrule
\end{tabular}
\end{adjustbox}
\end{table*}

The same division of labour holds when each component is measured on its own against the plain model
(Table~\ref{tab:app-task1-paired}). Retrieval buys in-distribution and ring accuracy, pretraining buys the shifted splits,
and the full recipe improves on the plain model on all four splits. On Task~2 (Table~\ref{tab:app-task2-paired}) neither
component changes the generated law in distribution or on the regional ring, where a sequence model trained on the
Eastern Goldfields already matches it. On the distal ring and across provinces both help. There the full recipe is the
only arm whose gain is significant on both splits, and on \textsc{pilbara-test} the two components are additive.

\subsubsection{Pretraining}
\label{app:ptrk-pretraining}
\label{app:province_holdout}

\paragraph{Corpus size.}
The pretraining corpus consists of the roughly 112k archive holes with at least two logged lithology layers. Table~\ref{tab:abl-ptscale} pretrains on nested random subsets of it for a fixed 24 epochs. In distribution the
corpus size never matters. Under shift the benefit grows with the corpus, but the splits differ in how much they need.
The distal ring gains from a few thousand holes on, whereas \textsc{pilbara-test} gains only from tens of thousands of
holes on. Below that, pretraining without retrieval is harmful across provinces, though not significantly: a
small corpus mostly repeats the Eastern Goldfields and teaches nothing that transfers.

\begin{table*}[t]
\centering
\footnotesize
\setlength{\tabcolsep}{3pt}
\caption{\textbf{Pretraining-corpus size.} Nested random subsets of the 112k-hole pretraining corpus, fine objective, 24 pretraining epochs each; $\Delta$ vs.\ no pretraining in the same retrieval setting. $\pm$: half-width of the 95\% hole-cluster bootstrap interval ($B{=}1000$); trained rows also resample seeds. macro-F1 over classes with $\geq$20 events; top-1 on OOD-R / OOD-D is macro-averaged over exploration reports. $\Delta$: paired difference in macro-F1 (points) with its 95\% interval; bold when the interval excludes zero.}
\label{tab:abl-ptscale}
\begin{adjustbox}{max width=\textwidth}
\begin{tabular}{@{}l c rrrr rrrr rrrr@{}}
\toprule
& & \multicolumn{4}{c}{\textbf{macro-F1 (\%)}} & \multicolumn{4}{c}{\textbf{top-1 (\%)}} & \multicolumn{4}{c}{\textbf{$\Delta$ macro-F1 vs.\ reference}} \\
\cmidrule(lr){3-6}\cmidrule(lr){7-10}\cmidrule(lr){11-14}
\textbf{Arm} & \textbf{runs} & Test & OOD-R & OOD-D & Pil-T & Test & OOD-R & OOD-D & Pil-T & Test & OOD-R & OOD-D & Pil-T \\
\midrule
no PT, no R & 3 & 59.5{\tiny$\pm$4.9} & 37.2{\tiny$\pm$3.6} & 28.5{\tiny$\pm$4.2} & 29.0{\tiny$\pm$3.0} & 58.5{\tiny$\pm$1.1} & 41.4{\tiny$\pm$2.4} & 36.3{\tiny$\pm$4.3} & 42.0{\tiny$\pm$1.4} & -- & -- & -- & -- \\
1.1k, no R & 3 & 59.4{\tiny$\pm$4.9} & 38.2{\tiny$\pm$3.3} & 29.9{\tiny$\pm$4.3} & 26.5{\tiny$\pm$3.7} & 58.5{\tiny$\pm$1.1} & 41.6{\tiny$\pm$2.3} & 36.3{\tiny$\pm$4.6} & 41.0{\tiny$\pm$1.4} & -0.1{\tiny\,[-1.4, +1.2]} & \textbf{+1.0{\tiny\,[+0.1, +2.2]}} & +1.3{\tiny\,[-0.3, +2.9]} & -2.5{\tiny\,[-4.6, +0.4]} \\
3.4k, no R & 3 & 59.2{\tiny$\pm$5.0} & 38.0{\tiny$\pm$3.4} & 30.6{\tiny$\pm$4.5} & 25.2{\tiny$\pm$3.2} & 58.5{\tiny$\pm$1.1} & 41.6{\tiny$\pm$2.4} & 36.3{\tiny$\pm$4.3} & 41.2{\tiny$\pm$1.5} & -0.3{\tiny\,[-1.6, +1.0]} & +0.8{\tiny\,[-0.4, +2.0]} & \textbf{+2.1{\tiny\,[+0.5, +3.9]}} & -3.8{\tiny\,[-5.0, +0.7]} \\
11k, no R & 3 & 59.4{\tiny$\pm$4.9} & 38.7{\tiny$\pm$3.6} & 30.4{\tiny$\pm$4.4} & 26.8{\tiny$\pm$3.2} & 58.5{\tiny$\pm$1.1} & 41.7{\tiny$\pm$2.3} & 36.4{\tiny$\pm$4.3} & 42.7{\tiny$\pm$1.4} & -0.1{\tiny\,[-1.2, +1.0]} & \textbf{+1.5{\tiny\,[+0.3, +2.8]}} & \textbf{+1.8{\tiny\,[+0.5, +3.3]}} & -2.2{\tiny\,[-3.6, +2.5]} \\
34k, no R & 3 & 59.1{\tiny$\pm$5.0} & 39.1{\tiny$\pm$3.5} & 32.1{\tiny$\pm$4.6} & 29.5{\tiny$\pm$3.6} & 58.5{\tiny$\pm$1.1} & 41.7{\tiny$\pm$2.4} & 36.7{\tiny$\pm$4.2} & 43.0{\tiny$\pm$1.3} & -0.4{\tiny\,[-1.7, +1.0]} & \textbf{+1.9{\tiny\,[+1.0, +3.9]}} & \textbf{+3.5{\tiny\,[+2.0, +5.0]}} & +0.6{\tiny\,[-1.8, +2.6]} \\
67k, no R & 3 & 59.0{\tiny$\pm$5.0} & 39.0{\tiny$\pm$3.3} & 34.1{\tiny$\pm$4.7} & 31.7{\tiny$\pm$3.2} & 58.5{\tiny$\pm$1.1} & 42.0{\tiny$\pm$2.5} & 37.3{\tiny$\pm$4.4} & 43.5{\tiny$\pm$1.5} & -0.5{\tiny\,[-1.9, +1.1]} & \textbf{+1.8{\tiny\,[+1.1, +4.1]}} & \textbf{+5.5{\tiny\,[+3.7, +7.1]}} & \textbf{+2.7{\tiny\,[+1.3, +4.6]}} \\
112k, no R & 3 & 58.7{\tiny$\pm$5.1} & 39.1{\tiny$\pm$3.3} & 32.9{\tiny$\pm$4.7} & 31.8{\tiny$\pm$3.1} & 58.5{\tiny$\pm$1.1} & 42.1{\tiny$\pm$2.3} & 37.1{\tiny$\pm$4.5} & 43.5{\tiny$\pm$1.6} & -0.8{\tiny\,[-2.2, +0.8]} & \textbf{+1.9{\tiny\,[+1.2, +3.5]}} & \textbf{+4.4{\tiny\,[+2.8, +6.0]}} & \textbf{+2.8{\tiny\,[+1.5, +4.6]}} \\
\midrule
no PT, + R & 3 & 69.5{\tiny$\pm$5.1} & 38.3{\tiny$\pm$3.7} & 33.9{\tiny$\pm$4.6} & 26.9{\tiny$\pm$3.4} & 73.4{\tiny$\pm$1.0} & 52.4{\tiny$\pm$2.2} & 46.2{\tiny$\pm$4.4} & 42.8{\tiny$\pm$1.2} & -- & -- & -- & -- \\
1.1k, + R & 3 & 69.3{\tiny$\pm$5.1} & 38.4{\tiny$\pm$3.6} & 35.7{\tiny$\pm$4.8} & 26.0{\tiny$\pm$3.6} & 73.4{\tiny$\pm$0.9} & 52.3{\tiny$\pm$2.3} & 46.1{\tiny$\pm$4.4} & 42.8{\tiny$\pm$1.3} & -0.2{\tiny\,[-1.5, +0.9]} & +0.1{\tiny\,[-0.7, +1.0]} & +1.7{\tiny\,[-0.1, +3.5]} & -0.9{\tiny\,[-3.7, +2.7]} \\
3.4k, + R & 3 & 69.5{\tiny$\pm$5.0} & 38.6{\tiny$\pm$3.7} & 35.8{\tiny$\pm$4.8} & 25.7{\tiny$\pm$3.5} & 73.4{\tiny$\pm$1.0} & 52.5{\tiny$\pm$2.2} & 46.2{\tiny$\pm$4.5} & 42.7{\tiny$\pm$1.4} & +0.1{\tiny\,[-1.3, +1.0]} & +0.3{\tiny\,[-0.6, +1.5]} & \textbf{+1.9{\tiny\,[+0.1, +3.7]}} & -1.1{\tiny\,[-3.8, +2.4]} \\
11k, + R & 3 & 70.1{\tiny$\pm$5.0} & 38.9{\tiny$\pm$3.7} & 37.1{\tiny$\pm$5.1} & 27.7{\tiny$\pm$3.5} & 73.6{\tiny$\pm$0.9} & 53.0{\tiny$\pm$2.2} & 48.6{\tiny$\pm$4.0} & 43.4{\tiny$\pm$1.2} & +0.6{\tiny\,[-0.6, +1.3]} & +0.6{\tiny\,[-0.5, +2.7]} & \textbf{+3.1{\tiny\,[+1.4, +5.0]}} & +0.8{\tiny\,[-2.1, +4.1]} \\
34k, + R & 3 & 69.7{\tiny$\pm$5.1} & 38.7{\tiny$\pm$3.5} & 38.1{\tiny$\pm$5.0} & 30.1{\tiny$\pm$2.9} & 73.5{\tiny$\pm$1.0} & 53.0{\tiny$\pm$2.3} & 48.8{\tiny$\pm$4.0} & 44.2{\tiny$\pm$1.1} & +0.2{\tiny\,[-1.0, +1.1]} & +0.4{\tiny\,[-0.3, +1.6]} & \textbf{+4.2{\tiny\,[+2.5, +5.9]}} & \textbf{+3.2{\tiny\,[+1.0, +6.1]}} \\
67k, + R & 3 & 69.3{\tiny$\pm$4.8} & 39.9{\tiny$\pm$3.9} & 38.4{\tiny$\pm$5.1} & 31.1{\tiny$\pm$2.9} & 73.6{\tiny$\pm$0.9} & 53.1{\tiny$\pm$2.2} & 49.1{\tiny$\pm$4.1} & 44.7{\tiny$\pm$1.2} & -0.2{\tiny\,[-1.5, +3.5]} & \textbf{+1.6{\tiny\,[+0.6, +3.7]}} & \textbf{+4.4{\tiny\,[+2.5, +6.2]}} & \textbf{+4.3{\tiny\,[+2.0, +7.2]}} \\
112k, + R & 3 & 69.1{\tiny$\pm$5.2} & 40.0{\tiny$\pm$4.0} & 37.3{\tiny$\pm$4.9} & 31.2{\tiny$\pm$2.9} & 73.7{\tiny$\pm$0.9} & 53.1{\tiny$\pm$2.3} & 49.2{\tiny$\pm$4.0} & 44.6{\tiny$\pm$1.2} & -0.4{\tiny\,[-1.7, +2.4]} & \textbf{+1.7{\tiny\,[+0.5, +3.9]}} & \textbf{+3.3{\tiny\,[+1.8, +5.1]}} & \textbf{+4.3{\tiny\,[+2.1, +7.4]}} \\
\bottomrule
\end{tabular}
\end{adjustbox}
\end{table*}

\paragraph{Data or optimisation?}
A fixed epoch count also gives a small corpus fewer optimisation steps. Table~\ref{tab:abl-ptcmatch} holds the number of
steps at that of the full corpus. With equal compute the curve saturates at about a tenth of the corpus. Apart from one significant dip across provinces
at 30\% without retrieval, only the two smallest fractions remain clearly worse. Much of the climb in Table~\ref{tab:abl-ptscale} beyond that point is
therefore optimisation rather than data. What the data itself buys is concentrated in the smallest corpora.

\begin{table*}[t]
\centering
\footnotesize
\setlength{\tabcolsep}{3pt}
\caption{\textbf{Compute-matched corpus size.} Corpus fractions pretrained for $24/\text{fraction}$ epochs, so that every checkpoint takes the same number of optimisation steps; $\Delta$ vs.\ the deployed full-corpus checkpoint in the same retrieval setting. $\pm$: half-width of the 95\% hole-cluster bootstrap interval ($B{=}1000$); trained rows also resample seeds. macro-F1 over classes with $\geq$20 events; top-1 on OOD-R / OOD-D is macro-averaged over exploration reports. $\Delta$: paired difference in macro-F1 (points) with its 95\% interval; bold when the interval excludes zero.}
\label{tab:abl-ptcmatch}
\begin{adjustbox}{max width=\textwidth}
\begin{tabular}{@{}l c rrrr rrrr rrrr@{}}
\toprule
& & \multicolumn{4}{c}{\textbf{macro-F1 (\%)}} & \multicolumn{4}{c}{\textbf{top-1 (\%)}} & \multicolumn{4}{c}{\textbf{$\Delta$ macro-F1 vs.\ reference}} \\
\cmidrule(lr){3-6}\cmidrule(lr){7-10}\cmidrule(lr){11-14}
\textbf{Arm} & \textbf{runs} & Test & OOD-R & OOD-D & Pil-T & Test & OOD-R & OOD-D & Pil-T & Test & OOD-R & OOD-D & Pil-T \\
\midrule
fine + chem (deployed), no R & 10 & 58.5{\tiny$\pm$5.1} & 39.2{\tiny$\pm$3.3} & 33.0{\tiny$\pm$4.5} & 31.7{\tiny$\pm$2.9} & 58.6{\tiny$\pm$1.1} & 40.8{\tiny$\pm$2.1} & 35.4{\tiny$\pm$2.9} & 44.0{\tiny$\pm$1.2} & -- & -- & -- & -- \\
1\% (2400 ep), no R & 3 & 58.6{\tiny$\pm$4.9} & 35.6{\tiny$\pm$3.7} & 27.5{\tiny$\pm$3.8} & 28.1{\tiny$\pm$2.9} & 58.2{\tiny$\pm$1.1} & 41.1{\tiny$\pm$2.4} & 34.9{\tiny$\pm$4.4} & 41.8{\tiny$\pm$1.5} & +0.1{\tiny\,[-1.1, +1.2]} & \textbf{-3.6{\tiny\,[-6.5, -2.5]}} & \textbf{-5.5{\tiny\,[-7.4, -2.7]}} & \textbf{-3.6{\tiny\,[-5.6, -2.1]}} \\
3\% (800 ep), no R & 3 & 58.7{\tiny$\pm$5.1} & 39.0{\tiny$\pm$3.6} & 29.4{\tiny$\pm$4.4} & 25.4{\tiny$\pm$3.1} & 58.5{\tiny$\pm$1.1} & 41.4{\tiny$\pm$2.4} & 35.8{\tiny$\pm$4.3} & 41.5{\tiny$\pm$1.4} & +0.2{\tiny\,[-0.6, +1.2]} & -0.2{\tiny\,[-1.9, +0.5]} & \textbf{-3.6{\tiny\,[-5.0, -2.1]}} & \textbf{-6.3{\tiny\,[-7.7, -2.7]}} \\
10\% (240 ep), no R & 3 & 57.9{\tiny$\pm$5.1} & 39.2{\tiny$\pm$3.8} & 33.2{\tiny$\pm$4.7} & 31.6{\tiny$\pm$2.9} & 58.5{\tiny$\pm$1.1} & 40.7{\tiny$\pm$2.2} & 35.5{\tiny$\pm$4.6} & 43.6{\tiny$\pm$1.6} & -0.5{\tiny\,[-1.3, +1.2]} & +0.0{\tiny\,[-1.8, +1.0]} & +0.2{\tiny\,[-1.1, +1.5]} & -0.1{\tiny\,[-1.6, +1.1]} \\
30\% (80 ep), no R & 3 & 58.7{\tiny$\pm$5.1} & 39.9{\tiny$\pm$3.6} & 33.5{\tiny$\pm$4.6} & 29.7{\tiny$\pm$3.0} & 58.5{\tiny$\pm$1.1} & 42.2{\tiny$\pm$2.4} & 37.3{\tiny$\pm$4.5} & 42.6{\tiny$\pm$1.5} & +0.3{\tiny\,[-0.3, +0.9]} & +0.7{\tiny\,[-0.5, +1.3]} & +0.5{\tiny\,[-0.8, +1.7]} & \textbf{-2.0{\tiny\,[-2.9, -1.0]}} \\
60\% (40 ep), no R & 3 & 58.7{\tiny$\pm$5.1} & 39.3{\tiny$\pm$3.6} & 33.4{\tiny$\pm$4.5} & 31.8{\tiny$\pm$2.9} & 58.6{\tiny$\pm$1.1} & 42.1{\tiny$\pm$2.4} & 37.2{\tiny$\pm$4.1} & 43.7{\tiny$\pm$1.4} & +0.3{\tiny\,[-0.7, +1.1]} & +0.1{\tiny\,[-0.8, +0.8]} & +0.4{\tiny\,[-0.8, +1.5]} & +0.0{\tiny\,[-0.8, +1.0]} \\
\midrule
fine + chem (deployed), + R & 10 & 69.2{\tiny$\pm$5.0} & 38.8{\tiny$\pm$3.5} & 37.2{\tiny$\pm$4.9} & 30.6{\tiny$\pm$3.0} & 73.7{\tiny$\pm$0.9} & 52.9{\tiny$\pm$2.2} & 48.5{\tiny$\pm$3.3} & 44.3{\tiny$\pm$1.1} & -- & -- & -- & -- \\
1\% (2400 ep), + R & 3 & 68.0{\tiny$\pm$5.3} & 33.9{\tiny$\pm$3.4} & 31.6{\tiny$\pm$4.3} & 27.3{\tiny$\pm$2.7} & 73.2{\tiny$\pm$1.0} & 51.2{\tiny$\pm$2.5} & 48.1{\tiny$\pm$2.7} & 42.7{\tiny$\pm$1.1} & \textbf{-1.3{\tiny\,[-3.9, -0.3]}} & \textbf{-4.9{\tiny\,[-7.5, -3.2]}} & \textbf{-5.6{\tiny\,[-7.7, -3.1]}} & \textbf{-3.4{\tiny\,[-5.6, -2.0]}} \\
3\% (800 ep), + R & 3 & 69.3{\tiny$\pm$5.3} & 38.3{\tiny$\pm$3.4} & 36.1{\tiny$\pm$5.1} & 24.2{\tiny$\pm$2.7} & 73.7{\tiny$\pm$1.0} & 52.0{\tiny$\pm$2.5} & 50.1{\tiny$\pm$2.9} & 41.8{\tiny$\pm$1.2} & +0.1{\tiny\,[-3.0, +0.9]} & -0.5{\tiny\,[-1.9, +0.4]} & -1.1{\tiny\,[-2.7, +0.7]} & \textbf{-6.5{\tiny\,[-7.4, -2.5]}} \\
10\% (240 ep), + R & 3 & 68.5{\tiny$\pm$5.2} & 40.1{\tiny$\pm$3.7} & 37.9{\tiny$\pm$5.1} & 30.9{\tiny$\pm$2.8} & 73.2{\tiny$\pm$1.0} & 51.8{\tiny$\pm$2.4} & 47.3{\tiny$\pm$4.3} & 44.4{\tiny$\pm$1.2} & -0.8{\tiny\,[-1.6, +1.5]} & +1.3{\tiny\,[-0.1, +2.2]} & +0.6{\tiny\,[-0.6, +2.0]} & +0.3{\tiny\,[-1.0, +1.3]} \\
30\% (80 ep), + R & 3 & 68.9{\tiny$\pm$5.4} & 39.4{\tiny$\pm$3.7} & 38.0{\tiny$\pm$4.9} & 30.1{\tiny$\pm$3.0} & 73.6{\tiny$\pm$0.9} & 53.0{\tiny$\pm$2.3} & 49.4{\tiny$\pm$4.2} & 44.1{\tiny$\pm$1.2} & -0.3{\tiny\,[-3.3, +0.4]} & +0.7{\tiny\,[-0.6, +1.5]} & +0.7{\tiny\,[-0.0, +1.4]} & -0.5{\tiny\,[-1.8, +0.6]} \\
60\% (40 ep), + R & 3 & 69.0{\tiny$\pm$5.2} & 39.1{\tiny$\pm$3.7} & 37.8{\tiny$\pm$4.9} & 30.3{\tiny$\pm$3.1} & 73.6{\tiny$\pm$1.0} & 52.6{\tiny$\pm$2.4} & 49.2{\tiny$\pm$4.3} & 44.6{\tiny$\pm$1.2} & -0.3{\tiny\,[-3.4, +0.2]} & +0.3{\tiny\,[-0.9, +1.2]} & +0.6{\tiny\,[-0.0, +1.2]} & -0.4{\tiny\,[-0.9, +0.5]} \\
\bottomrule
\end{tabular}
\end{adjustbox}
\end{table*}

\paragraph{Province diversity.}
Subsampling also removes provinces. Table~\ref{tab:abl-ptprov} fixes the corpus at 60k holes and draws it from one to
thirty provinces. \textsc{pilbara-test} responds to diversity. Its deficit against the full archive is largest for a
one- or two-province corpus and at least halves at thirty provinces, while the in-distribution score does not move. The
distal ring, which lies in the training province, responds much less. Volume alone does not transfer across provinces;
geological breadth does.

\begin{table*}[t]
\centering
\footnotesize
\setlength{\tabcolsep}{3pt}
\caption{\textbf{Province diversity of the pretraining corpus.} 60k-hole corpora drawn from the $n$ largest provinces; $\Delta$ vs.\ the deployed full-corpus checkpoint in the same retrieval setting. $\pm$: half-width of the 95\% hole-cluster bootstrap interval ($B{=}1000$); trained rows also resample seeds. macro-F1 over classes with $\geq$20 events; top-1 on OOD-R / OOD-D is macro-averaged over exploration reports. $\Delta$: paired difference in macro-F1 (points) with its 95\% interval; bold when the interval excludes zero.}
\label{tab:abl-ptprov}
\begin{adjustbox}{max width=\textwidth}
\begin{tabular}{@{}l c rrrr rrrr rrrr@{}}
\toprule
& & \multicolumn{4}{c}{\textbf{macro-F1 (\%)}} & \multicolumn{4}{c}{\textbf{top-1 (\%)}} & \multicolumn{4}{c}{\textbf{$\Delta$ macro-F1 vs.\ reference}} \\
\cmidrule(lr){3-6}\cmidrule(lr){7-10}\cmidrule(lr){11-14}
\textbf{Arm} & \textbf{runs} & Test & OOD-R & OOD-D & Pil-T & Test & OOD-R & OOD-D & Pil-T & Test & OOD-R & OOD-D & Pil-T \\
\midrule
fine + chem (deployed), no R & 10 & 58.5{\tiny$\pm$5.1} & 39.2{\tiny$\pm$3.3} & 33.0{\tiny$\pm$4.5} & 31.7{\tiny$\pm$2.9} & 58.6{\tiny$\pm$1.1} & 40.8{\tiny$\pm$2.1} & 35.4{\tiny$\pm$2.9} & 44.0{\tiny$\pm$1.2} & -- & -- & -- & -- \\
1 province, no R & 3 & 59.6{\tiny$\pm$4.9} & 38.4{\tiny$\pm$3.5} & 30.3{\tiny$\pm$4.5} & 25.3{\tiny$\pm$3.1} & 58.5{\tiny$\pm$1.1} & 41.7{\tiny$\pm$2.3} & 36.4{\tiny$\pm$4.5} & 42.0{\tiny$\pm$1.4} & +1.2{\tiny\,[-0.8, +2.3]} & \textbf{-0.8{\tiny\,[-2.1, -0.1]}} & \textbf{-2.7{\tiny\,[-4.2, -1.3]}} & \textbf{-6.4{\tiny\,[-7.5, -2.4]}} \\
2 provinces, no R & 3 & 59.2{\tiny$\pm$4.9} & 38.7{\tiny$\pm$3.8} & 30.6{\tiny$\pm$4.3} & 26.4{\tiny$\pm$3.2} & 58.5{\tiny$\pm$1.1} & 41.6{\tiny$\pm$2.4} & 36.5{\tiny$\pm$4.1} & 42.2{\tiny$\pm$1.3} & +0.7{\tiny\,[-0.8, +2.1]} & -0.4{\tiny\,[-2.3, +0.1]} & \textbf{-2.4{\tiny\,[-3.7, -1.2]}} & \textbf{-5.4{\tiny\,[-6.7, -1.7]}} \\
4 provinces, no R & 3 & 59.0{\tiny$\pm$5.1} & 38.8{\tiny$\pm$3.6} & 31.9{\tiny$\pm$4.5} & 26.7{\tiny$\pm$3.1} & 58.5{\tiny$\pm$1.1} & 41.6{\tiny$\pm$2.4} & 36.6{\tiny$\pm$4.2} & 42.7{\tiny$\pm$1.5} & +0.6{\tiny\,[-0.9, +1.8]} & -0.4{\tiny\,[-1.6, +0.2]} & -1.2{\tiny\,[-2.5, +0.0]} & \textbf{-5.0{\tiny\,[-6.4, -1.5]}} \\
8 provinces, no R & 3 & 59.4{\tiny$\pm$5.0} & 39.0{\tiny$\pm$3.6} & 30.6{\tiny$\pm$4.4} & 26.4{\tiny$\pm$3.2} & 58.5{\tiny$\pm$1.1} & 41.7{\tiny$\pm$2.3} & 36.4{\tiny$\pm$4.2} & 42.3{\tiny$\pm$1.4} & +0.9{\tiny\,[-0.5, +2.0]} & -0.2{\tiny\,[-1.5, +0.4]} & \textbf{-2.4{\tiny\,[-3.7, -1.2]}} & \textbf{-5.4{\tiny\,[-6.5, -1.4]}} \\
30 provinces, no R & 3 & 59.2{\tiny$\pm$4.9} & 38.2{\tiny$\pm$3.3} & 30.0{\tiny$\pm$4.4} & 28.5{\tiny$\pm$3.5} & 58.5{\tiny$\pm$1.1} & 41.5{\tiny$\pm$2.5} & 36.1{\tiny$\pm$4.6} & 42.5{\tiny$\pm$1.4} & +0.8{\tiny\,[-0.8, +2.0]} & \textbf{-1.0{\tiny\,[-2.0, -0.3]}} & \textbf{-3.0{\tiny\,[-4.6, -1.5]}} & \textbf{-3.2{\tiny\,[-5.3, -1.4]}} \\
\midrule
fine + chem (deployed), + R & 10 & 69.2{\tiny$\pm$5.0} & 38.8{\tiny$\pm$3.5} & 37.2{\tiny$\pm$4.9} & 30.6{\tiny$\pm$3.0} & 73.7{\tiny$\pm$0.9} & 52.9{\tiny$\pm$2.2} & 48.5{\tiny$\pm$3.3} & 44.3{\tiny$\pm$1.1} & -- & -- & -- & -- \\
1 province, + R & 3 & 69.7{\tiny$\pm$5.0} & 38.3{\tiny$\pm$3.6} & 35.1{\tiny$\pm$4.6} & 26.6{\tiny$\pm$3.6} & 73.4{\tiny$\pm$0.9} & 52.3{\tiny$\pm$2.2} & 46.3{\tiny$\pm$4.3} & 43.5{\tiny$\pm$1.4} & +0.4{\tiny\,[-3.0, +1.6]} & -0.5{\tiny\,[-1.8, +0.7]} & \textbf{-2.2{\tiny\,[-3.6, -0.9]}} & \textbf{-4.1{\tiny\,[-6.7, -1.4]}} \\
2 provinces, + R & 3 & 69.7{\tiny$\pm$4.9} & 39.0{\tiny$\pm$3.7} & 35.9{\tiny$\pm$4.9} & 25.9{\tiny$\pm$3.0} & 73.5{\tiny$\pm$1.0} & 53.0{\tiny$\pm$2.1} & 48.1{\tiny$\pm$3.9} & 43.1{\tiny$\pm$1.4} & +0.4{\tiny\,[-3.1, +1.6]} & +0.2{\tiny\,[-1.6, +1.5]} & -1.3{\tiny\,[-2.9, +0.1]} & \textbf{-4.8{\tiny\,[-6.3, -1.7]}} \\
4 provinces, + R & 3 & 69.5{\tiny$\pm$5.0} & 38.5{\tiny$\pm$3.5} & 38.0{\tiny$\pm$5.0} & 27.4{\tiny$\pm$3.3} & 73.5{\tiny$\pm$1.0} & 52.9{\tiny$\pm$2.2} & 48.8{\tiny$\pm$4.1} & 43.8{\tiny$\pm$1.2} & +0.3{\tiny\,[-3.2, +1.3]} & -0.3{\tiny\,[-1.4, +0.4]} & +0.8{\tiny\,[-0.5, +2.1]} & \textbf{-3.2{\tiny\,[-5.5, -1.2]}} \\
8 provinces, + R & 3 & 70.0{\tiny$\pm$5.0} & 39.1{\tiny$\pm$3.8} & 36.8{\tiny$\pm$4.9} & 29.1{\tiny$\pm$3.0} & 73.6{\tiny$\pm$0.9} & 52.8{\tiny$\pm$2.2} & 48.6{\tiny$\pm$4.0} & 44.1{\tiny$\pm$1.2} & +0.8{\tiny\,[-2.8, +1.8]} & +0.3{\tiny\,[-1.8, +1.7]} & -0.4{\tiny\,[-2.1, +0.9]} & \textbf{-1.6{\tiny\,[-2.7, -0.4]}} \\
30 provinces, + R & 3 & 70.2{\tiny$\pm$5.0} & 39.2{\tiny$\pm$3.7} & 36.7{\tiny$\pm$5.0} & 29.4{\tiny$\pm$3.0} & 73.6{\tiny$\pm$1.0} & 53.0{\tiny$\pm$2.2} & 48.3{\tiny$\pm$4.2} & 43.8{\tiny$\pm$1.2} & +1.0{\tiny\,[-2.6, +1.9]} & +0.4{\tiny\,[-1.3, +1.5]} & -0.6{\tiny\,[-2.3, +1.0]} & \textbf{-1.2{\tiny\,[-2.3, -0.2]}} \\
\bottomrule
\end{tabular}
\end{adjustbox}
\end{table*}

\paragraph{Held-out provinces.}
The archive covers all of Western Australia and therefore contains unlabelled Pilbara holes. A gain on
\textsc{pilbara-test} could reflect a transferable prior or mere exposure to the target province. We repeat pretraining,
with an identical recipe, on two reduced archives. \emph{Pilbara held out} removes every archive hole inside the bounding
box of the \textsc{pilbara-test} collars dilated by 100\,km, about a quarter of the archive; the nearest remaining hole
is more than 100\,km from any Pilbara collar. \emph{Pilbara and EG held out} also removes the bounding box of the two OOD
rings dilated by 100\,km, which covers the training footprint as well, so that pretraining sees neither province
evaluated in this paper. This removes about half of the archive. Fine-tuning, seeds and the retrieval pool are unchanged.
Most of the cross-province gain survives. With retrieval, four fifths or more of the \textsc{pilbara-test} gain remains
($82\%$ with Pilbara held out, $96\%$ with the Eastern Goldfields held out as well). Without retrieval the gain over the
label-smoothed GRU is smaller to begin with, and holding Pilbara out keeps about half of it ($45\%$; $74\%$ with the
Eastern Goldfields also held out). Top-1 accuracy on \textsc{pilbara-test} does not change beyond seed noise
(Table~\ref{tab:province_holdout}). Removing the Eastern Goldfields as well costs nothing further, and the rings lose at
most about a point. The gain therefore largely reflects stratigraphic structure learned from other provinces. The residual
difference is the part that does come from having read the target province, and it is what a practitioner gains by
pretraining on the full archive.

\begin{table}[h]
\centering
\footnotesize
\setlength{\tabcolsep}{3pt}
\caption{\textbf{Province-held-out pretraining}, Task~1 on DrillBench, layer-level GRU, 10 seeds. Same convention as Table~\ref{tab:app-task1-backbones}: macro-F1 over classes with $\geq$20 events, top-1 macro-averaged over reports on the two rings; $\pm$: half-width of the 95\% hole-cluster bootstrap interval. $\Delta$ rows: paired difference to the full-archive row above (same resample); intervals that exclude zero are bold. All rows use label smoothing ($\varepsilon{=}0.3$).}
\label{tab:province_holdout}
\begin{adjustbox}{width=\linewidth}
\begin{tabular}{@{}l rrrr rrrr@{}}
\toprule
& \multicolumn{4}{c}{\textbf{macro-F1 (\%)}} & \multicolumn{4}{c}{\textbf{top-1 (\%)}} \\
\cmidrule(lr){2-5}\cmidrule(lr){6-9}
\textbf{Pretraining corpus} & Test & OOD-R & OOD-D & Pil-T & Test & OOD-R & OOD-D & Pil-T \\
\midrule
\multicolumn{9}{@{}l}{\textit{Without retrieval}} \\
None (GRU) & 59.4{\tiny$\pm$4.9} & 37.4{\tiny$\pm$3.5} & 28.3{\tiny$\pm$3.9} & 28.2{\tiny$\pm$3.0} & 58.6{\tiny$\pm$1.1} & 40.5{\tiny$\pm$1.9} & 34.7{\tiny$\pm$3.1} & 42.0{\tiny$\pm$1.4} \\
Full archive (GRU + PT) & 58.5{\tiny$\pm$5.1} & 39.2{\tiny$\pm$3.2} & 33.0{\tiny$\pm$4.3} & 31.7{\tiny$\pm$2.9} & 58.6{\tiny$\pm$1.1} & 40.8{\tiny$\pm$1.8} & 35.4{\tiny$\pm$3.0} & 44.0{\tiny$\pm$1.3} \\
Pilbara held out & 58.1{\tiny$\pm$5.1} & 38.3{\tiny$\pm$3.4} & 32.1{\tiny$\pm$4.2} & 29.8{\tiny$\pm$2.9} & 58.5{\tiny$\pm$1.1} & 40.7{\tiny$\pm$1.9} & 35.2{\tiny$\pm$3.0} & 43.8{\tiny$\pm$1.3} \\
\quad $\Delta$ vs.\ full archive & -0.3{\tiny\,[-1.0, +0.4]} & \textbf{-0.9{\tiny\,[-1.9, -0.1]}} & -0.9{\tiny\,[-1.9, +0.0]} & \textbf{-1.9{\tiny\,[-2.6, -0.9]}} & -0.1{\tiny\,[-0.3, +0.2]} & -0.1{\tiny\,[-1.1, +0.9]} & -0.2{\tiny\,[-3.4, +2.7]} & -0.2{\tiny\,[-1.2, +0.7]} \\
Pilbara and EG held out & 58.5{\tiny$\pm$5.2} & 38.5{\tiny$\pm$3.3} & 31.9{\tiny$\pm$4.1} & 30.8{\tiny$\pm$2.7} & 58.6{\tiny$\pm$1.1} & 40.9{\tiny$\pm$1.9} & 35.6{\tiny$\pm$3.1} & 43.4{\tiny$\pm$1.3} \\
\quad $\Delta$ vs.\ full archive & -0.0{\tiny\,[-0.9, +0.8]} & -0.7{\tiny\,[-1.6, +0.2]} & \textbf{-1.1{\tiny\,[-1.9, -0.3]}} & -0.9{\tiny\,[-3.0, +0.2]} & -0.0{\tiny\,[-0.3, +0.2]} & +0.1{\tiny\,[-1.0, +1.0]} & +0.2{\tiny\,[-2.6, +3.8]} & -0.6{\tiny\,[-1.6, +0.4]} \\
\addlinespace[3pt]
\multicolumn{9}{@{}l}{\textit{With retrieval}} \\
None (GRU + R) & 70.0{\tiny$\pm$5.0} & 37.4{\tiny$\pm$3.4} & 33.9{\tiny$\pm$4.5} & 26.9{\tiny$\pm$3.1} & 73.5{\tiny$\pm$0.9} & 52.5{\tiny$\pm$2.0} & 46.9{\tiny$\pm$3.3} & 42.8{\tiny$\pm$1.2} \\
Full archive (GRU + PT + R) & 69.2{\tiny$\pm$4.9} & 38.8{\tiny$\pm$3.5} & 37.2{\tiny$\pm$4.6} & 30.6{\tiny$\pm$3.0} & 73.7{\tiny$\pm$0.9} & 52.9{\tiny$\pm$2.1} & 48.5{\tiny$\pm$3.3} & 44.3{\tiny$\pm$1.2} \\
Pilbara held out & 69.5{\tiny$\pm$5.1} & 38.6{\tiny$\pm$3.3} & 37.5{\tiny$\pm$4.8} & 30.0{\tiny$\pm$2.8} & 73.6{\tiny$\pm$0.9} & 52.5{\tiny$\pm$2.1} & 48.5{\tiny$\pm$3.2} & 44.4{\tiny$\pm$1.2} \\
\quad $\Delta$ vs.\ full archive & +0.3{\tiny\,[-3.0, +0.8]} & -0.2{\tiny\,[-0.9, +0.7]} & +0.3{\tiny\,[-0.7, +1.2]} & -0.7{\tiny\,[-1.8, +0.2]} & -0.1{\tiny\,[-0.5, +0.3]} & -0.4{\tiny\,[-1.2, +0.3]} & -0.0{\tiny\,[-2.7, +2.6]} & +0.1{\tiny\,[-0.5, +0.6]} \\
Pilbara and EG held out & 69.5{\tiny$\pm$5.3} & 39.3{\tiny$\pm$3.5} & 37.1{\tiny$\pm$4.7} & 30.5{\tiny$\pm$2.5} & 73.6{\tiny$\pm$0.9} & 52.4{\tiny$\pm$2.2} & 48.3{\tiny$\pm$3.2} & 44.9{\tiny$\pm$1.2} \\
\quad $\Delta$ vs.\ full archive & +0.2{\tiny\,[-2.6, +0.8]} & +0.5{\tiny\,[-0.6, +1.3]} & -0.2{\tiny\,[-1.2, +0.8]} & -0.2{\tiny\,[-2.1, +0.9]} & -0.1{\tiny\,[-0.5, +0.3]} & -0.5{\tiny\,[-1.4, +0.1]} & -0.2{\tiny\,[-2.8, +2.2]} & +0.6{\tiny\,[-0.1, +1.2]} \\
\addlinespace[3pt]
\bottomrule
\end{tabular}
\end{adjustbox}
\end{table}

\paragraph{Objective.}
Table~\ref{tab:abl-ptobj} varies the pretraining targets on the full corpus. The fine-grained lithology objective is the
component that matters. Replacing it by the 13-class coarse objective costs a significant amount on all three shifted
splits, with or without retrieval and with or without chemistry. The chemistry losses themselves are not measurable at
this scale: switching them off makes no split worse, and the one significant difference, on the regional ring without
retrieval, favours switching them off. We keep them in the released
recipe because they do no harm and make the geochemistry channel of the archive available, but no transfer claim rests on
them.

\begin{table*}[t]
\centering
\footnotesize
\setlength{\tabcolsep}{3pt}
\caption{\textbf{Pretraining objective.} Full corpus; fine: 37-code targets with coarse fallback; coarse: 13-class targets; chem: auxiliary geochemistry losses. $\Delta$ vs.\ the deployed checkpoint (fine + chem) in the same retrieval setting. $\pm$: half-width of the 95\% hole-cluster bootstrap interval ($B{=}1000$); trained rows also resample seeds. macro-F1 over classes with $\geq$20 events; top-1 on OOD-R / OOD-D is macro-averaged over exploration reports. $\Delta$: paired difference in macro-F1 (points) with its 95\% interval; bold when the interval excludes zero.}
\label{tab:abl-ptobj}
\begin{adjustbox}{max width=\textwidth}
\begin{tabular}{@{}l c rrrr rrrr rrrr@{}}
\toprule
& & \multicolumn{4}{c}{\textbf{macro-F1 (\%)}} & \multicolumn{4}{c}{\textbf{top-1 (\%)}} & \multicolumn{4}{c}{\textbf{$\Delta$ macro-F1 vs.\ reference}} \\
\cmidrule(lr){3-6}\cmidrule(lr){7-10}\cmidrule(lr){11-14}
\textbf{Arm} & \textbf{runs} & Test & OOD-R & OOD-D & Pil-T & Test & OOD-R & OOD-D & Pil-T & Test & OOD-R & OOD-D & Pil-T \\
\midrule
fine + chem (deployed), no R & 10 & 58.5{\tiny$\pm$5.1} & 39.2{\tiny$\pm$3.3} & 33.0{\tiny$\pm$4.5} & 31.7{\tiny$\pm$2.9} & 58.6{\tiny$\pm$1.1} & 40.8{\tiny$\pm$2.1} & 35.4{\tiny$\pm$2.9} & 44.0{\tiny$\pm$1.2} & -- & -- & -- & -- \\
fine, chem losses off, no R & 3 & 58.8{\tiny$\pm$5.1} & 40.6{\tiny$\pm$3.3} & 33.7{\tiny$\pm$4.7} & 31.1{\tiny$\pm$3.1} & 58.6{\tiny$\pm$1.1} & 42.0{\tiny$\pm$2.2} & 37.5{\tiny$\pm$4.3} & 44.7{\tiny$\pm$1.7} & +0.3{\tiny\,[-1.1, +1.6]} & \textbf{+1.4{\tiny\,[+0.8, +3.0]}} & +0.7{\tiny\,[-0.4, +1.7]} & -0.7{\tiny\,[-1.5, +0.8]} \\
coarse + chem, no R & 3 & 58.0{\tiny$\pm$5.6} & 36.2{\tiny$\pm$3.6} & 26.6{\tiny$\pm$4.2} & 27.6{\tiny$\pm$2.5} & 58.5{\tiny$\pm$1.1} & 41.6{\tiny$\pm$2.3} & 35.7{\tiny$\pm$4.3} & 43.5{\tiny$\pm$1.6} & -0.4{\tiny\,[-2.6, +1.0]} & \textbf{-3.0{\tiny\,[-4.8, -1.2]}} & \textbf{-6.4{\tiny\,[-7.9, -4.4]}} & \textbf{-4.1{\tiny\,[-7.8, -2.5]}} \\
coarse, no chemistry, no R & 3 & 57.4{\tiny$\pm$5.7} & 35.7{\tiny$\pm$3.2} & 26.2{\tiny$\pm$4.1} & 27.4{\tiny$\pm$2.8} & 58.4{\tiny$\pm$1.2} & 40.4{\tiny$\pm$2.2} & 34.6{\tiny$\pm$4.3} & 42.8{\tiny$\pm$1.8} & -1.0{\tiny\,[-3.1, +0.6]} & \textbf{-3.4{\tiny\,[-4.9, -1.3]}} & \textbf{-6.8{\tiny\,[-8.1, -4.8]}} & \textbf{-4.4{\tiny\,[-8.3, -2.7]}} \\
\midrule
fine + chem (deployed), + R & 10 & 69.2{\tiny$\pm$5.0} & 38.8{\tiny$\pm$3.5} & 37.2{\tiny$\pm$4.9} & 30.6{\tiny$\pm$3.0} & 73.7{\tiny$\pm$0.9} & 52.9{\tiny$\pm$2.2} & 48.5{\tiny$\pm$3.3} & 44.3{\tiny$\pm$1.1} & -- & -- & -- & -- \\
fine, chem losses off, + R & 3 & 69.3{\tiny$\pm$5.2} & 39.2{\tiny$\pm$3.5} & 38.2{\tiny$\pm$5.0} & 31.6{\tiny$\pm$3.1} & 73.6{\tiny$\pm$1.0} & 52.7{\tiny$\pm$2.3} & 49.0{\tiny$\pm$4.2} & 45.4{\tiny$\pm$1.4} & +0.0{\tiny\,[-3.4, +0.7]} & +0.4{\tiny\,[-0.6, +1.7]} & +1.0{\tiny\,[-0.3, +2.0]} & +1.0{\tiny\,[-0.0, +2.2]} \\
coarse + chem, + R & 3 & 68.1{\tiny$\pm$6.0} & 36.7{\tiny$\pm$3.5} & 31.2{\tiny$\pm$4.8} & 27.4{\tiny$\pm$2.5} & 73.5{\tiny$\pm$1.0} & 52.2{\tiny$\pm$2.2} & 45.6{\tiny$\pm$4.5} & 45.3{\tiny$\pm$1.5} & -1.1{\tiny\,[-4.2, +0.9]} & \textbf{-2.1{\tiny\,[-3.7, -1.0]}} & \textbf{-6.0{\tiny\,[-8.0, -3.2]}} & \textbf{-3.3{\tiny\,[-7.2, -1.7]}} \\
coarse, no chemistry, + R & 3 & 67.9{\tiny$\pm$5.9} & 36.4{\tiny$\pm$3.6} & 30.6{\tiny$\pm$4.5} & 26.3{\tiny$\pm$2.3} & 73.3{\tiny$\pm$1.0} & 51.8{\tiny$\pm$2.3} & 45.8{\tiny$\pm$4.5} & 43.5{\tiny$\pm$1.4} & -1.3{\tiny\,[-4.4, +1.0]} & \textbf{-2.4{\tiny\,[-3.7, -1.3]}} & \textbf{-6.6{\tiny\,[-8.0, -4.6]}} & \textbf{-4.3{\tiny\,[-8.5, -2.8]}} \\
\bottomrule
\end{tabular}
\end{adjustbox}
\end{table*}

\paragraph{Labelled data.}
Table~\ref{tab:abl-dsize} fine-tunes on random subsets of \textsc{train}. Pretraining is worth most when labels are
scarce. At a thousand labelled holes it lifts every split, \textsc{pilbara-test} included, by a wide margin, and a
pretrained model fine-tuned on a thousand holes already beats a model trained from scratch on all of \textsc{train}
across provinces. As labels grow, its in-distribution gain vanishes and its shifted-split gain settles to the values of
Table~\ref{tab:abl-ladder}. Retrieval behaves the other way. Its in-distribution gain grows with the labelled set,
because the labelled part of the pool grows with it, and it adds nothing on \textsc{pilbara-test} at any size.

\begin{table*}[t]
\centering
\footnotesize
\setlength{\tabcolsep}{3pt}
\caption{\textbf{Labelled training holes.} Random subsets of \textsc{train}; the labelled part of the retrieval pool is restricted to the same subset and the archive part is unchanged; $\Delta$ vs.\ the plain model at the same size. $\pm$: half-width of the 95\% hole-cluster bootstrap interval ($B{=}1000$); trained rows also resample seeds. macro-F1 over classes with $\geq$20 events; top-1 on OOD-R / OOD-D is macro-averaged over exploration reports. $\Delta$: paired difference in macro-F1 (points) with its 95\% interval; bold when the interval excludes zero.}
\label{tab:abl-dsize}
\begin{adjustbox}{max width=\textwidth}
\begin{tabular}{@{}l c rrrr rrrr rrrr@{}}
\toprule
& & \multicolumn{4}{c}{\textbf{macro-F1 (\%)}} & \multicolumn{4}{c}{\textbf{top-1 (\%)}} & \multicolumn{4}{c}{\textbf{$\Delta$ macro-F1 vs.\ reference}} \\
\cmidrule(lr){3-6}\cmidrule(lr){7-10}\cmidrule(lr){11-14}
\textbf{Arm} & \textbf{runs} & Test & OOD-R & OOD-D & Pil-T & Test & OOD-R & OOD-D & Pil-T & Test & OOD-R & OOD-D & Pil-T \\
\midrule
1k, plain & 3 & 42.5{\tiny$\pm$5.5} & 17.0{\tiny$\pm$2.9} & 19.4{\tiny$\pm$2.8} & 17.4{\tiny$\pm$3.6} & 53.0{\tiny$\pm$1.2} & 35.0{\tiny$\pm$2.1} & 31.0{\tiny$\pm$2.8} & 34.0{\tiny$\pm$1.9} & -- & -- & -- & -- \\
1k, + PT & 3 & 56.5{\tiny$\pm$4.9} & 38.3{\tiny$\pm$3.7} & 29.7{\tiny$\pm$3.8} & 35.9{\tiny$\pm$2.8} & 57.4{\tiny$\pm$1.2} & 39.3{\tiny$\pm$2.1} & 32.5{\tiny$\pm$2.6} & 45.7{\tiny$\pm$1.6} & \textbf{+14.1{\tiny\,[+7.9, +18.8]}} & \textbf{+21.3{\tiny\,[+15.7, +24.4]}} & \textbf{+10.3{\tiny\,[+8.0, +14.6]}} & \textbf{+18.5{\tiny\,[+14.8, +22.6]}} \\
1k, + R & 3 & 43.6{\tiny$\pm$5.8} & 19.9{\tiny$\pm$3.9} & 20.0{\tiny$\pm$3.4} & 17.8{\tiny$\pm$3.2} & 56.7{\tiny$\pm$1.1} & 39.2{\tiny$\pm$2.2} & 35.8{\tiny$\pm$3.9} & 34.3{\tiny$\pm$2.3} & +1.1{\tiny\,[-4.0, +6.5]} & +2.9{\tiny\,[-1.8, +6.6]} & +0.6{\tiny\,[-1.7, +3.0]} & +0.4{\tiny\,[-3.8, +4.7]} \\
1k, + PT + R & 3 & 58.7{\tiny$\pm$4.8} & 36.5{\tiny$\pm$3.1} & 31.9{\tiny$\pm$4.0} & 36.7{\tiny$\pm$2.9} & 61.4{\tiny$\pm$1.5} & 45.8{\tiny$\pm$2.6} & 43.1{\tiny$\pm$2.8} & 45.5{\tiny$\pm$2.0} & \textbf{+16.2{\tiny\,[+10.2, +20.7]}} & \textbf{+19.5{\tiny\,[+15.9, +22.9]}} & \textbf{+12.5{\tiny\,[+9.9, +16.7]}} & \textbf{+19.4{\tiny\,[+15.5, +23.5]}} \\
3k, plain & 3 & 50.2{\tiny$\pm$5.9} & 23.6{\tiny$\pm$2.8} & 23.4{\tiny$\pm$3.7} & 24.0{\tiny$\pm$2.3} & 56.9{\tiny$\pm$1.2} & 37.1{\tiny$\pm$2.1} & 31.3{\tiny$\pm$2.6} & 38.9{\tiny$\pm$1.6} & -- & -- & -- & -- \\
3k, + PT & 3 & 57.3{\tiny$\pm$5.0} & 40.5{\tiny$\pm$3.4} & 31.3{\tiny$\pm$3.8} & 35.6{\tiny$\pm$3.2} & 58.0{\tiny$\pm$1.2} & 40.2{\tiny$\pm$2.1} & 33.2{\tiny$\pm$2.8} & 45.4{\tiny$\pm$1.5} & \textbf{+7.1{\tiny\,[+1.8, +11.5]}} & \textbf{+16.9{\tiny\,[+12.0, +18.6]}} & \textbf{+7.9{\tiny\,[+5.8, +11.0]}} & \textbf{+11.6{\tiny\,[+9.3, +15.9]}} \\
3k, + R & 3 & 56.2{\tiny$\pm$6.3} & 24.8{\tiny$\pm$2.8} & 27.6{\tiny$\pm$3.9} & 21.8{\tiny$\pm$2.3} & 66.0{\tiny$\pm$1.1} & 48.9{\tiny$\pm$2.2} & 44.5{\tiny$\pm$4.8} & 38.2{\tiny$\pm$1.6} & \textbf{+6.0{\tiny\,[+2.4, +10.1]}} & +1.3{\tiny\,[-0.1, +4.8]} & \textbf{+4.1{\tiny\,[+2.1, +5.9]}} & -2.2{\tiny\,[-3.7, +0.8]} \\
3k, + PT + R & 3 & 62.5{\tiny$\pm$5.2} & 41.3{\tiny$\pm$3.3} & 33.5{\tiny$\pm$4.2} & 36.4{\tiny$\pm$3.0} & 66.6{\tiny$\pm$1.1} & 48.4{\tiny$\pm$2.7} & 40.6{\tiny$\pm$3.1} & 46.0{\tiny$\pm$1.3} & \textbf{+12.3{\tiny\,[+7.6, +16.5]}} & \textbf{+17.7{\tiny\,[+13.4, +19.6]}} & \textbf{+10.1{\tiny\,[+7.7, +12.5]}} & \textbf{+12.4{\tiny\,[+10.3, +16.5]}} \\
9k, plain & 3 & 57.4{\tiny$\pm$5.1} & 30.8{\tiny$\pm$5.5} & 26.4{\tiny$\pm$3.9} & 26.3{\tiny$\pm$2.9} & 58.2{\tiny$\pm$1.1} & 39.0{\tiny$\pm$1.9} & 33.6{\tiny$\pm$4.2} & 40.4{\tiny$\pm$1.7} & -- & -- & -- & -- \\
9k, + PT & 3 & 58.1{\tiny$\pm$4.8} & 40.4{\tiny$\pm$3.5} & 33.1{\tiny$\pm$4.7} & 34.4{\tiny$\pm$2.8} & 58.5{\tiny$\pm$1.1} & 39.9{\tiny$\pm$2.1} & 33.8{\tiny$\pm$2.7} & 44.6{\tiny$\pm$1.3} & +0.7{\tiny\,[-1.0, +5.9]} & \textbf{+9.5{\tiny\,[+4.8, +14.3]}} & \textbf{+6.7{\tiny\,[+4.4, +8.8]}} & \textbf{+8.1{\tiny\,[+5.9, +10.9]}} \\
9k, + R & 3 & 65.5{\tiny$\pm$5.5} & 33.0{\tiny$\pm$5.0} & 29.6{\tiny$\pm$4.2} & 23.5{\tiny$\pm$4.6} & 70.7{\tiny$\pm$1.0} & 50.0{\tiny$\pm$2.4} & 44.0{\tiny$\pm$4.6} & 40.1{\tiny$\pm$1.6} & \textbf{+8.1{\tiny\,[+6.2, +10.6]}} & +2.2{\tiny\,[-4.2, +7.7]} & \textbf{+3.2{\tiny\,[+1.7, +4.8]}} & -2.8{\tiny\,[-6.4, +1.9]} \\
9k, + PT + R & 3 & 66.0{\tiny$\pm$5.1} & 39.1{\tiny$\pm$3.9} & 37.4{\tiny$\pm$4.6} & 33.9{\tiny$\pm$3.1} & 70.8{\tiny$\pm$1.0} & 51.3{\tiny$\pm$2.4} & 46.7{\tiny$\pm$4.5} & 45.0{\tiny$\pm$1.5} & \textbf{+8.6{\tiny\,[+6.7, +14.3]}} & \textbf{+8.3{\tiny\,[+3.2, +14.2]}} & \textbf{+11.0{\tiny\,[+8.8, +13.4]}} & \textbf{+7.6{\tiny\,[+5.3, +10.8]}} \\
27k (all), plain & 3 & 59.5{\tiny$\pm$4.8} & 37.2{\tiny$\pm$3.5} & 28.5{\tiny$\pm$4.3} & 29.0{\tiny$\pm$3.0} & 58.5{\tiny$\pm$1.1} & 41.4{\tiny$\pm$2.4} & 36.3{\tiny$\pm$4.2} & 42.0{\tiny$\pm$1.3} & -- & -- & -- & -- \\
27k (all), + PT & 3 & 58.8{\tiny$\pm$5.0} & 39.0{\tiny$\pm$3.2} & 33.4{\tiny$\pm$4.7} & 31.6{\tiny$\pm$2.9} & 58.5{\tiny$\pm$1.1} & 42.0{\tiny$\pm$2.4} & 37.3{\tiny$\pm$4.4} & 43.4{\tiny$\pm$1.5} & -0.7{\tiny\,[-2.3, +1.1]} & \textbf{+1.8{\tiny\,[+0.8, +3.4]}} & \textbf{+4.8{\tiny\,[+3.3, +6.3]}} & \textbf{+2.6{\tiny\,[+1.5, +4.2]}} \\
27k (all), + R & 3 & 69.5{\tiny$\pm$4.9} & 38.3{\tiny$\pm$3.6} & 33.9{\tiny$\pm$4.8} & 26.9{\tiny$\pm$3.5} & 73.4{\tiny$\pm$0.9} & 52.4{\tiny$\pm$2.3} & 46.2{\tiny$\pm$4.5} & 42.8{\tiny$\pm$1.2} & \textbf{+10.0{\tiny\,[+8.1, +12.4]}} & +1.1{\tiny\,[-0.6, +2.7]} & \textbf{+5.4{\tiny\,[+3.8, +6.9]}} & -2.1{\tiny\,[-4.8, +0.6]} \\
27k (all), + PT + R & 3 & 69.2{\tiny$\pm$5.1} & 39.1{\tiny$\pm$3.6} & 37.5{\tiny$\pm$4.9} & 31.1{\tiny$\pm$3.0} & 73.7{\tiny$\pm$1.0} & 53.0{\tiny$\pm$2.4} & 49.1{\tiny$\pm$4.1} & 44.4{\tiny$\pm$1.3} & \textbf{+9.7{\tiny\,[+7.7, +12.8]}} & \textbf{+1.9{\tiny\,[+0.2, +4.2]}} & \textbf{+9.0{\tiny\,[+7.3, +10.7]}} & \textbf{+2.1{\tiny\,[+0.7, +4.4]}} \\
\bottomrule
\end{tabular}
\end{adjustbox}
\end{table*}

\subsubsection{Retrieval}
\label{app:ptrk-retrieval}

\paragraph{Number of neighbours.}
Table~\ref{tab:abl-topk} varies $k$. In distribution accuracy keeps rising up to the largest $k$ tried. Across provinces
it moves the other way, significantly at the largest $k$: the farther neighbours of a Pilbara hole resemble it less than
the nearest few, and averaging them dilutes the evidence. A single neighbour already gives about half of the gain on the
distal ring and is slightly better than $k{=}10$ across provinces. The deployed $k{=}10$ sits on the plateau of the shifted splits; a user who cares only about in-distribution
accuracy should raise it. Restricting neighbours to a 1\,km radius hurts on the distal ring, where the nearest evidence
lies farther away, and re-ranking candidates by mutual agreement does not help anywhere. Distance is the better selection
rule.

\begin{table*}[t]
\centering
\footnotesize
\setlength{\tabcolsep}{3pt}
\caption{\textbf{Number of retrieved neighbours.} Full recipe; radius: neighbours restricted to 1\,km; consensus: candidates re-ranked by agreement with each other. $\Delta$ vs.\ the deployed $k{=}10$. $\pm$: half-width of the 95\% hole-cluster bootstrap interval ($B{=}1000$); trained rows also resample seeds. macro-F1 over classes with $\geq$20 events; top-1 on OOD-R / OOD-D is macro-averaged over exploration reports. $\Delta$: paired difference in macro-F1 (points) with its 95\% interval; bold when the interval excludes zero.}
\label{tab:abl-topk}
\begin{adjustbox}{max width=\textwidth}
\begin{tabular}{@{}l c rrrr rrrr rrrr@{}}
\toprule
& & \multicolumn{4}{c}{\textbf{macro-F1 (\%)}} & \multicolumn{4}{c}{\textbf{top-1 (\%)}} & \multicolumn{4}{c}{\textbf{$\Delta$ macro-F1 vs.\ reference}} \\
\cmidrule(lr){3-6}\cmidrule(lr){7-10}\cmidrule(lr){11-14}
\textbf{Arm} & \textbf{runs} & Test & OOD-R & OOD-D & Pil-T & Test & OOD-R & OOD-D & Pil-T & Test & OOD-R & OOD-D & Pil-T \\
\midrule
no retrieval & 3 & 58.8{\tiny$\pm$5.1} & 39.0{\tiny$\pm$3.2} & 33.4{\tiny$\pm$4.6} & 31.6{\tiny$\pm$2.9} & 58.5{\tiny$\pm$1.1} & 42.0{\tiny$\pm$2.4} & 37.3{\tiny$\pm$4.3} & 43.4{\tiny$\pm$1.4} & \textbf{-10.4{\tiny\,[-13.3, -8.9]}} & -0.1{\tiny\,[-1.6, +1.3]} & \textbf{-4.1{\tiny\,[-5.5, -2.9]}} & +0.5{\tiny\,[-1.3, +1.9]} \\
$k{=}1$ & 3 & 63.1{\tiny$\pm$5.0} & 39.9{\tiny$\pm$3.5} & 35.6{\tiny$\pm$4.9} & 32.5{\tiny$\pm$3.0} & 64.6{\tiny$\pm$1.0} & 47.1{\tiny$\pm$2.4} & 42.8{\tiny$\pm$4.5} & 44.7{\tiny$\pm$1.4} & \textbf{-6.1{\tiny\,[-7.2, -3.5]}} & +0.8{\tiny\,[-0.9, +1.9]} & \textbf{-1.9{\tiny\,[-3.2, -1.1]}} & \textbf{+1.4{\tiny\,[+0.5, +2.4]}} \\
$k{=}3$ & 3 & 65.7{\tiny$\pm$5.1} & 39.6{\tiny$\pm$3.7} & 35.5{\tiny$\pm$4.9} & 31.5{\tiny$\pm$3.1} & 68.9{\tiny$\pm$1.0} & 48.7{\tiny$\pm$2.3} & 44.6{\tiny$\pm$4.1} & 44.6{\tiny$\pm$1.3} & \textbf{-3.4{\tiny\,[-4.4, -0.6]}} & +0.4{\tiny\,[-1.1, +1.3]} & \textbf{-2.0{\tiny\,[-3.1, -1.4]}} & +0.3{\tiny\,[-0.9, +1.6]} \\
$k{=}5$ & 3 & 66.7{\tiny$\pm$5.1} & 38.7{\tiny$\pm$3.5} & 36.7{\tiny$\pm$4.9} & 32.3{\tiny$\pm$3.0} & 70.9{\tiny$\pm$0.9} & 50.3{\tiny$\pm$2.5} & 46.9{\tiny$\pm$4.2} & 44.4{\tiny$\pm$1.3} & \textbf{-2.5{\tiny\,[-3.4, -0.6]}} & -0.4{\tiny\,[-1.2, +0.7]} & \textbf{-0.8{\tiny\,[-1.4, -0.2]}} & \textbf{+1.2{\tiny\,[+0.3, +2.1]}} \\
$k{=}10$ & 3 & 69.2{\tiny$\pm$5.1} & 39.1{\tiny$\pm$3.6} & 37.5{\tiny$\pm$5.0} & 31.1{\tiny$\pm$2.9} & 73.7{\tiny$\pm$0.9} & 53.0{\tiny$\pm$2.3} & 49.1{\tiny$\pm$4.0} & 44.4{\tiny$\pm$1.3} & -- & -- & -- & -- \\
$k{=}20$ & 3 & 70.1{\tiny$\pm$5.4} & 37.8{\tiny$\pm$3.5} & 37.3{\tiny$\pm$5.0} & 30.4{\tiny$\pm$2.9} & 76.1{\tiny$\pm$1.0} & 54.1{\tiny$\pm$2.6} & 49.2{\tiny$\pm$3.6} & 44.2{\tiny$\pm$1.2} & +0.9{\tiny\,[-2.4, +1.9]} & -1.4{\tiny\,[-2.3, +0.4]} & -0.2{\tiny\,[-1.3, +0.5]} & -0.7{\tiny\,[-1.4, +0.0]} \\
$k{=}40$ & 3 & 71.1{\tiny$\pm$5.2} & 38.5{\tiny$\pm$3.8} & 38.3{\tiny$\pm$5.2} & 30.0{\tiny$\pm$3.1} & 77.5{\tiny$\pm$0.9} & 53.9{\tiny$\pm$2.6} & 50.2{\tiny$\pm$3.3} & 44.4{\tiny$\pm$1.3} & +1.9{\tiny\,[-0.4, +3.1]} & -0.6{\tiny\,[-2.3, +0.5]} & +0.8{\tiny\,[-0.1, +1.6]} & \textbf{-1.1{\tiny\,[-1.8, -0.1]}} \\
$k{=}10$, radius 1\,km & 3 & 69.3{\tiny$\pm$4.9} & 39.3{\tiny$\pm$3.7} & 36.1{\tiny$\pm$4.8} & 31.3{\tiny$\pm$3.1} & 73.6{\tiny$\pm$0.9} & 48.3{\tiny$\pm$2.3} & 46.6{\tiny$\pm$4.8} & 44.0{\tiny$\pm$1.2} & +0.2{\tiny\,[-0.7, +2.0]} & +0.2{\tiny\,[-1.2, +1.1]} & \textbf{-1.4{\tiny\,[-2.0, -0.9]}} & +0.2{\tiny\,[-0.5, +1.2]} \\
$k{=}10$, consensus & 3 & 68.2{\tiny$\pm$5.3} & 36.9{\tiny$\pm$3.2} & 37.2{\tiny$\pm$5.0} & 31.7{\tiny$\pm$2.9} & 72.1{\tiny$\pm$1.0} & 53.3{\tiny$\pm$2.4} & 45.4{\tiny$\pm$3.2} & 44.5{\tiny$\pm$1.2} & -1.0{\tiny\,[-4.1, +0.2]} & -2.2{\tiny\,[-3.2, +0.6]} & -0.4{\tiny\,[-1.8, +0.5]} & +0.6{\tiny\,[-0.1, +1.7]} \\
\bottomrule
\end{tabular}
\end{adjustbox}
\end{table*}

\paragraph{Pool composition.}
Table~\ref{tab:abl-pool} changes what the model can retrieve from. Searching only the training split leaves
in-distribution accuracy unchanged but costs both rings heavily. Their nearest training holes are kilometres away, and the
archive supplies the neighbours they use. Across provinces the training-only pool is no worse, so the archive's Pilbara
holes are not what carries the cross-province result. Table~\ref{tab:abl-poolcurve} makes this explicit by thinning the
357 Pilbara holes that an earlier release set aside as a target-region subset. Removing any number of them leaves
\textsc{pilbara-test} unchanged within noise in either pool, which is why the subset was folded back into the archive.

\begin{table*}[t]
\centering
\footnotesize
\setlength{\tabcolsep}{3pt}
\caption{\textbf{Retrieval pool.} Full recipe; target context: the 357 Pilbara holes formerly released as a fine-tuning subset. $\Delta$ vs.\ the deployed train + archive pool. $\pm$: half-width of the 95\% hole-cluster bootstrap interval ($B{=}1000$); trained rows also resample seeds. macro-F1 over classes with $\geq$20 events; top-1 on OOD-R / OOD-D is macro-averaged over exploration reports. $\Delta$: paired difference in macro-F1 (points) with its 95\% interval; bold when the interval excludes zero.}
\label{tab:abl-pool}
\begin{adjustbox}{max width=\textwidth}
\begin{tabular}{@{}l c rrrr rrrr rrrr@{}}
\toprule
& & \multicolumn{4}{c}{\textbf{macro-F1 (\%)}} & \multicolumn{4}{c}{\textbf{top-1 (\%)}} & \multicolumn{4}{c}{\textbf{$\Delta$ macro-F1 vs.\ reference}} \\
\cmidrule(lr){3-6}\cmidrule(lr){7-10}\cmidrule(lr){11-14}
\textbf{Arm} & \textbf{runs} & Test & OOD-R & OOD-D & Pil-T & Test & OOD-R & OOD-D & Pil-T & Test & OOD-R & OOD-D & Pil-T \\
\midrule
none & 10 & 58.5{\tiny$\pm$5.1} & 39.2{\tiny$\pm$3.2} & 33.0{\tiny$\pm$4.2} & 31.7{\tiny$\pm$2.8} & 58.6{\tiny$\pm$1.1} & 40.8{\tiny$\pm$1.8} & 35.4{\tiny$\pm$3.1} & 44.0{\tiny$\pm$1.3} & \textbf{-10.8{\tiny\,[-13.2, -9.1]}} & +0.4{\tiny\,[-1.6, +1.8]} & \textbf{-4.2{\tiny\,[-5.5, -3.3]}} & +1.1{\tiny\,[-0.4, +2.3]} \\
train only & 3 & 69.5{\tiny$\pm$5.0} & 30.8{\tiny$\pm$2.5} & 29.7{\tiny$\pm$4.1} & 32.3{\tiny$\pm$3.2} & 73.6{\tiny$\pm$1.0} & 37.9{\tiny$\pm$2.1} & 33.8{\tiny$\pm$3.8} & 43.4{\tiny$\pm$1.6} & +0.3{\tiny\,[-1.3, +0.9]} & \textbf{-8.0{\tiny\,[-9.2, -4.7]}} & \textbf{-7.5{\tiny\,[-9.4, -5.9]}} & \textbf{+1.7{\tiny\,[+0.3, +3.0]}} \\
train + target context & 3 & 69.5{\tiny$\pm$4.9} & 30.8{\tiny$\pm$2.5} & 29.7{\tiny$\pm$4.0} & 31.5{\tiny$\pm$3.0} & 73.6{\tiny$\pm$1.0} & 37.9{\tiny$\pm$2.1} & 33.8{\tiny$\pm$3.9} & 43.3{\tiny$\pm$1.7} & +0.3{\tiny\,[-1.3, +0.9]} & \textbf{-8.0{\tiny\,[-9.2, -4.7]}} & \textbf{-7.5{\tiny\,[-9.4, -5.9]}} & +0.8{\tiny\,[-0.3, +1.8]} \\
train + archive & 10 & 69.2{\tiny$\pm$5.0} & 38.8{\tiny$\pm$3.4} & 37.2{\tiny$\pm$4.6} & 30.6{\tiny$\pm$2.9} & 73.7{\tiny$\pm$0.9} & 52.9{\tiny$\pm$2.1} & 48.5{\tiny$\pm$3.2} & 44.3{\tiny$\pm$1.1} & -- & -- & -- & -- \\
\bottomrule
\end{tabular}
\end{adjustbox}
\end{table*}

\begin{table*}[t]
\centering
\footnotesize
\setlength{\tabcolsep}{3pt}
\caption{\textbf{Target-region holes in the retrieval pool.} Full recipe; $n$ of the 357 Pilbara context holes kept in the pool. $\Delta$ vs.\ $n{=}357$ in the same pool. $\pm$: half-width of the 95\% hole-cluster bootstrap interval ($B{=}1000$); trained rows also resample seeds. macro-F1 over classes with $\geq$20 events; top-1 on OOD-R / OOD-D is macro-averaged over exploration reports. $\Delta$: paired difference in macro-F1 (points) with its 95\% interval; bold when the interval excludes zero.}
\label{tab:abl-poolcurve}
\begin{adjustbox}{max width=\textwidth}
\begin{tabular}{@{}l c rrrr rrrr rrrr@{}}
\toprule
& & \multicolumn{4}{c}{\textbf{macro-F1 (\%)}} & \multicolumn{4}{c}{\textbf{top-1 (\%)}} & \multicolumn{4}{c}{\textbf{$\Delta$ macro-F1 vs.\ reference}} \\
\cmidrule(lr){3-6}\cmidrule(lr){7-10}\cmidrule(lr){11-14}
\textbf{Arm} & \textbf{runs} & Test & OOD-R & OOD-D & Pil-T & Test & OOD-R & OOD-D & Pil-T & Test & OOD-R & OOD-D & Pil-T \\
\midrule
train pool, $n{=}$0 & 3 & 69.5{\tiny$\pm$5.0} & 30.8{\tiny$\pm$2.5} & 29.7{\tiny$\pm$4.3} & 32.3{\tiny$\pm$3.2} & 73.6{\tiny$\pm$0.9} & 37.9{\tiny$\pm$2.2} & 33.8{\tiny$\pm$3.7} & 43.4{\tiny$\pm$1.6} & +0.0{\tiny\,[-0.9, +0.9]} & +0.0{\tiny\,[-0.5, +0.4]} & +0.0{\tiny\,[-1.1, +1.1]} & +0.8{\tiny\,[-0.4, +1.9]} \\
train pool, $n{=}$10 & 3 & 69.5{\tiny$\pm$5.0} & 30.8{\tiny$\pm$2.5} & 29.7{\tiny$\pm$4.3} & 31.9{\tiny$\pm$3.1} & 73.6{\tiny$\pm$1.0} & 37.9{\tiny$\pm$2.2} & 33.8{\tiny$\pm$4.0} & 43.1{\tiny$\pm$1.7} & +0.0{\tiny\,[-0.8, +1.6]} & +0.0{\tiny\,[-0.5, +0.4]} & +0.0{\tiny\,[-1.2, +1.2]} & +0.4{\tiny\,[-0.6, +1.4]} \\
train pool, $n{=}$50 & 3 & 69.5{\tiny$\pm$4.9} & 30.8{\tiny$\pm$2.5} & 29.7{\tiny$\pm$4.4} & 30.6{\tiny$\pm$3.4} & 73.6{\tiny$\pm$1.0} & 37.9{\tiny$\pm$2.3} & 33.8{\tiny$\pm$3.9} & 43.4{\tiny$\pm$1.5} & +0.0{\tiny\,[-0.8, +1.3]} & +0.0{\tiny\,[-0.5, +0.4]} & +0.0{\tiny\,[-1.1, +1.1]} & -0.9{\tiny\,[-2.5, +0.6]} \\
train pool, $n{=}$100 & 3 & 69.5{\tiny$\pm$4.9} & 30.8{\tiny$\pm$2.5} & 29.7{\tiny$\pm$4.3} & 30.8{\tiny$\pm$3.2} & 73.6{\tiny$\pm$1.0} & 37.9{\tiny$\pm$2.2} & 33.8{\tiny$\pm$3.7} & 43.5{\tiny$\pm$1.6} & +0.0{\tiny\,[-0.7, +0.9]} & +0.0{\tiny\,[-0.5, +0.4]} & +0.0{\tiny\,[-1.2, +1.1]} & -0.7{\tiny\,[-1.8, +0.4]} \\
train pool, $n{=}$200 & 3 & 69.5{\tiny$\pm$5.1} & 30.8{\tiny$\pm$2.5} & 29.7{\tiny$\pm$4.3} & 31.4{\tiny$\pm$2.9} & 73.6{\tiny$\pm$1.0} & 37.9{\tiny$\pm$2.2} & 33.8{\tiny$\pm$3.8} & 43.6{\tiny$\pm$1.5} & +0.0{\tiny\,[-0.7, +0.9]} & +0.0{\tiny\,[-0.5, +0.4]} & +0.0{\tiny\,[-1.1, +1.2]} & -0.1{\tiny\,[-0.9, +0.8]} \\
train pool, $n{=}$357 & 3 & 69.5{\tiny$\pm$5.0} & 30.8{\tiny$\pm$2.5} & 29.7{\tiny$\pm$4.4} & 31.5{\tiny$\pm$3.1} & 73.6{\tiny$\pm$1.0} & 37.9{\tiny$\pm$2.2} & 33.8{\tiny$\pm$3.8} & 43.3{\tiny$\pm$1.6} & -- & -- & -- & -- \\
train + archive, $n{=}$0 & 3 & 69.2{\tiny$\pm$5.0} & 39.1{\tiny$\pm$3.6} & 37.5{\tiny$\pm$5.0} & 31.2{\tiny$\pm$3.0} & 73.7{\tiny$\pm$0.9} & 53.0{\tiny$\pm$2.3} & 49.1{\tiny$\pm$4.2} & 43.7{\tiny$\pm$1.1} & +0.0{\tiny\,[-1.1, +0.8]} & -0.0{\tiny\,[-0.9, +1.0]} & +0.0{\tiny\,[-0.3, +0.3]} & +0.1{\tiny\,[-0.4, +1.0]} \\
train + archive, $n{=}$10 & 3 & 69.2{\tiny$\pm$5.0} & 39.1{\tiny$\pm$3.7} & 37.5{\tiny$\pm$4.9} & 31.1{\tiny$\pm$3.1} & 73.7{\tiny$\pm$0.9} & 53.0{\tiny$\pm$2.3} & 49.1{\tiny$\pm$4.2} & 44.0{\tiny$\pm$1.2} & +0.0{\tiny\,[-0.9, +1.2]} & +0.0{\tiny\,[-0.9, +1.0]} & +0.0{\tiny\,[-0.3, +0.3]} & -0.1{\tiny\,[-0.8, +0.8]} \\
train + archive, $n{=}$50 & 3 & 69.2{\tiny$\pm$5.0} & 39.1{\tiny$\pm$3.6} & 37.5{\tiny$\pm$5.0} & 30.7{\tiny$\pm$3.1} & 73.7{\tiny$\pm$0.9} & 53.0{\tiny$\pm$2.3} & 49.1{\tiny$\pm$4.3} & 44.4{\tiny$\pm$1.2} & +0.0{\tiny\,[-0.8, +0.7]} & +0.0{\tiny\,[-0.9, +0.9]} & +0.0{\tiny\,[-0.3, +0.3]} & -0.4{\tiny\,[-1.3, +0.6]} \\
train + archive, $n{=}$100 & 3 & 69.2{\tiny$\pm$5.0} & 39.1{\tiny$\pm$3.6} & 37.5{\tiny$\pm$5.0} & 31.0{\tiny$\pm$3.0} & 73.7{\tiny$\pm$0.9} & 53.0{\tiny$\pm$2.4} & 49.1{\tiny$\pm$4.2} & 44.5{\tiny$\pm$1.2} & +0.0{\tiny\,[-0.8, +1.0]} & -0.0{\tiny\,[-0.9, +0.9]} & +0.0{\tiny\,[-0.3, +0.3]} & -0.1{\tiny\,[-0.8, +0.7]} \\
train + archive, $n{=}$200 & 3 & 69.2{\tiny$\pm$5.0} & 39.1{\tiny$\pm$3.7} & 37.5{\tiny$\pm$5.0} & 31.2{\tiny$\pm$3.0} & 73.7{\tiny$\pm$0.9} & 53.0{\tiny$\pm$2.3} & 49.1{\tiny$\pm$4.3} & 44.5{\tiny$\pm$1.2} & +0.0{\tiny\,[-0.8, +0.7]} & -0.0{\tiny\,[-0.9, +0.9]} & +0.0{\tiny\,[-0.3, +0.3]} & +0.1{\tiny\,[-0.5, +0.9]} \\
train + archive, $n{=}$357 & 3 & 69.2{\tiny$\pm$5.0} & 39.1{\tiny$\pm$3.6} & 37.5{\tiny$\pm$5.0} & 31.1{\tiny$\pm$3.0} & 73.7{\tiny$\pm$0.9} & 53.0{\tiny$\pm$2.4} & 49.1{\tiny$\pm$4.1} & 44.4{\tiny$\pm$1.3} & -- & -- & -- & -- \\
\bottomrule
\end{tabular}
\end{adjustbox}
\end{table*}

\paragraph{The same pool for the spatial baselines.}
A larger pool could favour our model simply by showing it more holes. We therefore refit spatial $k$-NN, indicator
kriging and SIS on the same train + archive pool (Table~\ref{tab:app-task1-baselines}). The archive lifts them under
every shift but leaves them far below the recipe on every shifted split (Table~\ref{tab:app-task1-paired}). What the recipe
adds is not access to the pool but a sequence model that reads the neighbours as context.

\paragraph{Design choices.}
Table~\ref{tab:abl-rkknob} changes the retrieval module one choice at a time. Neighbour dropout, elevation alignment,
the self-exclusion radius and a learned gate on the retrieval term all stay within about a point of the deployed setting.
The gate, initialised open, learns nothing that the additive fusion does not already do. The one change that matters is
how neighbour evidence is read. The deployed module lets a neighbour vote only at depths where one of its beds starts, so
across provinces only about a third of the cells receive any evidence. Letting every bed vote over its whole thickness
gives the largest in-distribution gain of any change we tried, but costs significantly on \textsc{pilbara-test}: denser
evidence from the same neighbours helps where they are right and hurts where they are not. We report it as the main
untapped in-distribution gain and leave it out of the deployed recipe, whose target is transfer.

\begin{table*}[t]
\centering
\footnotesize
\setlength{\tabcolsep}{3pt}
\caption{\textbf{Retrieval design choices.} Full recipe, one change at a time. $\Delta$ vs.\ the deployed setting. $\pm$: half-width of the 95\% hole-cluster bootstrap interval ($B{=}1000$); trained rows also resample seeds. macro-F1 over classes with $\geq$20 events; top-1 on OOD-R / OOD-D is macro-averaged over exploration reports. $\Delta$: paired difference in macro-F1 (points) with its 95\% interval; bold when the interval excludes zero.}
\label{tab:abl-rkknob}
\begin{adjustbox}{max width=\textwidth}
\begin{tabular}{@{}l c rrrr rrrr rrrr@{}}
\toprule
& & \multicolumn{4}{c}{\textbf{macro-F1 (\%)}} & \multicolumn{4}{c}{\textbf{top-1 (\%)}} & \multicolumn{4}{c}{\textbf{$\Delta$ macro-F1 vs.\ reference}} \\
\cmidrule(lr){3-6}\cmidrule(lr){7-10}\cmidrule(lr){11-14}
\textbf{Arm} & \textbf{runs} & Test & OOD-R & OOD-D & Pil-T & Test & OOD-R & OOD-D & Pil-T & Test & OOD-R & OOD-D & Pil-T \\
\midrule
deployed & 10 & 69.2{\tiny$\pm$4.9} & 38.8{\tiny$\pm$3.5} & 37.2{\tiny$\pm$4.9} & 30.6{\tiny$\pm$2.9} & 73.7{\tiny$\pm$0.9} & 52.9{\tiny$\pm$2.2} & 48.5{\tiny$\pm$3.2} & 44.3{\tiny$\pm$1.1} & -- & -- & -- & -- \\
neighbour dropout 0 & 3 & 69.7{\tiny$\pm$5.0} & 39.4{\tiny$\pm$3.8} & 37.8{\tiny$\pm$5.0} & 30.9{\tiny$\pm$2.9} & 73.8{\tiny$\pm$1.0} & 53.4{\tiny$\pm$2.2} & 49.4{\tiny$\pm$4.2} & 44.2{\tiny$\pm$1.3} & +0.4{\tiny\,[-0.5, +1.7]} & +0.6{\tiny\,[-0.9, +2.4]} & +0.6{\tiny\,[-0.7, +1.7]} & +0.2{\tiny\,[-0.6, +0.9]} \\
neighbour dropout 0.6 & 3 & 68.9{\tiny$\pm$5.3} & 39.0{\tiny$\pm$3.6} & 37.3{\tiny$\pm$5.0} & 31.9{\tiny$\pm$3.0} & 73.6{\tiny$\pm$1.0} & 52.6{\tiny$\pm$2.4} & 49.2{\tiny$\pm$4.2} & 44.9{\tiny$\pm$1.3} & -0.3{\tiny\,[-3.3, +0.3]} & +0.2{\tiny\,[-0.9, +1.2]} & +0.0{\tiny\,[-0.6, +0.5]} & \textbf{+1.2{\tiny\,[+0.6, +2.1]}} \\
no elevation alignment & 3 & 70.2{\tiny$\pm$4.9} & 39.5{\tiny$\pm$3.7} & 37.6{\tiny$\pm$4.9} & 30.9{\tiny$\pm$3.1} & 74.8{\tiny$\pm$1.0} & 53.6{\tiny$\pm$2.3} & 49.6{\tiny$\pm$4.1} & 44.3{\tiny$\pm$1.4} & \textbf{+0.9{\tiny\,[+0.1, +3.0]}} & +0.7{\tiny\,[-0.7, +1.6]} & +0.4{\tiny\,[-0.2, +0.9]} & +0.2{\tiny\,[-0.6, +1.1]} \\
no 25\,m self-exclusion & 3 & 70.3{\tiny$\pm$4.9} & 39.0{\tiny$\pm$3.6} & 37.5{\tiny$\pm$5.0} & 30.4{\tiny$\pm$2.9} & 74.3{\tiny$\pm$1.0} & 52.5{\tiny$\pm$2.4} & 49.5{\tiny$\pm$4.0} & 44.2{\tiny$\pm$1.2} & \textbf{+1.0{\tiny\,[+0.2, +3.5]}} & +0.2{\tiny\,[-1.1, +1.0]} & +0.2{\tiny\,[-0.4, +0.8]} & -0.2{\tiny\,[-1.2, +0.8]} \\
learned gate & 3 & 69.1{\tiny$\pm$5.1} & 39.1{\tiny$\pm$3.6} & 37.5{\tiny$\pm$5.0} & 31.1{\tiny$\pm$2.9} & 73.7{\tiny$\pm$0.9} & 53.0{\tiny$\pm$2.3} & 49.1{\tiny$\pm$4.3} & 44.4{\tiny$\pm$1.3} & -0.1{\tiny\,[-0.9, +0.5]} & +0.4{\tiny\,[-0.8, +1.2]} & +0.3{\tiny\,[-0.2, +0.7]} & +0.4{\tiny\,[-0.3, +1.2]} \\
learned gate + base-of-cover & 3 & 69.1{\tiny$\pm$5.1} & 39.1{\tiny$\pm$3.6} & 37.5{\tiny$\pm$4.9} & 31.1{\tiny$\pm$3.0} & 73.7{\tiny$\pm$0.9} & 53.0{\tiny$\pm$2.3} & 49.1{\tiny$\pm$4.3} & 44.4{\tiny$\pm$1.2} & -0.1{\tiny\,[-1.0, +0.7]} & +0.4{\tiny\,[-0.8, +1.2]} & +0.3{\tiny\,[-0.2, +0.7]} & +0.4{\tiny\,[-0.3, +1.2]} \\
bed-filled votes & 3 & 73.6{\tiny$\pm$5.6} & 38.5{\tiny$\pm$3.7} & 38.0{\tiny$\pm$5.1} & 29.4{\tiny$\pm$2.7} & 81.1{\tiny$\pm$0.9} & 55.2{\tiny$\pm$2.4} & 51.8{\tiny$\pm$3.9} & 44.0{\tiny$\pm$1.1} & \textbf{+4.4{\tiny\,[+1.6, +6.0]}} & -0.3{\tiny\,[-2.3, +0.7]} & +0.7{\tiny\,[-0.7, +2.1]} & \textbf{-1.2{\tiny\,[-2.5, -0.3]}} \\
\bottomrule
\end{tabular}
\end{adjustbox}
\end{table*}

\subsubsection{Where each component acts}
\label{app:ptrk-where}

Two analyses of where the components act are reported with the Task~1 and Task~2 results and are not repeated here.
Cover versus bedrock (Table~\ref{tab:app-task1-zones}) shows that retrieval is mainly a cover mechanism and pretraining
mainly a bedrock one, with the distal ring as the exception. The five-backbone comparison
(Tables~\ref{tab:app-task1-backbones} and~\ref{tab:app-task2-backbones}) shows that the division of labour of
Table~\ref{tab:abl-ladder} holds for every backbone on both tasks.

\paragraph{Token unit.}
Table~\ref{tab:app-granularity} keeps the Transformer backbone and the recipe fixed and changes only the token: a 1\,m
cell or a whole stratum. Retrieval helps in distribution at both levels. Pretraining does not: at the cell level it
lowers macro-F1 on most splits, whereas at the stratum level it raises it on all three shifted splits. The full recipe
at the cell level is therefore well below the stratum-level one under every shift. At one-metre resolution most targets
are \texttt{CONT}, so the pretraining signal is dominated by bed thickness rather than by lithological succession. The
stratum token removes that imbalance, which is why the recipe is built on it.

\providecolor{unitgray}{HTML}{E5E7EB}
\begin{table*}[t]
\centering
\footnotesize
\setlength{\tabcolsep}{3pt}
\caption{\textbf{Token unit: 1\,m cell vs.\ stratum.} The same Transformer backbone and the same recipe, retrieval (R) and corpus pretraining (PT), with a 1\,m cell or a whole stratum as the token. AR: plain autoregressive model. $\pm$: half-width of the 95\% hole-cluster bootstrap interval ($B{=}1000$); trained rows also resample seeds. macro-F1 over classes with $\geq$20 events; top-1 on OOD-R / OOD-D is macro-averaged over exploration reports.}
\label{tab:app-granularity}
\begin{adjustbox}{width=\textwidth}
\begin{tabular}{@{}l c rrrr rrrr@{}}
\toprule
& & \multicolumn{4}{c}{\textbf{macro-F1 (\%)}} & \multicolumn{4}{c}{\textbf{top-1 accuracy (\%)}} \\
\cmidrule(lr){3-6}\cmidrule(lr){7-10}
\textbf{Method} & \textbf{runs} & Test & OOD-R & OOD-D & Pil-T & Test & OOD-R & OOD-D & Pil-T \\
\midrule
\rowcolor{unitgray}
\multicolumn{10}{@{}l}{\textbf{Cell level (1\,m token, Transformer)}} \\
AR & 1 & 60.3{\tiny$\pm$5.1} & 26.5{\tiny$\pm$3.1} & 23.6{\tiny$\pm$3.3} & 22.5{\tiny$\pm$2.5} & 60.6{\tiny$\pm$1.2} & 39.7{\tiny$\pm$2.1} & 34.8{\tiny$\pm$3.0} & 38.2{\tiny$\pm$1.4} \\
+ R & 3 & \best{70.6}{\tiny$\pm$5.5} & 28.2{\tiny$\pm$2.8} & 25.0{\tiny$\pm$3.6} & 19.9{\tiny$\pm$1.9} & \best{80.6}{\tiny$\pm$0.9} & 51.4{\tiny$\pm$2.5} & 46.0{\tiny$\pm$3.5} & 39.0{\tiny$\pm$1.5} \\
+ PT & 3 & 53.3{\tiny$\pm$5.1} & 26.7{\tiny$\pm$2.7} & 19.4{\tiny$\pm$2.8} & 19.1{\tiny$\pm$2.3} & 57.9{\tiny$\pm$1.1} & 37.9{\tiny$\pm$2.0} & 33.5{\tiny$\pm$2.6} & 31.0{\tiny$\pm$1.6} \\
+ PT + R & 3 & 64.4{\tiny$\pm$5.6} & 28.2{\tiny$\pm$3.6} & 23.5{\tiny$\pm$3.5} & 18.5{\tiny$\pm$2.4} & 76.5{\tiny$\pm$1.1} & 49.9{\tiny$\pm$2.6} & 44.3{\tiny$\pm$4.2} & 30.7{\tiny$\pm$2.5} \\
\midrule
\rowcolor{unitgray}
\multicolumn{10}{@{}l}{\textbf{Layer level (stratum token, Transformer)}} \\
AR & 10 & 58.2{\tiny$\pm$5.1} & 34.4{\tiny$\pm$3.8} & 26.7{\tiny$\pm$3.9} & 24.4{\tiny$\pm$2.8} & 58.2{\tiny$\pm$1.1} & 40.2{\tiny$\pm$2.0} & 35.3{\tiny$\pm$3.1} & 38.1{\tiny$\pm$1.3} \\
+ R & 3 & 69.8{\tiny$\pm$4.9} & 36.6{\tiny$\pm$3.6} & 34.8{\tiny$\pm$4.6} & 24.0{\tiny$\pm$3.7} & 73.7{\tiny$\pm$1.0} & 52.0{\tiny$\pm$2.2} & \best{48.4}{\tiny$\pm$4.3} & 39.3{\tiny$\pm$1.4} \\
+ PT & 3 & 57.5{\tiny$\pm$5.2} & 38.8{\tiny$\pm$3.3} & 30.8{\tiny$\pm$3.9} & \best{28.1}{\tiny$\pm$3.1} & 58.1{\tiny$\pm$1.1} & 40.9{\tiny$\pm$2.3} & 36.4{\tiny$\pm$4.2} & 40.7{\tiny$\pm$1.3} \\
+ PT + R & 3 & 68.2{\tiny$\pm$5.3} & \best{39.6}{\tiny$\pm$3.2} & \best{34.9}{\tiny$\pm$4.2} & 26.0{\tiny$\pm$3.6} & 73.4{\tiny$\pm$1.0} & \best{52.5}{\tiny$\pm$2.4} & 48.3{\tiny$\pm$4.9} & \best{41.8}{\tiny$\pm$1.3} \\
\bottomrule
\end{tabular}
\end{adjustbox}
\end{table*}

\subsubsection{Tokenisation grid}
\label{app:ptrk-grid}

Changing the cell spacing changes the task, so the benchmark is frozen at 1\,m. Table~\ref{tab:abl-grid} retrains the
plain model and the full recipe at 0.5\,m and 2\,m as a sensitivity probe. The plain model is indifferent to the finer
grid and loses at the coarser one, where coarsening erases layer starts. The full recipe at 0.5\,m is slightly worse in
distribution and unchanged under shift. A likely reason is that a neighbour votes only in the cell where one of its beds
starts, and a half-metre cell is missed more often once the look-up is shifted by elevation. At 2\,m the full recipe is
unchanged in distribution and loses across provinces. Neither alternative improves any split significantly, and each
costs on at least one, so we keep the 1\,m grid, which also matches the logging resolution of the source data.

\begin{table*}[t]
\centering
\footnotesize
\setlength{\tabcolsep}{3pt}
\caption{\textbf{Tokenisation grid (without label smoothing).} Plain GRU and full recipe retrained at each spacing; $\Delta$ vs.\ 1\,m in the same arm. $\pm$: half-width of the 95\% hole-cluster bootstrap interval ($B{=}1000$); trained rows also resample seeds. macro-F1 over classes with $\geq$20 events; top-1 on OOD-R / OOD-D is macro-averaged over exploration reports. $\Delta$: paired difference in macro-F1 (points) with its 95\% interval; bold when the interval excludes zero.}
\label{tab:abl-grid}
\begin{adjustbox}{max width=\textwidth}
\begin{tabular}{@{}l c rrrr rrrr rrrr@{}}
\toprule
& & \multicolumn{4}{c}{\textbf{macro-F1 (\%)}} & \multicolumn{4}{c}{\textbf{top-1 (\%)}} & \multicolumn{4}{c}{\textbf{$\Delta$ macro-F1 vs.\ reference}} \\
\cmidrule(lr){3-6}\cmidrule(lr){7-10}\cmidrule(lr){11-14}
\textbf{Arm} & \textbf{runs} & Test & OOD-R & OOD-D & Pil-T & Test & OOD-R & OOD-D & Pil-T & Test & OOD-R & OOD-D & Pil-T \\
\midrule
0.5 m, plain & 3 & 60.2{\tiny$\pm$5.4} & 36.0{\tiny$\pm$3.6} & 27.2{\tiny$\pm$4.1} & 27.8{\tiny$\pm$2.5} & 59.5{\tiny$\pm$1.1} & 41.4{\tiny$\pm$2.4} & 36.4{\tiny$\pm$4.5} & 41.1{\tiny$\pm$1.2} & +0.7{\tiny\,[-3.0, +1.6]} & +1.1{\tiny\,[-0.3, +2.8]} & +1.1{\tiny\,[-0.5, +2.4]} & +1.3{\tiny\,[-0.4, +2.9]} \\
1 m, plain & 10 & 59.5{\tiny$\pm$4.8} & 34.9{\tiny$\pm$3.2} & 26.1{\tiny$\pm$3.8} & 26.5{\tiny$\pm$2.7} & 58.6{\tiny$\pm$1.1} & 40.2{\tiny$\pm$2.0} & 34.5{\tiny$\pm$3.2} & 40.2{\tiny$\pm$1.4} & -- & -- & -- & -- \\
2 m, plain & 3 & 55.5{\tiny$\pm$4.8} & 30.1{\tiny$\pm$3.1} & 23.6{\tiny$\pm$3.5} & 25.2{\tiny$\pm$2.8} & 55.4{\tiny$\pm$1.0} & 37.4{\tiny$\pm$2.2} & 31.8{\tiny$\pm$4.6} & 36.7{\tiny$\pm$2.2} & \textbf{-4.0{\tiny\,[-7.0, -1.0]}} & \textbf{-4.7{\tiny\,[-9.0, -1.5]}} & -2.5{\tiny\,[-5.4, +0.2]} & -1.4{\tiny\,[-4.5, +0.3]} \\
0.5 m, full recipe & 3 & 66.4{\tiny$\pm$5.7} & 39.8{\tiny$\pm$3.7} & 37.8{\tiny$\pm$5.0} & 31.6{\tiny$\pm$3.0} & 70.0{\tiny$\pm$1.0} & 52.4{\tiny$\pm$2.2} & 44.8{\tiny$\pm$5.0} & 45.2{\tiny$\pm$1.4} & \textbf{-2.8{\tiny\,[-6.7, -1.3]}} & +1.0{\tiny\,[-0.6, +2.1]} & +0.6{\tiny\,[-1.2, +1.6]} & +1.0{\tiny\,[-0.0, +2.0]} \\
1 m, full recipe & 10 & 69.2{\tiny$\pm$4.9} & 38.8{\tiny$\pm$3.6} & 37.2{\tiny$\pm$4.9} & 30.6{\tiny$\pm$3.1} & 73.7{\tiny$\pm$0.9} & 52.9{\tiny$\pm$2.2} & 48.5{\tiny$\pm$3.2} & 44.3{\tiny$\pm$1.1} & -- & -- & -- & -- \\
2 m, full recipe & 3 & 68.2{\tiny$\pm$5.2} & 41.0{\tiny$\pm$3.4} & 36.0{\tiny$\pm$4.7} & 27.7{\tiny$\pm$3.4} & 72.7{\tiny$\pm$1.0} & 53.5{\tiny$\pm$2.5} & 46.6{\tiny$\pm$3.1} & 40.9{\tiny$\pm$1.2} & -1.1{\tiny\,[-6.0, +1.3]} & +2.2{\tiny\,[-2.0, +3.4]} & -1.2{\tiny\,[-4.7, +2.3]} & \textbf{-3.0{\tiny\,[-6.6, -0.9]}} \\
\bottomrule
\end{tabular}
\end{adjustbox}
\end{table*}
\end{document}